%% file: main_bmvc.tex
\documentclass[10pt,twocolumn,letterpaper]{article}

\usepackage[pagenumbers]{cvpr}

\definecolor{cvprblue}{rgb}{0.21,0.49,0.74}
\usepackage[pagebackref,breaklinks,colorlinks,allcolors=cvprblue]{hyperref}

\usepackage{xcolor}
\usepackage{bm}
\usepackage{booktabs}
\usepackage{graphicx}
\usepackage{multirow}
\usepackage{tabularx}
\usepackage[normalem]{ulem}
\useunder{\uline}{\ul}{}
\usepackage{tabularray}
\usepackage{bbding}
\usepackage[export]{adjustbox}
\usepackage{fontawesome}

\input{preamble}

\newcommand{\namelong}{Pose-free Articulated Object Modeling from Sparse Views\xspace}
\newcommand{\nameshort}{PAOLI\xspace}
\title{{\nameshort:} \namelong}

\author{%
  Jianning Deng \quad Kartic Subr \quad Hakan Bilen\\
  University of Edinburgh\\
  {\tt\small \{jianning.deng, k.subr, h.bilen\}@ed.ac.uk}
}

\begin{document}

\maketitle

\input{sec/00abstract}
\input{sec/01introduction}

\input{sec/02relatedwork}

\input{sec/03method}
\input{sec/04experiments}
\input{sec/05conclusion}

{\small
  \bibliographystyle{plainnat}
  \bibliography{main}
}

\end{document}

%% file: preamble.tex
\usepackage{colortbl}         

\usepackage{booktabs, multirow}
\usepackage{tabularray}
\usepackage{bbding}
\usepackage[font=small]{caption}
\usepackage[export]{adjustbox}  

\usepackage{fontawesome}

\renewcommand{\eg}{\textit{e.g.}}

\usepackage[capitalize]{cleveref}
\crefname{section}{Sec.}{Secs.}
\Crefname{section}{Section}{Sections}
\Crefname{table}{Table}{Tables}
\crefname{table}{Tab.}{Tabs.}
\Crefname{figure}{Figure}{Figures}
\crefname{figure}{Fig.}{Figs.}

%% file: sec/00abstract.tex
\begin{abstract}

We present a method for modeling articulated objects from sparse images with unknown camera poses. Existing approaches require dense multi-view observations and ground-truth camera poses shared across articulation states, whereas our approach operates with as few as four unposed views per articulation state, without requiring ground-truth poses or shared cross-state calibration. The key challenge is that sparse-view reconstruction can recover each articulation state independently, but does not provide the cross-state correspondences required for part segmentation and kinematic reasoning. We address this by optimizing a dense deformation field that aligns independently reconstructed states and establishes dense correspondences across them. An iterative robust estimation procedure then separates static and moving parts and recovers their kinematic parameters. Finally, we refine geometry, appearance, and motion using self-supervised losses that enforce cross-view and cross-state consistency. Experiments on standard benchmarks and real-world examples show that our method produces accurate articulated object representations under substantially weaker input assumptions than prior work.

\end{abstract}

%% file: sec/01introduction.tex
\section{Introduction}

Articulated object modeling aims to reconstruct objects composed of multiple rigid parts connected by joints, such as doors, drawers, and boxes, and to recover their geometry, part structure, and motion parameters. 
This capability is essential for robotic systems that must reason about and interact with everyday objects, yet existing methods typically require dense multi-view observations across articulation states, making them difficult to deploy in practical settings.
In this work, we study a substantially more practical setting: articulated object modeling from sparse, unposed observations.

Early approaches~\cite{qian2022understanding, jiang2022opd, sun2023opdmulti} rely on rich supervision such as 3D object shapes, articulation parameters, or part segmentations. 
While effective, this approach depends on high-quality 3D scans and labor-intensive manual annotations, which are both costly and difficult to scale.
Recent methods reduce supervision by learning from unlabeled multiple views; rather than relying on manual labels, these approaches infer object structure and kinematics from observed motion \cite{liu2023paris,deng2024articulate,guo2025articulatedgs}. 
For instance, PARIS~\cite{liu2023paris} and ArticulateYourNeRF~\cite{deng2024articulate} follow an analysis-by-synthesis approach, learning object representations from pairs of multi-view images, using Neural Radiance Fields (NeRF)~\cite{mildenhall2021nerf}.
These models jointly recover 3D geometry and appearance while separating moving parts from static ones. 
ArticulatedGS~\cite{guo2025articulatedgs} further improves efficiency and accuracy by replacing NeRF with Gaussian Splatting (GS)~\cite{kerbl20233d}.

Despite these advances, existing state-of-the-art methods still rely on \textit{dense, paired observations}, requiring around 100 paired views of the same object across two articulation states. 
Such dense capture is rarely practical.
Moreover, these methods assume that all images are captured in a \textbf{shared, pre-calibrated coordinate system}, i.e., that the object’s static base is perfectly aligned across articulation states via known camera poses. 
In practice, this level of calibration is costly and rarely achievable, severely limiting deployment.
In contrast, our setting requires only four unposed images per articulation state, substantially reducing capture requirements while removing the need for shared camera calibration across states.


\begin{figure*}[t]
    \centering
    \includegraphics[width=0.95\textwidth]{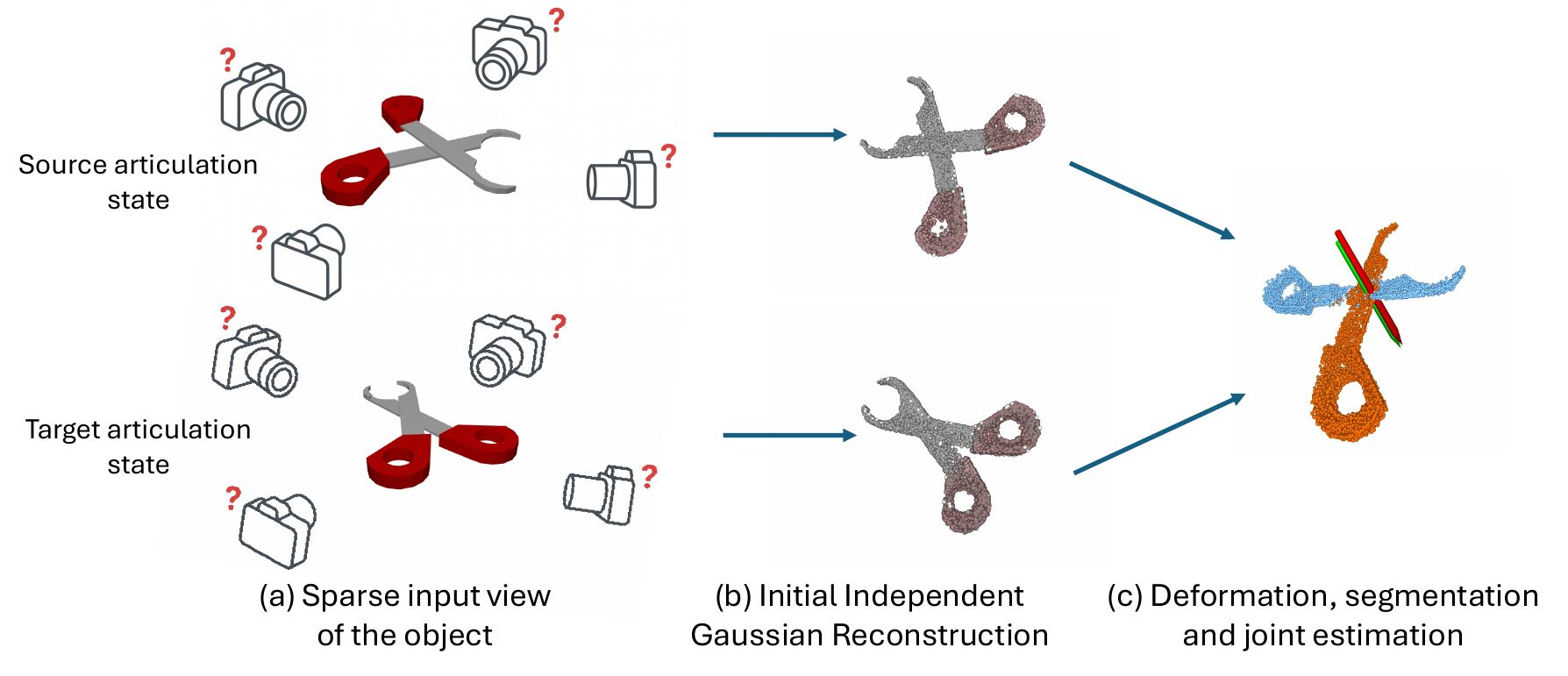}
    \caption{
    Given a few unposed views of an articulated object across different articulation states, \textbf{\nameshort} reconstructs its 3D geometry and appearance while estimating part segmentation and the joint axes of moving parts in 3D. The resulting articulated model can be rendered from novel viewpoints and under novel articulation configurations. }
    \vspace{-1em}
\end{figure*}

We study articulated object reconstruction and modeling from \textit{sparse, unposed views}: a more realistic setting than dense calibrated capture, but also a substantially more challenging one. 
Standard 3D reconstruction methods, including structure-from-motion (SfM)~\cite{schoenberger2016sfm, schoenberger2016vote} and recent neural reconstruction methods~\cite{wang2025vggt, wang2024dust3r}, typically assume static scenes and therefore cannot directly handle object articulation. One possible workaround is to reconstruct each articulation state independently. However, this produces reconstructions in separate coordinate frames, creating a severe misalignment problem. As a result, reasoning about articulation requires first establishing reliable 3D-to-3D correspondences between unaligned reconstructions.

A natural approach is to match sparse keypoints across states. However, sparse keypoints are unreliable under large articulations and on textureless surfaces, and they provide insufficient coverage for dense part segmentation. They are therefore ill-suited to the sparse, unposed setting we target.

We introduce \textit{\nameshort}, a method that converts independently reconstructed articulation states into a coherent articulated object model. Given sparse, unposed views of each articulation state, we first obtain an independent Gaussian Splatting (GS) reconstruction using an off-the-shelf sparse-view reconstruction method~\cite{xu2025freesplatter, fujimura2025ufv}. These per-state reconstructions provide coarse geometry, but they are not assumed to share camera poses, object coordinates, or part correspondences across states.
Note that our method assumes that the sparse-view reconstruction provides sufficient coarse geometry for downstream alignment; we analyze the effect of reconstruction quality in our experiments in ~\cref{para:initialization_robustness}.
Rather than proposing a new sparse-view reconstruction method, \textit{\nameshort} addresses the cross-state reasoning problem that remains after each articulation state has been reconstructed independently.

\textit{\nameshort} estimates dense cross-state correspondences by optimizing a deformation field that warps the source reconstruction into alignment with the target state, using both 3D geometric consistency and 2D photometric rendering losses. Once the source has been brought into the target frame, we establish dense correspondences by matching the deformed source model to the target model with 1-Nearest-Neighbor (1-NN). Finally, we separate rigid parts using an iterative robust estimation procedure: we repeatedly cluster inlier correspondences with a pose estimator~\cite{yang2020teaser}, thereby isolating static and moving components and recovering their kinematic parameters. In summary, our main contributions are three fold:
\begin{enumerate}
    \item \textbf{Sparse, unposed articulated modeling:} To our knowledge, we are the first to formulate articulated object modeling from independently reconstructed sparse, unposed articulation states. We identify dense 3D-to-3D correspondence between unaligned per-state reconstructions as the central missing step.
    \item \textbf{Dense correspondence from independent reconstructions:} We propose a test-time deformation-based correspondence method that operates on independently reconstructed GS models. Unlike sparse-view reconstruction methods, which can reconstruct each state separately, our method estimates the cross-state dense correspondences needed for part segmentation and kinematic recovery.
    \item \textbf{Complete articulated modeling pipeline:} We present a complete pipeline that combines off-the-shelf sparse-view reconstruction with our dense correspondence and robust part-motion estimation modules to recover object geometry, part segmentation, and kinematic parameters from sparse, unposed observations.
\end{enumerate}

Together, these contributions show that sparse, unposed articulated object modeling becomes feasible when the cross-state correspondence problem is addressed directly through object-specific dense deformation.

%% file: sec/02relatedwork.tex
\section{Related Work}

\paragraph{Pose-free 3D reconstruction} 
Classical Structure from Motion (SfM)~\cite{agarwal2011building,liu2024robust,schoenberger2016sfm,schoenberger2016mvs} jointly estimates camera poses and scene geometry via image registration, triangulation, and bundle adjustment. 
Recent learning-based reconstruction methods~\cite{yang2025fast3r,wang2024dust3r,leroy2024grounding,wang2025vggt} achieve high-quality reconstructions without explicit feature matching.
DUST3R~\cite{wang2024dust3r} uses transformers to predict dense point maps and poses from image pairs, though bundle adjustment is still required. MASt3R~\cite{leroy2024grounding} improves scalability with coarse-to-fine matching for high-resolution inputs, while VGGT~\cite{wang2025vggt} extends this paradigm with multi-task learning to jointly predict point maps, poses, depth, and tracks.
These methods provide powerful initialization for sparse, unposed reconstruction, but they assume a static scene and therefore do not recover cross-state correspondences, part decomposition, or joint parameters for articulated objects.

\paragraph{Sparse View Gaussian Splats} 
Neural rendering~\cite{mildenhall2021nerf,kerbl20233d,huang20242d,wang2021neus,muller2022instant} enables photorealistic novel view synthesis by rendering from learned 3D representations.
More recently, 3D GS~\cite{kerbl20233d} has emerged as particularly compelling, offering efficient, high-quality representations~\cite{fan2024lightgaussian,cheng2024gaussianpro,lu2023scaffold,lee2024compact,yan2024multi}. 
Standard GS requires dense multi-view images with accurate poses.
To relax this, some works use depth priors~\cite{li2024dngaussian,zhu2024fsgs,chung2024depth} or diffusion guidance~\cite{liu2023zero,kong2025generative,long2024wonder3d}, while others directly predict Gaussian parameters with feed-forward networks~\cite{ye2024no,xu2025freesplatter,szymanowicz2024splatter,tang2024lgm,charatan2024pixelsplat,chen2024mvsplat}. 
Splatter Image~\cite{szymanowicz2024splatter} and LGM~\cite{tang2024lgm} handle single images, while MVSPlat~\cite{chen2024mvsplat} and PixelSplat~\cite{charatan2024pixelsplat} target sparse multi-view with known poses. FreeSplatter~\cite{xu2025freesplatter} and NoPoSplat~\cite{ye2024no} remove pose supervision by jointly predicting poses and Gaussians, but remain limited to static scenes without articulated motion. UFV-splatter~\cite{fujimura2025ufv} improves robustness to image viewpoints by finetuning FreeSplatter with images from unfavorable views, which enhances reconstruction and camera pose estimation.
These methods are complementary to our work: they provide per-state geometry from sparse, unposed images, whereas our goal is to align independently reconstructed articulation states and recover part-level motion.

\paragraph{Articulated Object Reconstruction}
Reconstructing articulated objects is challenging due to coupled geometry and kinematics. Some methods focus on structure or motion alone~\cite{jiang2022ditto,le2024articulate,che2024op,fu2024capt,liu2023building,kawana2023detection}, while others use generative part-level graphs~\cite{liu2024cage,lei2023nap}. More recent works jointly model geometry and appearance through optimization~\cite{mu2021sdf,tseng2022cla,wei2022self,liu2023paris,deng2024articulate,weng2024neural,liu2025building,guo2025articulatedgs}. 
DITTO~\cite{jiang2022ditto} is an early work on modeling two-part articulated objects from point-cloud input, but it assumes observations in an aligned coordinate frame and uses a pre-trained feed-forward model.
PARIS~\cite{liu2023paris} reconstructs parts and motions from dense multi-view scans with known poses. ArticulateYourNeRF~\cite{deng2024articulate} improves robustness through iterative segmentation and motion learning. ArticulatedGS~\cite{guo2025articulatedgs}, ArtGS~\cite{liu2025building}, and DTA~\cite{weng2024neural} further improve reconstruction quality using GS representations or depth priors.

In contrast to optimization-based approaches, data-driven methods learn articulation priors from large-scale datasets, enabling fast and scalable prediction of articulated structure without per-instance optimization ~\cite{chen2024urdformer, liu2024singapo,le2025articulate,li2025art}.
Articulate-Anything~\cite{le2025articulate} employs a vision-language model as an actor-critic policy that iteratively synthesizes URDF code from single images, and retrieves part mesh from a fix mesh set to achieve generalizable articulated object modeling across diverse categories.
URDF-Anything+~\cite{wu2026urdf} further advances this paradigm with an autoregressive diffusion transformer (DiT) that sequentially generates parts and joints in a unified framework.

Existing methods typically rely on dense, calibrated multi-view inputs or struggle to generalize beyond limited training categories. In contrast, our approach leverages pre-trained static 3D reconstruction models as a powerful prior, enabling high-fidelity articulated reconstruction without 3D supervision. Consequently, our method generalizes across diverse categories while operating directly from sparse views without requiring ground-truth camera extrinsics or shared cross-state calibration.




\paragraph{3D-3D Correspondence and Matching}
Finding 3D-to-3D correspondences is a fundamental problem. 
Many approaches rely on sparse keypoints, ranging from classical descriptors such as FPFH~\cite{rusu2009fpfh} to modern learned detectors~\cite{li2019usip,zohaib2023sc3k}.
Recent GS registration methods such as GaussReg~\cite{chang2024gaussreg} establish coarse 3D keypoint matches between Gaussian representations. However, coarse keypoint matching can be insufficient for accurate dense registration, especially when articulated motion requires part-level correspondences.
A more robust alternative is to learn dense correspondences via deformation fields~\cite{groueix20183d,deng2021deformed,cao2023self,le2024integrating,chen2025dv}. 
Many of these approaches are designed for meshes or point clouds, and often rely on pre-trained deformation models rather than test-time optimization over 3D GS representations.
In contrast, our method introduces a test-time, optimization-based framework to learn a dense deformation field specifically for GS, allowing us to solve the articulated motion problem. We then use a robust iterative algorithm to segment these dense correspondences into distinct rigid parts and their kinematics.

%% file: sec/03method.tex
\section{Method}
\label{sec:method}

We address articulated object reconstruction from sparse images with unknown camera poses. 
We consider objects composed of rigid parts connected by 1-DoF revolute or prismatic joints. 
Given two image sets, $\mathcal{I}^s$ and $\mathcal{I}^t$, each containing sparse views of the same object in a fixed articulation state, our goal is to reconstruct an articulated 3D model with geometry, appearance, part segmentation, and joint parameters. 
The articulation state differs across the two sets, while camera poses are unknown.

\begin{figure*}[t]
    \centering
    \includegraphics[width=0.95\linewidth]{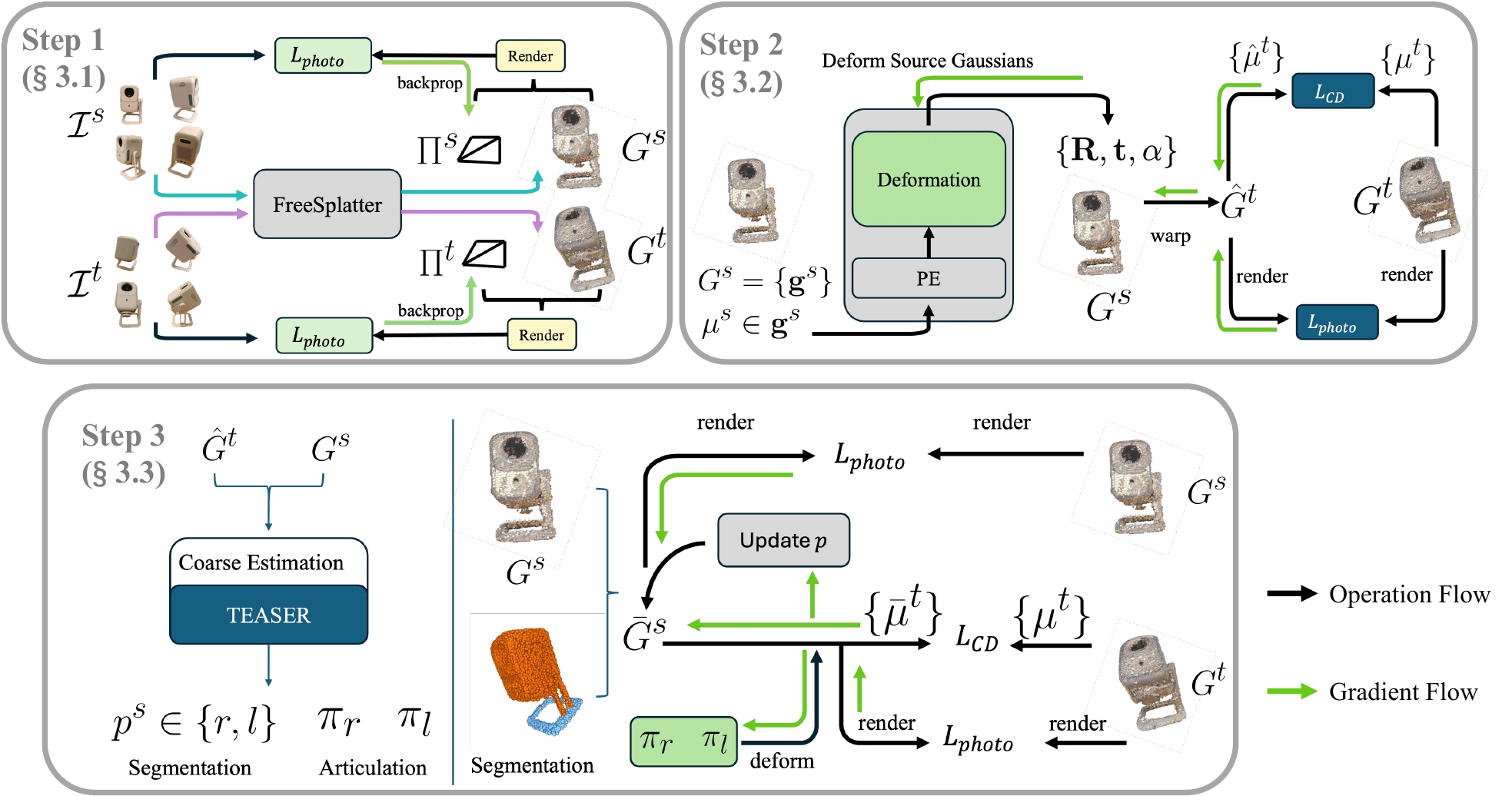}
    \vspace{1em}
    \caption{Overview of our method pipeline. Given two sets of K-view images (K=4) of an articulated object in different states, our approach reconstructs high-fidelity 3D geometry and estimates articulation parameters through a three-stage process: (1) initialization of Gaussian splats and camera parameters (top left), (2) Gaussian alignment via deformation fields (top right), and (3) joint optimization of geometry and articulation parameters (bottom).}
    \label{fig:method_pipeline}
\end{figure*}

Our pipeline has three stages, as shown in \cref{fig:method_pipeline}. First, we independently reconstruct a 3D GS model and estimate camera parameters for each image set. Second, we optimize a dense deformation field that aligns the independently reconstructed states and yields dense cross-state correspondences. Third, we use these correspondences to initialize part segmentation and rigid transformations, followed by staged joint optimization of geometry, part assignments, appearance, and articulation parameters. 

\subsection{Initialization of 3D Gaussian Splats}
\label{sec:initgs}
Let $\mathcal{I}^s = \{I^s_k\}_{k=1}^{K}$ and $\mathcal{I}^t = \{I^t_k\}_{k=1}^{K}$ denote the source and target image sets, each containing $K$ (\eg, 4) unposed views.
We first construct initial 3D Gaussian splats $G^s$ and $G^t$ independently. 
Each set $G = \{\mathbf{g}_i\}_{i=1}^{N}$ consists of $N$ Gaussian points, where each point $\mathbf{g}= (\bm{\mu}, \mathbf{q}, \mathbf{s}, o, \mathbf{c})$ is parameterized by its mean $\bm{\mu}$, rotation $\mathbf{q}$, scale $\mathbf{s}$, opacity $o$, and color $\mathbf{c}$.

Since FreeSplatter is designed for static scenes, we apply it independently to the source set $\mathcal{I}^s$ and the target set $\mathcal{I}^t$. This process yields our two initial 3D Gaussian Splatting models, $G^s$ and $G^t$, each in its own arbitrary coordinate system. This step also provides initial camera parameters (both extrinsics and intrinsics) for each set. 
Because all views are assumed to be captured with the same camera model, we average the estimated intrinsics from the two independently reconstructed sets and keep the resulting intrinsic matrix fixed during subsequent optimization.
However, the initial Gaussians and extrinsic estimates are not perfectly pixel-aligned. To refine this, we freeze the GS and optimize only the extrinsic parameters ($\Pi^s, \Pi^t$) for each set by minimizing a photometric rendering loss. This refinement provides a better initialization for the subsequent steps.


\subsection{Gaussian Correspondences via Deformation Fields}
\label{sec:deform}
A central challenge is aligning independently reconstructed articulation states that live in different coordinate systems.
A natural approach is to estimate this alignment from sparse 3D keypoint matches~\cite{li2019usip,zohaib2023sc3k}. However, sparse keypoints are unreliable for dense analysis and struggle on textureless surfaces.
A more robust alternative is to find a dense alignment. A naive strategy would be to optimize the mean and rotation of each Gaussian point independently, but this breaks local rigidity, since it allows unconstrained updates of individual parameters, whereas true object motion should remain rigid within each part.

To address this, we learn a per-scene deformation field, parameterized by a lightweight feed-forward network, $\mathcal{F}_{\text{deform}}$. This network is optimized from scratch for each object and predicts rigid transformations for Gaussian points in $G^s$ using their positional encodings, producing a deformed target representation $\hat{G}^t$. 
By restricting the positional encoding to low frequencies, we bias the deformation field toward smooth, locally coherent transformations, which discourages unconstrained point-wise motion.
Specifically, $\mathcal{F}_{\text{deform}}$ maps the positional encoding of Gaussian means to a rigid transformation consisting of a rotation, translation, and scaling factor:
\begin{equation}
    \{\mathbf{R}, \mathbf{t}, \alpha\} = \mathcal{F}_{\text{deform}}(\text{PE}(\bm{\mu}))
    \label{eq:deformation_network}
\end{equation}
where $\mathbf{R} \in \text{SO}(3)$ is parameterized by Euler angles $\in \mathbb{R}^3$ and converted to rotation matrices, $\mathbf{t} \in \mathbb{R}^3$ represents the translation vector, $\alpha \in \mathbb{R}$ is a scaling factor, and $\text{PE}(\bm{\mu})$ is the positional encoding of the Gaussian means. In practice, we compute average scaling factor $\bar{\alpha} = \frac{1}{N}\sum_1^N \alpha_i$ to serve as global scale.
This transformation is then applied to the source Gaussian points to obtain the deformed target representation.

The means of the source Gaussian points $\bm{\mu} \in \mathbb{R}^3$ are encoded as:
\begin{equation}
    \text{PE}(\bm{\mu}) = [\sin(2^{k_0} \pi \bm{\mu}), \cos(2^{k_0} \pi \bm{\mu})]
    \label{eq:pe}
\end{equation}
where $k_0$ represents the base frequency parameter. The parameter $k_0$ directly influences the rigidity of the deformation field: smaller values result in lower-frequency encodings that naturally constrain the deformation field to be more rigid, effectively guiding the network's capacity to learn low-frequency non-rigid transformations. 

\paragraph{Deformation Network Training}
To optimize the parameters of $\mathcal{F}_{\text{deform}}$, we minimize the Chamfer distance $\mathcal{L}_{\text{CD}}$ between the means of the deformed source Gaussian $\hat{G}^t$ and the target one ${G}^t$, as well as $\ell_1$ photometric loss $\mathcal{L}_{\text{photo}}$ between their rendered images $\mathcal{R}(\hat{G}^t), \mathcal{R}(G^t)$ respectively:
\begin{equation}
    \min_{\mathcal{F}_{\text{deform}}} \mathcal{L}_{\text{CD}}(\hat{G}^t, G^t) + 
    \mathcal{L}_{\text{photo}}(\mathcal{R}(\hat{G}^t), \mathcal{R}(G^t))
    \label{eq:deformation_training}
\end{equation} where $\hat{G}^t = \{\hat{\mathbf{g}}_i^t\}_{i=1}^{N^s}$ represents the deformed source Gaussians, with each point $\hat{\mathbf{g}}^t$ obtained by applying the transformation predicted by $\pi = \mathcal{F}_{\text{deform}}(\bm{\mu})$ to the corresponding source point $\mathbf{g}^s \in G^s$, where $\pi$ is a similarity transformation containing rotation $\mathbf{R}$, translation $\mathbf{t}$, and scaling $s$. Specifically, to obtain the deformed target Gaussian points $\hat{\mathbf{g}}^t = (\hat{\bm{\mu}}^t, \hat{\mathbf{q}}^t, \mathbf{s}^s, o^s, \mathbf{c}^s)$, we first compute the transformation for each point $\mathbf{g}^s = (\bm{\mu}^s, \mathbf{q}^s, \mathbf{s}^s, o^s, \mathbf{c}^s)$ using \cref{eq:deformation_network},
and apply the transformation to the means and rotation of the source Gaussian points. With $\hat{\bm{\mu}}^t = \alpha \mathbf{R} \bm{\mu}^s + \mathbf{t}$ and $\hat{\mathbf{q}}^t = \mathbf{q}^s \otimes \mathbf{q}(\mathbf{R})$, we can obtain the deformed target Gaussian points $\hat{\mathbf{g}}^t = (\hat{\bm{\mu}}^t, \hat{\mathbf{q}}^t, \mathbf{s}^s, o^s, \mathbf{c}^s)$, where $\otimes$ denotes quaternion multiplication and $\mathbf{q}(\mathbf{R})$ converts the rotation matrix $\mathbf{R}$ to its quaternion representation.
For simplicity, we denote this process as $\hat{\mathbf{g}} = \pi \circ \mathbf{g}$.

After optimizing $\mathcal{F}_{\text{deform}}$, we establish dense correspondences by nearest-neighbor matching between the deformed source means and the target means:
\begin{equation}
    j(i) = \arg\min_j \left\| \hat{\bm{\mu}}^t_i - \bm{\mu}^t_j \right\|_2 .
\end{equation}
This gives a correspondence set
$\mathcal{C} = \{(\bm{\mu}^s_i, \bm{\mu}^t_{j(i)})\}_{i=1}^{N^s}$,
which is used for robust part-motion initialization.

\subsection{Joint Optimization}
\label{sec:jointopt}

To obtain an initial segmentation and transformations, we leverage TEASER++~\cite{yang2020teaser}, a robust pose estimation algorithm. For clarity, we first describe the common two-part (root/leaf) scenario (see \cref{sec:multi_part} for multi-part). 
We first run TEASER++ on the full correspondence set $\mathcal{C}$ to estimate the dominant rigid transform. Its inliers are assigned to the root part $r$. We then run TEASER++ on the remaining correspondences to estimate the moving-part transform, assigning its inliers to the leaf part $l$.
This yields initial transformations $\pi_r, \pi_l$ and part labels $p^s$.
Because both the initial geometry from FreeSplatter and the part-pose estimates from TEASER++ are coarse approximations, we refine them through staged joint optimization. The staged design reduces sensitivity to poor initialization: Phase 1 stabilizes large-scale rigid motions, Phase 2 refines geometry and segmentation, and Phase 3 improves pose and appearance.


We define our full learnable model and its parameters. We initialize the learnable Gaussian splats $\bar{G}^s = \{\bar{\mathbf{g}}^s_i\}_{i=1}^{N^s}$, where each $\bar{\mathbf{g}}^s = (\bm{\mu}^s, \mathbf{q}^s, \mathbf{s}^s, o^s, \mathbf{c}^s, p^s)$ is augmented with its initial part label $p^s$. We also treat the rigid transformations $\pi_r$ and $\pi_l$ as learnable parameters. To compensate for non-rigidity and initial estimation errors, we introduce a learnable, per-Gaussian displacement term $\delta\bm{\mu}$.
To simplify the notation, we define an auxiliary function $f(\bar{\mathbf{g}}^s_i, \delta\bm{\mu}_i, \pi)$. This function first selects the appropriate rigid transformation $\pi_{p^s}$ from the set $\pi$ (using the part label $p^s$ contained within $\bar{\mathbf{g}}^s_i$), applies it, and then adds the displacement $\delta\bm{\mu}_i$ to the transformed mean. This yields the final deformed Gaussian splats $\bar{G}^t = \{ \bar{\mathbf{g}}^t_i | \bar{\mathbf{g}}^t_i = f(\bar{\mathbf{g}}^s_i, \delta\bm{\mu}_i, \pi), \bar{\mathbf{g}}^s_i \in \bar{G}^s \}$.

\noindent\textbf{Phase 1: Rigid Pose Refinement} The goal of this phase is to stabilize the coarse poses from TEASER++. We apply two constraints: the source Gaussian geometry $\bar{G}^s$ is frozen, and the non-rigid displacement $\delta\bm{\mu}$ is set to zero. Using the fixed initial segmentation $p^s$, we only optimize the rigid transformations $\pi = \{\pi_r, \pi_l\}$ by minimizing a joint loss:
\begin{equation}
\begin{aligned}
    \mathcal{L}_{\text{joint}} ={}& \mathcal{L}_{\text{CD}}(\bar{G}^t, G^t) + \mathcal{L}_{\text{photo}}(\mathcal{R}(\bar{G}^t), \mathcal{R}(G^t)) \\
    &+ \mathcal{L}_{\text{photo}}(\mathcal{R}(\bar{G}^s), \mathcal{R}(G^s))
\end{aligned}
\end{equation} 
The optimization objective for this phase is:
\begin{equation}
    (\pi_r^*, \pi_l^*) = \arg\min_{\pi_r, \pi_l} \mathcal{L}_{\text{joint}}
\end{equation}

\noindent\textbf{Phase 2: Geometry and Occlusion Refinement} 
With the rigid poses stabilized in Phase 1, we now address the inaccuracies in the underlying geometry and the initial part segmentation. In this phase, we unfreeze the parameters in $\bar{G}^s$ and activate the learnable displacement term $\delta\bm{\mu}$. 
This per-Gaussian term compensates for residual non-rigid errors, which often arise from incorrect part assignments or imperfect initialization.
We encourage the displacement to be sparse by penalizing it with $\mathcal{L}_{\text{rigid}} = \sum_{i=1}^N \left\| \delta\bm{\mu}_i \right\|_1$.

Crucially, this displacement term also serves as a heuristic to correct mis-segmented points. As suggested in \cite{deng2024articulate}, we periodically update segmentation labels $p^s$. We assume points with a large $\delta\bm{\mu}$ are mis-assigned, and we re-assign their label $p$ based on the smallest rigid alignment error:
\begin{equation}
    p_i^s = \arg\min_{p \in \{r,l\}}\left\|\pi_p \circ \bar{\bm{\mu}}^s_i-\bm{\mu}^t_{j(i)}\right\|_2,
    \label{eq:segmentation_update}
\end{equation} where $j(i)$ denotes the target correspondence defined in \cref{sec:deform}.

The full optimization objective minimizes the joint loss plus the rigidity penalty, updating all parameters (geometry, transformations, and displacements) simultaneously:
\begin{equation}
    \min_{\{\delta\bm{\mu}\}, \bar{G}^s, \pi_r, \pi_l} \mathcal{L}_{\text{joint}} + \mathcal{L}_{\text{rigid}}
\end{equation}

\noindent\textbf{Phase 3: Pose and Appearance Refinement} The previous phases refined the geometry and produced an as-rigid-as-possible part segmentation. In this final phase, we lock this geometry (freezing $\bar{G}^s$ and setting $\delta\bm{\mu} = \mathbf{0}$) to focus exclusively on finetuning the per-part pose estimation ($\pi_r, \pi_l$) and visual quality (colors $\mathbf{c}$). This is supervised purely by the photometric loss against the original input images:
\begin{equation}
     \min_{\{\pi_r, \pi_l, \mathbf{c}\}} \mathcal{L}_{\text{photo}}(\mathcal{R}(\bar{G}^s), I^s) + \mathcal{L}_{\text{photo}}(\mathcal{R}(\bar{G}^t), I^t).
\end{equation}
In practice, we use 500 iterations for the first phase, 2500 iterations for the second phase, and 2000 iterations for the final phase.
Finally, we decompose the articulation parameters into axis $\mathbf{a} \in \mathbb{R}^3$, angle $\theta \in \mathbb{R}$, and pivot point $\mathbf{p} \in \mathbb{R}^3$ for revolute joints, 
and an axis $\mathbf{a}$ and translation magnitude $d$ for prismatic joints.

\subsection{Multiple Moving Part Modeling}
\label{sec:multi_part}


\begin{figure}[t]
    \centering
    \includegraphics[width=0.95\columnwidth]{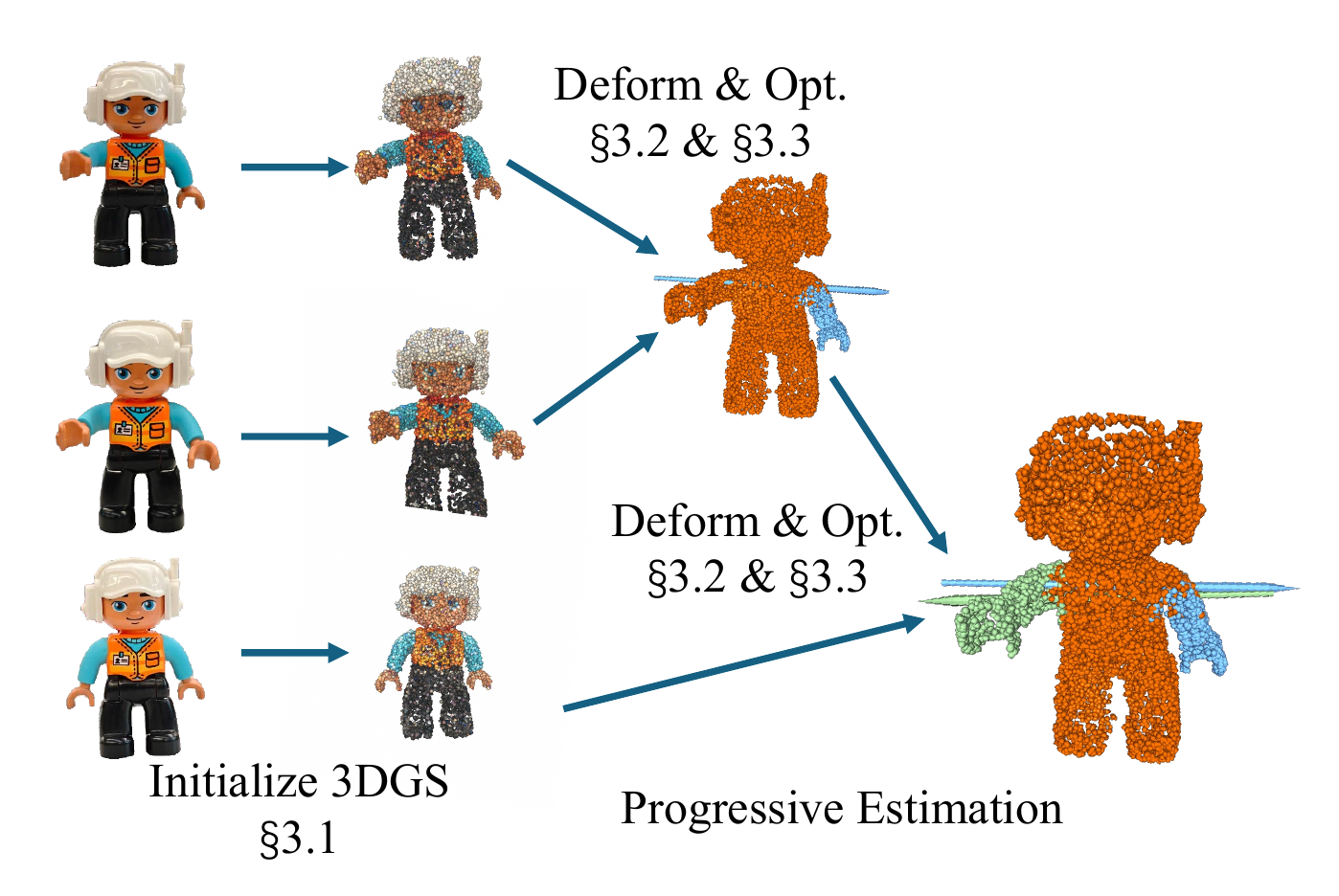}
    \caption{Illustration of the extension to multiple parts. We progressively estimate the segmentation and articulation one part at a time with extra images for new target articulation states. }
    \label{fig:multi_parts}
\end{figure}
For objects with multiple moving parts, we apply the two-state pipeline iteratively. 
Given a common source image set $\mathcal{I}^s$ and target sets $\{\mathcal{I}^{t_m}\}_{m=1}^{M}$, each target set is chosen such that only one additional part moves relative to the source. 
As illustrated in \cref{fig:multi_parts}, for each pair $(\mathcal{I}^s,\mathcal{I}^{t_m})$, we estimate dense correspondences, recover the segmentation and articulation parameters of the moving part, and accumulate the recovered part into the final articulated model. 
This extension therefore assumes that moving parts can be isolated across the available articulation states, which avoids ambiguous correspondence clusters and incorrect part-motion assignment. 


%% file: sec/04experiments.tex
\section{Experiments}

\input{tables/bigresults.tex}
\subsection{Experimental Setup}
 
\paragraph{Dataset and Metrics} Following the evaluation protocol established by previous work~\cite{liu2023paris,deng2024articulate,guo2025articulatedgs}, we evaluate our method on the synthetic 3D PartNet-Mobility dataset~\cite{Mo_2019_CVPR}.
Our evaluation metrics encompass three key aspects: (1) For rendering, we use PSNR (Peak Signal-to-Noise Ratio) and SSIM (Structural Similarity Index) to evaluate the quality of the rendered images for unseen views in both source and target articulation states. (2) For articulation parameter estimation, we use the angular error, position error of the estimated motion axis and additional motion distance to evaluate the accuracy of articulation parameters as in previous work~\cite{deng2024articulate, liu2023paris, guo2025articulatedgs}. More details are provided in the supplementary material. (3) For 3D reconstruction precision, we use the Chamfer Distance (CD) to evaluate the precision of the estimated 3D geometry. 
For PSNR, SSIM and CD, we report the average values of the two articulation states.

\paragraph{Baselines} To our knowledge, no existing work addresses self-supervised articulated object modeling from sparse, unposed images. Our evaluation is therefore twofold: a component-wise analysis to validate our correspondence method and a full-pipeline comparison against state-of-the-art methods. For the component analysis, we fix our pipeline and swap our correspondence module with two alternatives: (1) GaussReg~\cite{chang2024gaussreg}, a pre-trained rigid GS registration method that provides 1024 sparse keypoint correspondences, and (2) FM-Dense~\cite{cao2023self}, a pre-trained non-rigid matching network, to which we feed 20480 downsampled points from our GS to obtain dense correspondences. For the full-pipeline comparison, we evaluate against AGS with the same number of images ($K=4$) and ground truth, aligned camera parameters. Besides, we use the VGGT~\cite{wang2025vggt} reconstructed point clouds to initialize the AGS for better reconstruction. We will make our code and models publicly available.

\subsection{Results}
\paragraph{Quantitative Evaluation}

Results are presented in \cref{tab:quantitative}. 
We first compare our method against AGS. The results show that AGS, even with GT aligned camera poses and VGGT initialization, fails to converge in a sparse 4-view setting. This is because its optimization, designed for dense data, introduces significant noise from sparse inputs, and the limited views are insufficient for its motion estimation to converge. These factors lead to large errors, demonstrating that existing SOTA methods are not robust to this sparse-input scenario.

\begin{figure}[t]
    \centering
    \resizebox{\columnwidth}{!}{%
        \SetTblrInner{rowsep=1pt,colsep=1pt}
        \begin{tblr}{c|c|c|c|c}
        Corr. & Oven & Storage & USB & Laptop \\
        \hline
        GaussReg
        & \includegraphics[valign=c, width=0.1\textwidth]{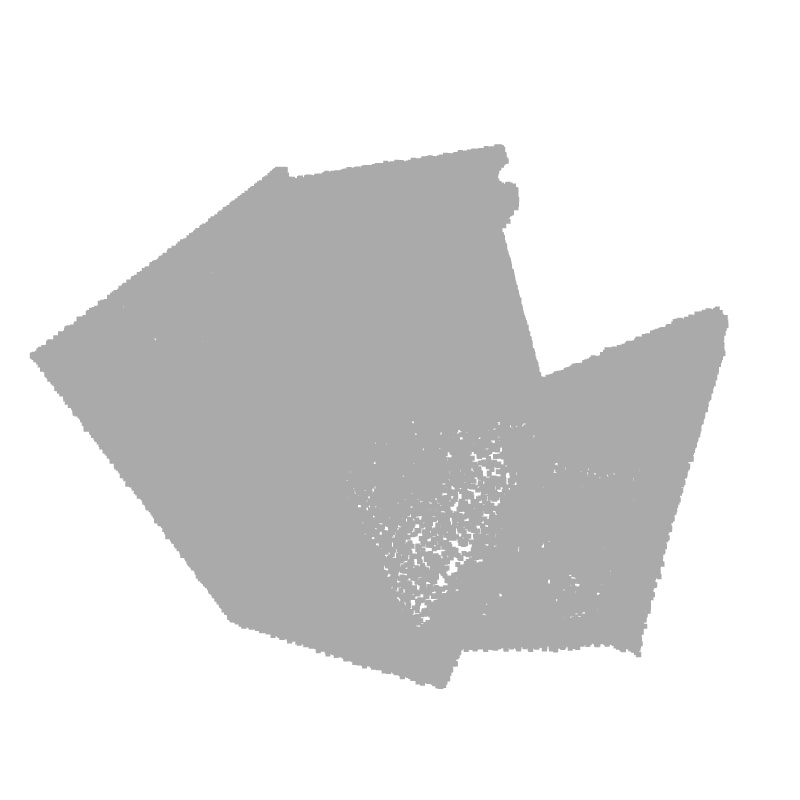}
        & \includegraphics[valign=c, width=0.1\textwidth]{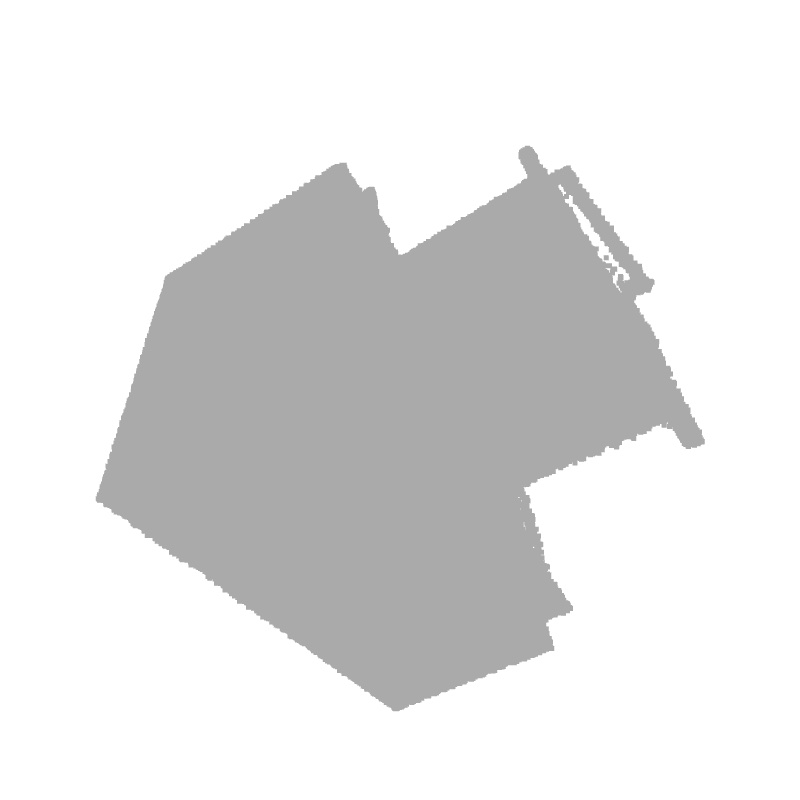}
        & \includegraphics[valign=c, width=0.1\textwidth]{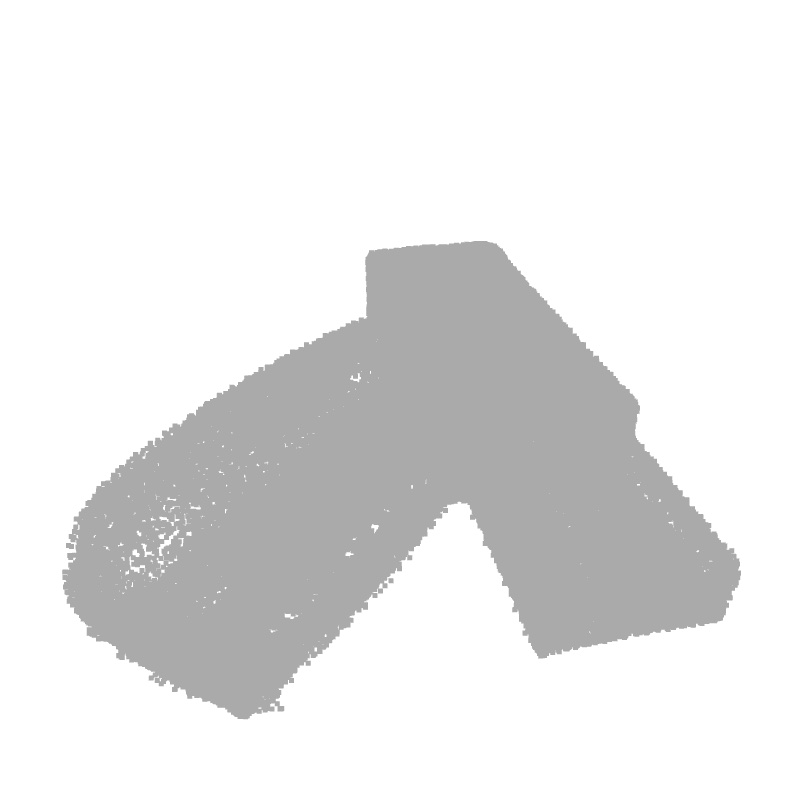}
        &\includegraphics[valign=c, width=0.1\textwidth]{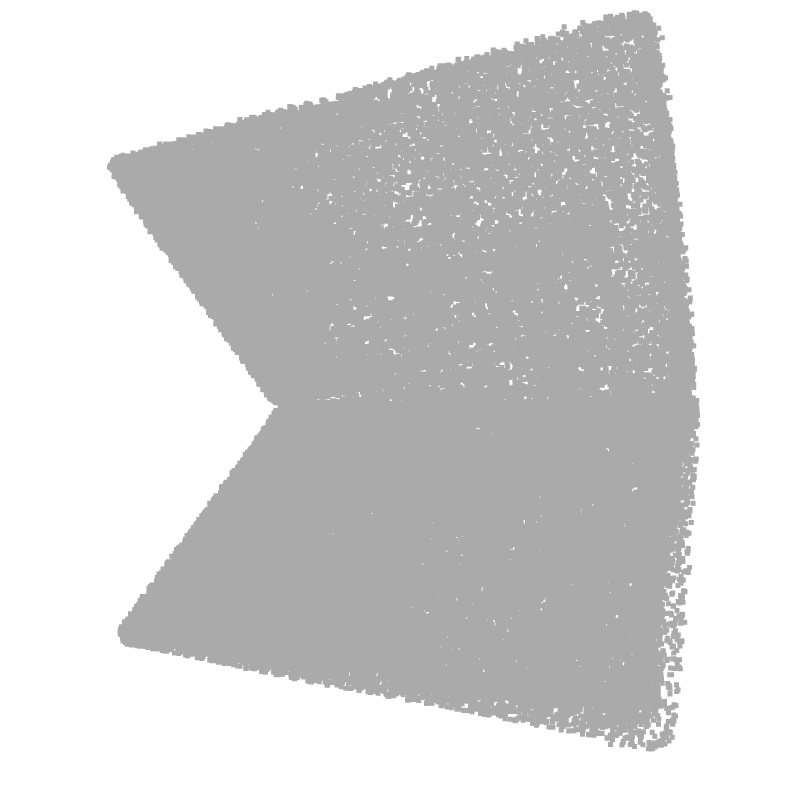}
        \\
        FM-Dense
        & \includegraphics[valign=c, width=0.1\textwidth]{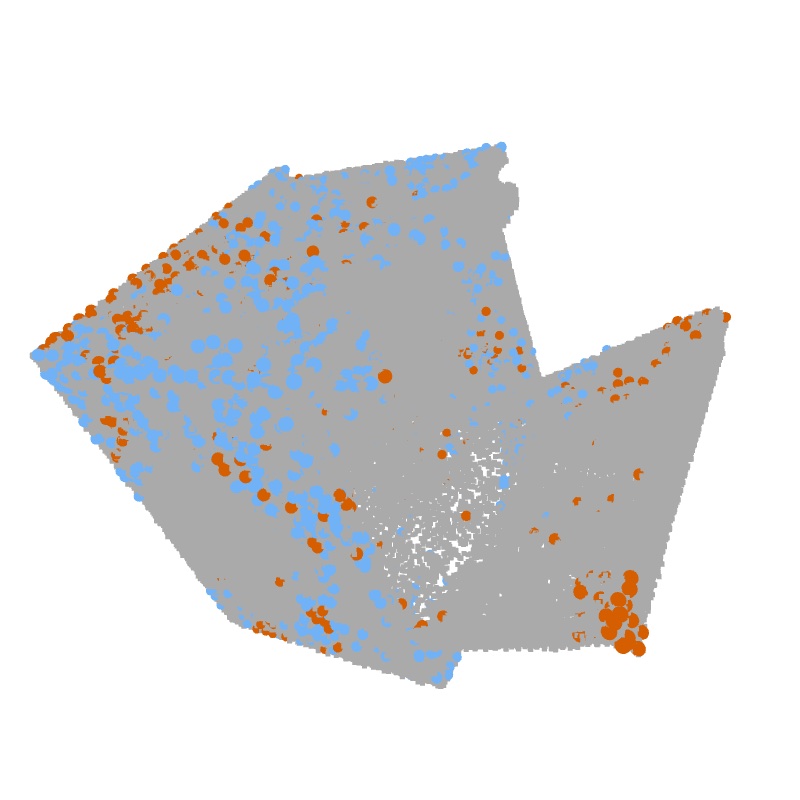}
        & \includegraphics[valign=c, width=0.1\textwidth]{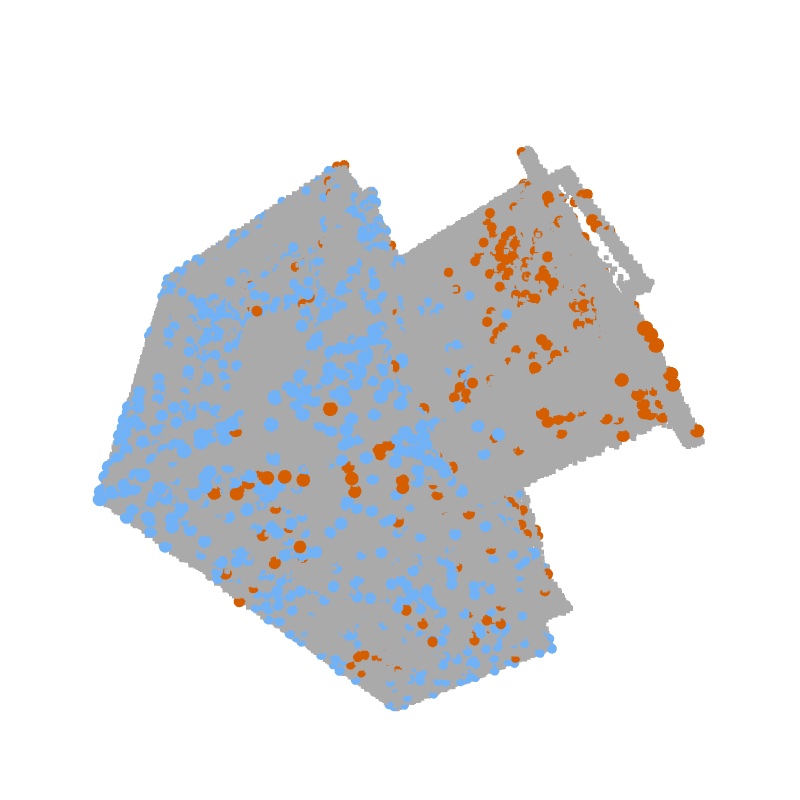}
        & \includegraphics[valign=c, width=0.1\textwidth]{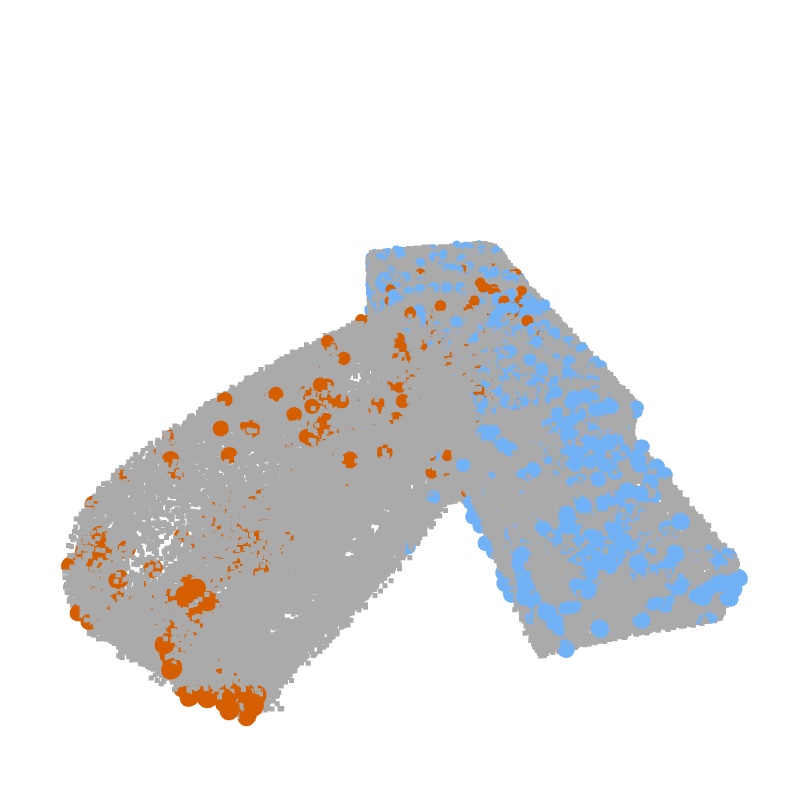}
        &\includegraphics[valign=c, width=0.1\textwidth]{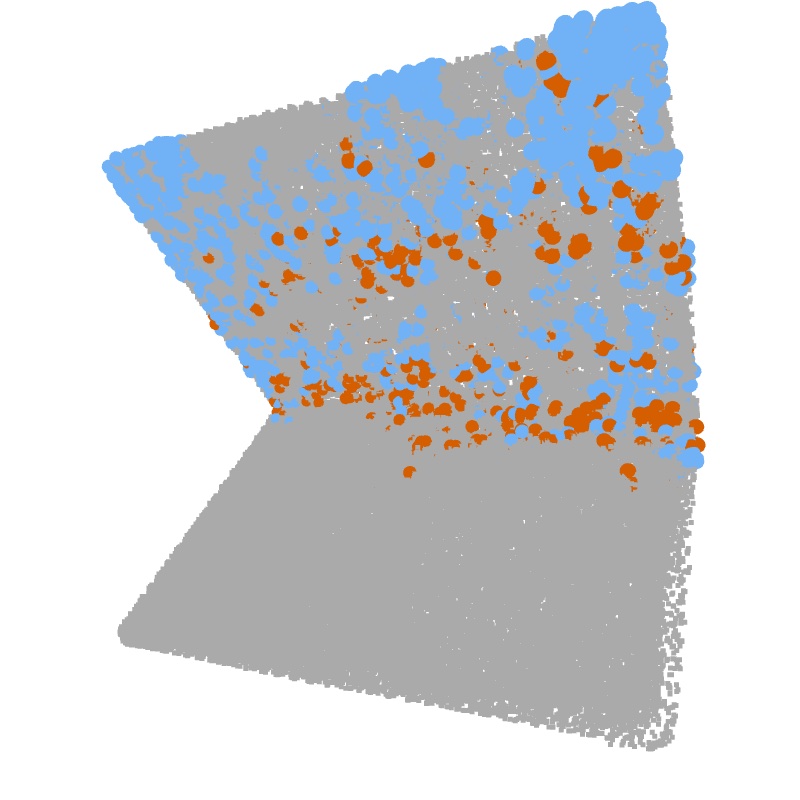}
        \\
        Ours
        & \includegraphics[valign=c, width=0.1\textwidth]{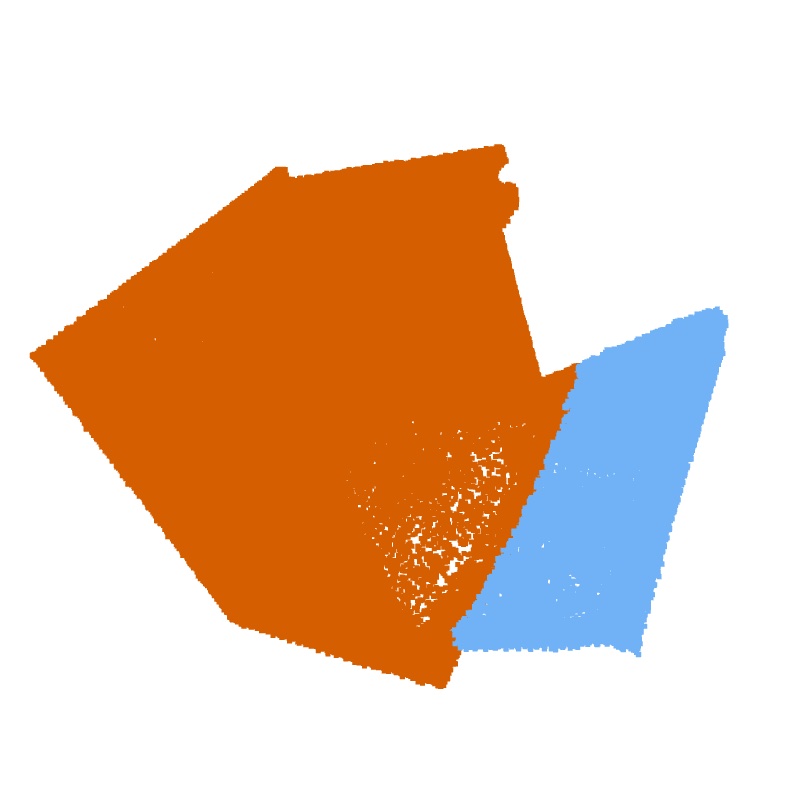}
        & \includegraphics[valign=c, width=0.1\textwidth]{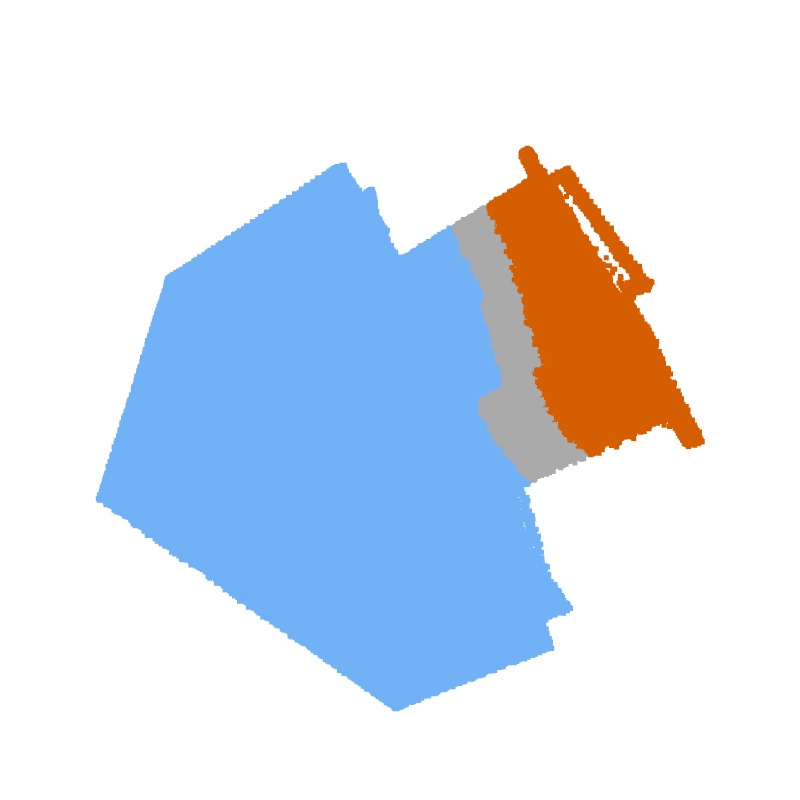}
        & \includegraphics[valign=c, width=0.1\textwidth]{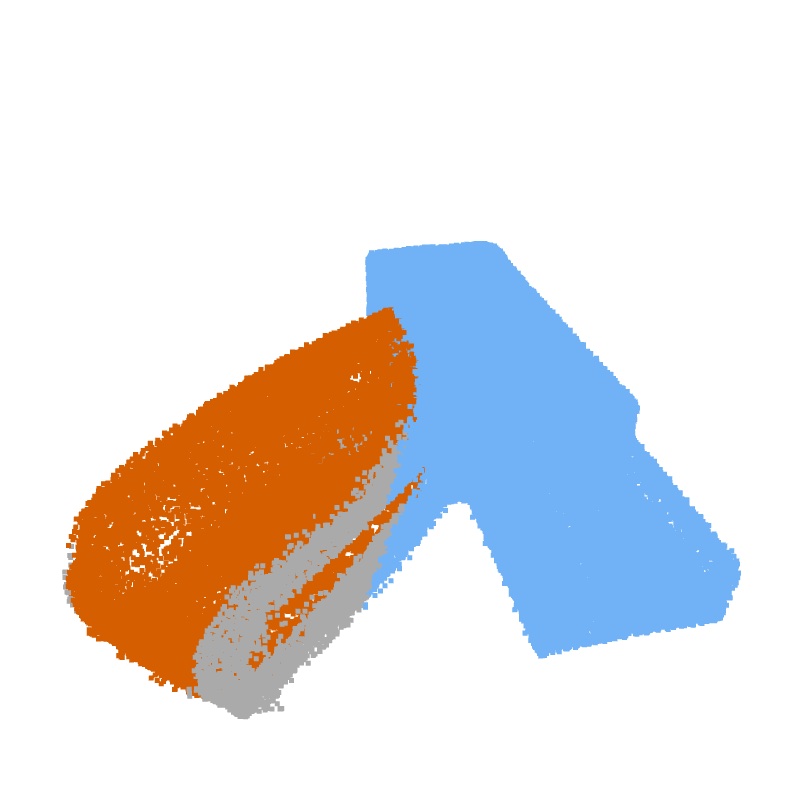}
        &\includegraphics[valign=c, width=0.1\textwidth]{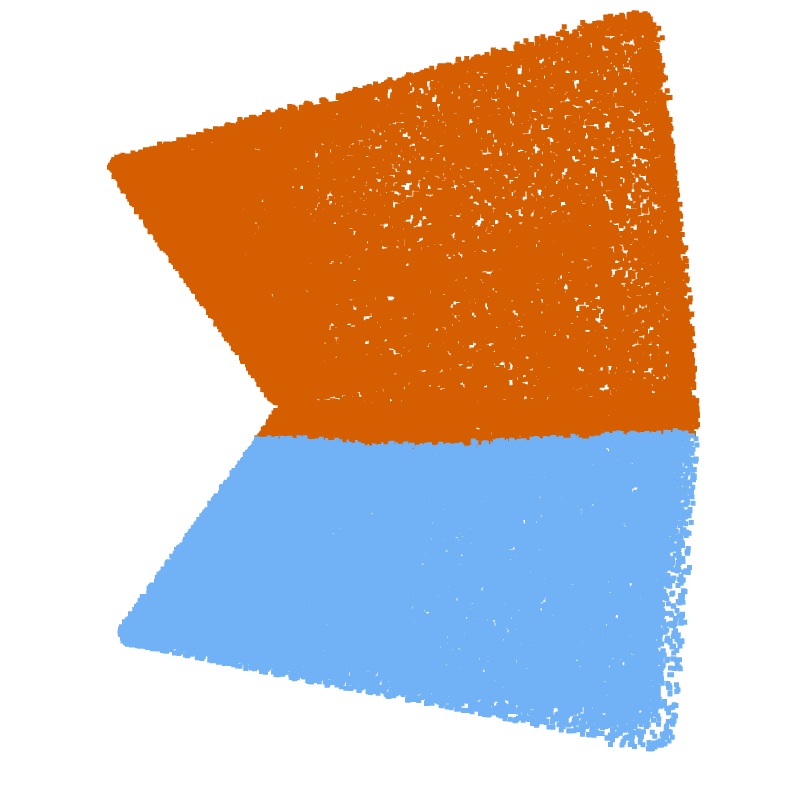}
        
        \end{tblr}
    }
    \caption{\textbf{Qualitative analysis of correspondence methods.} We visualize the part segmentations computed by our TEASER~\cite{yang2020teaser} solver, which is fed correspondences from each method. Inliers for the two main segmented parts are colored blue and orange; outliers and non-correspondences are shown in gray. Better viewed in color and zoom in. }
    \label{fig:qual_corrs}
\end{figure}

\begin{figure}[t]
    \centering
    \resizebox{\columnwidth}{!}{%
        \SetTblrInner{rowsep=1pt,colsep=1pt}
        \begin{tblr}{c|c|c|c|c}
        \SetCell[r=2]{c,m}GT & \SetCell[r=2]{c,m}AGS & \SetCell[r=2]{c,m}AGS-Full & \SetCell[c=2]{c}{Ours}&  \\
        \cline{3-5}
        &  &  &  FM-Dense & Ours \\
        \hline
        \includegraphics[valign=c, width=0.1\textwidth]{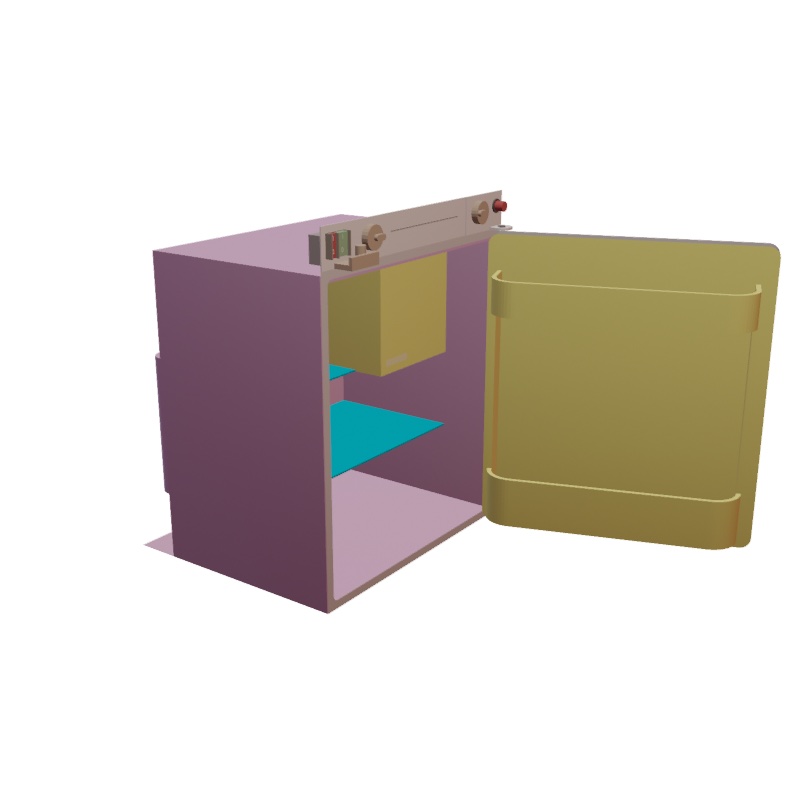}
        &\includegraphics[valign=c, width=0.1\textwidth]{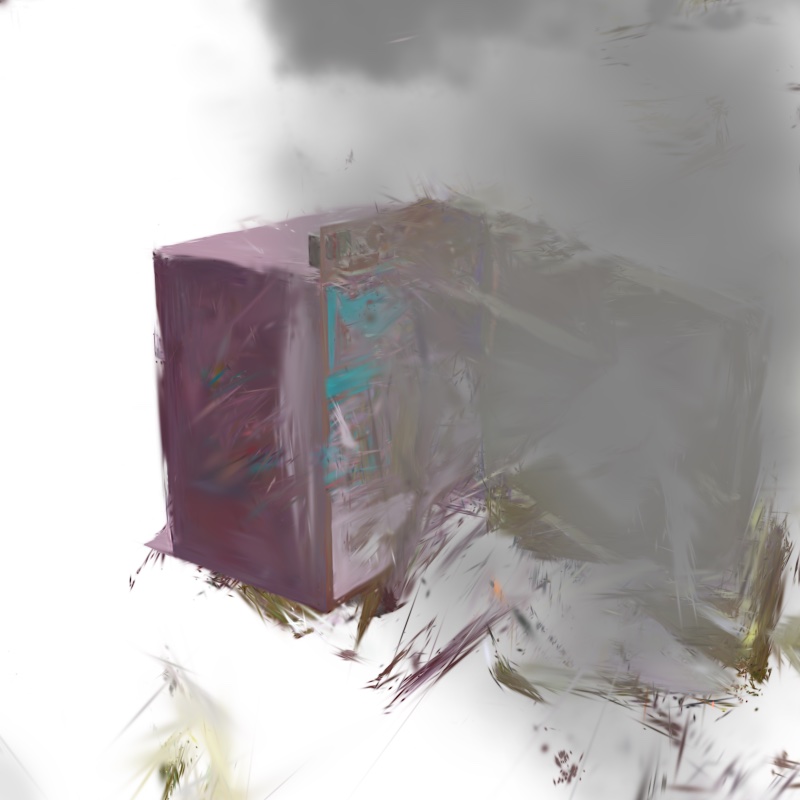}
        &\includegraphics[valign=c, width=0.1\textwidth]{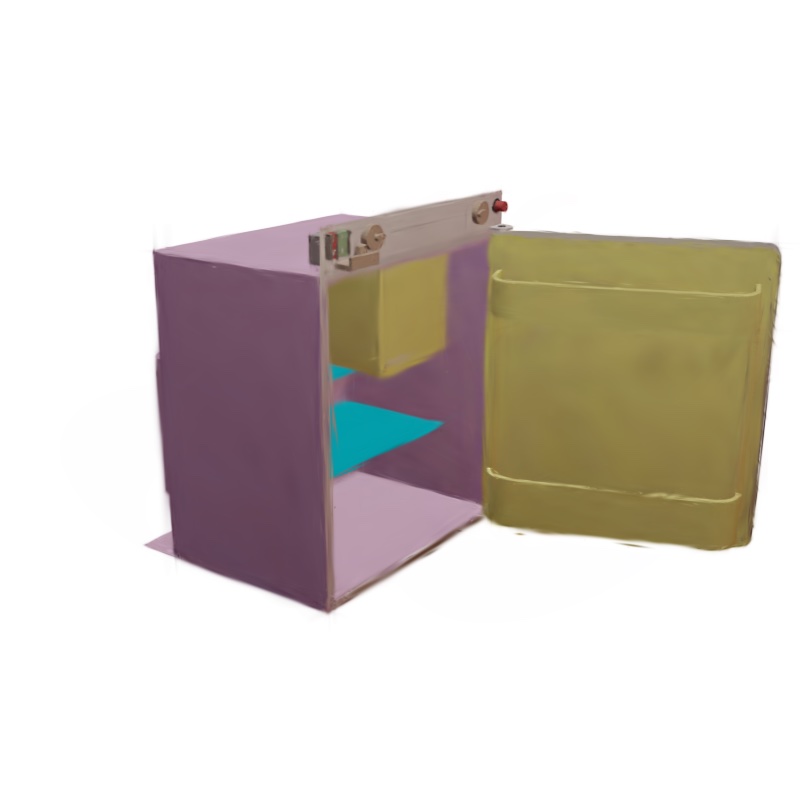}
        &\includegraphics[valign=c, width=0.1\textwidth]{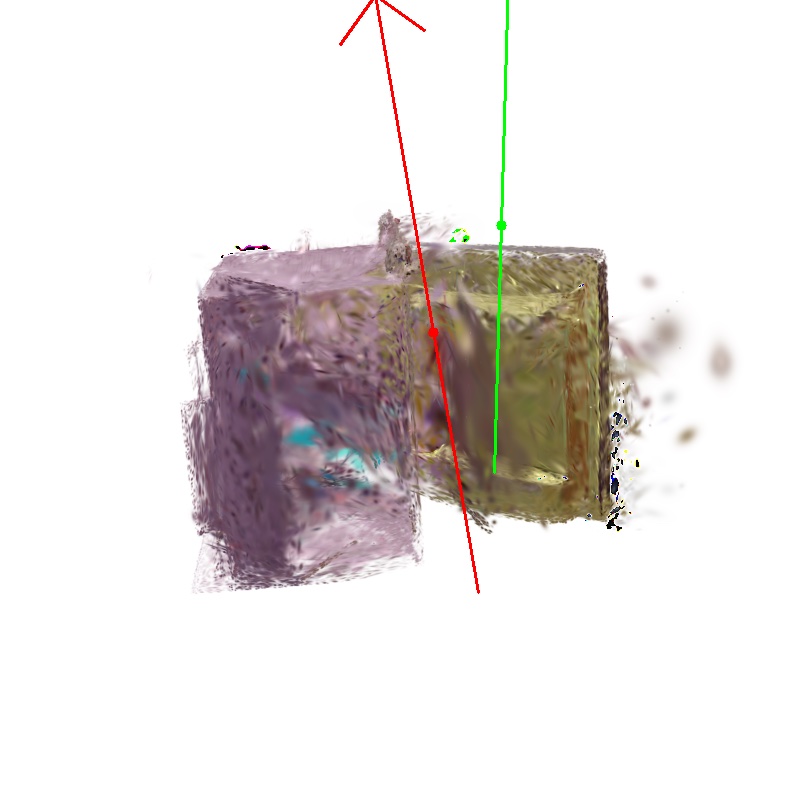}
        &\includegraphics[valign=c, width=0.1\textwidth]{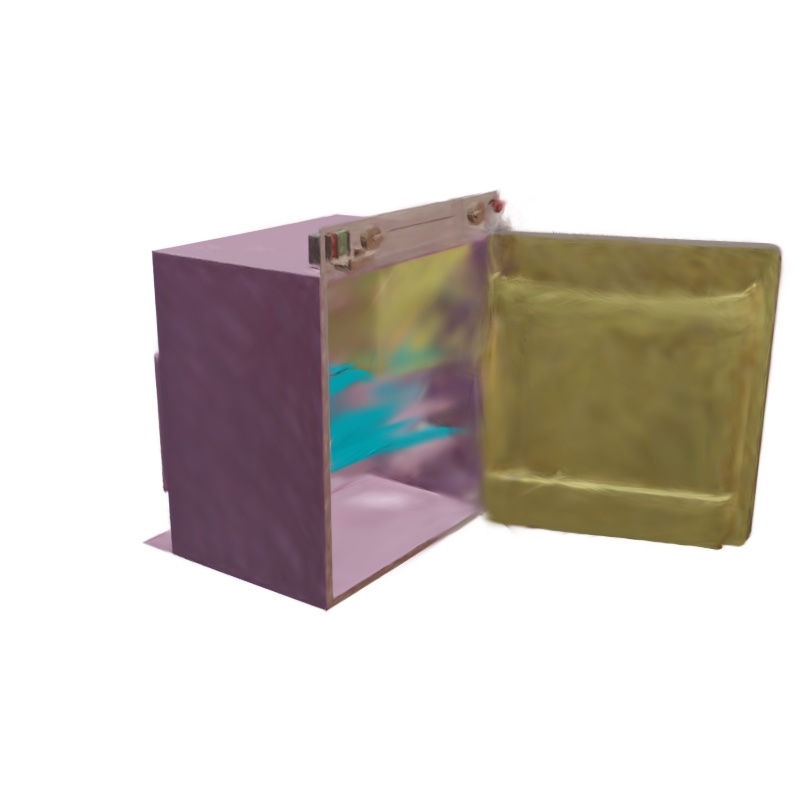}\\
        \includegraphics[valign=c, width=0.1\textwidth]{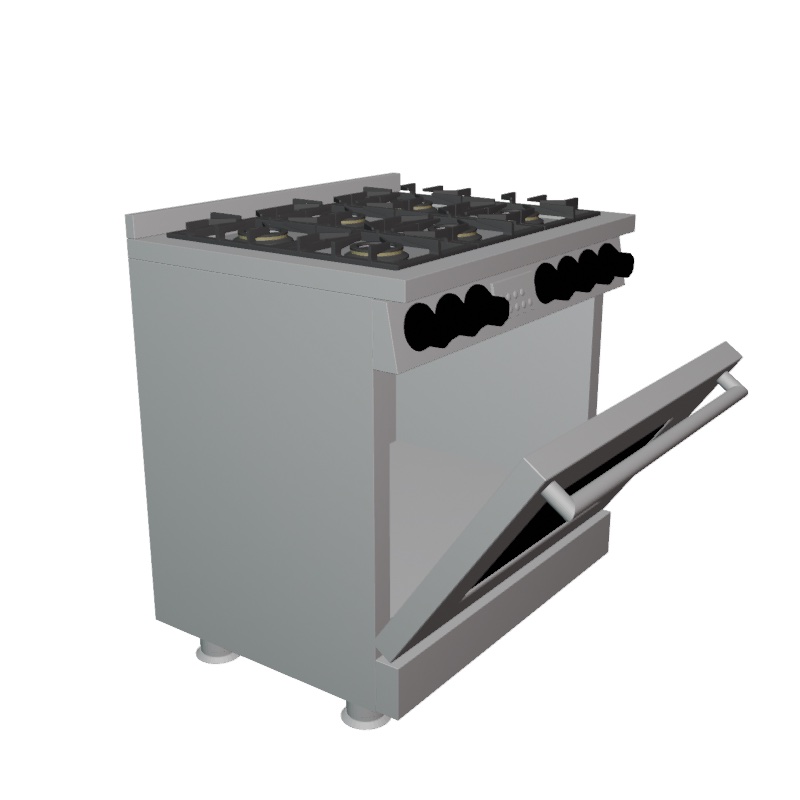}
        &\includegraphics[valign=c, width=0.1\textwidth]{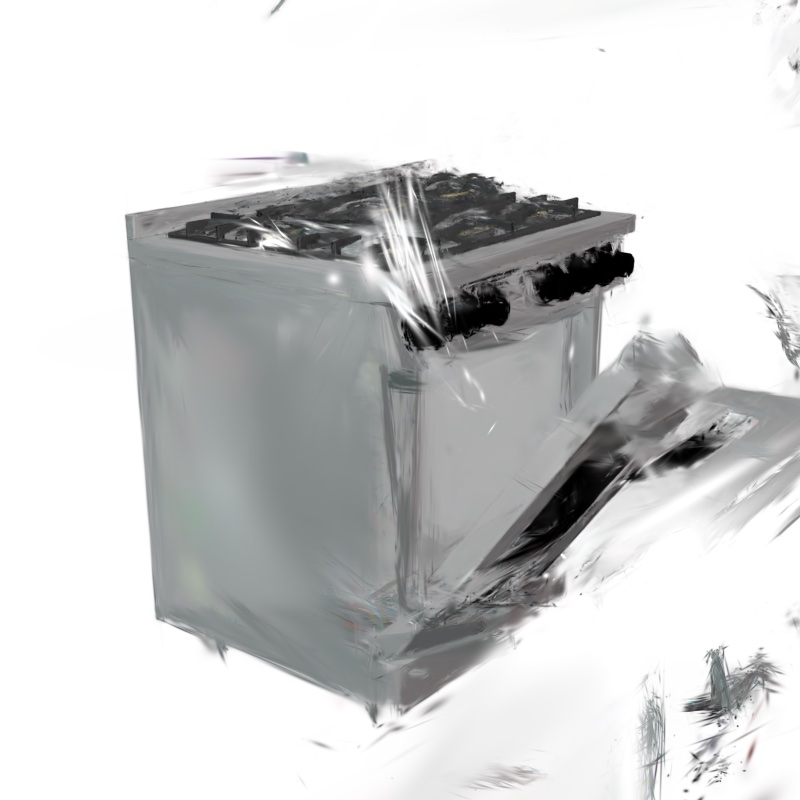}
        &\includegraphics[valign=c, width=0.1\textwidth]{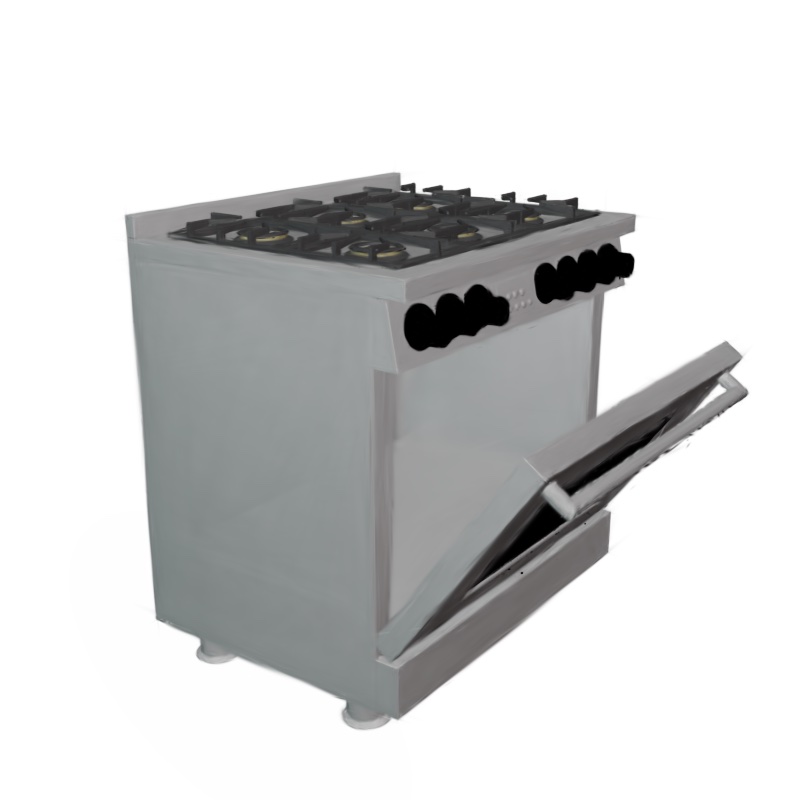}
        & failed
        &\includegraphics[valign=c, width=0.1\textwidth]{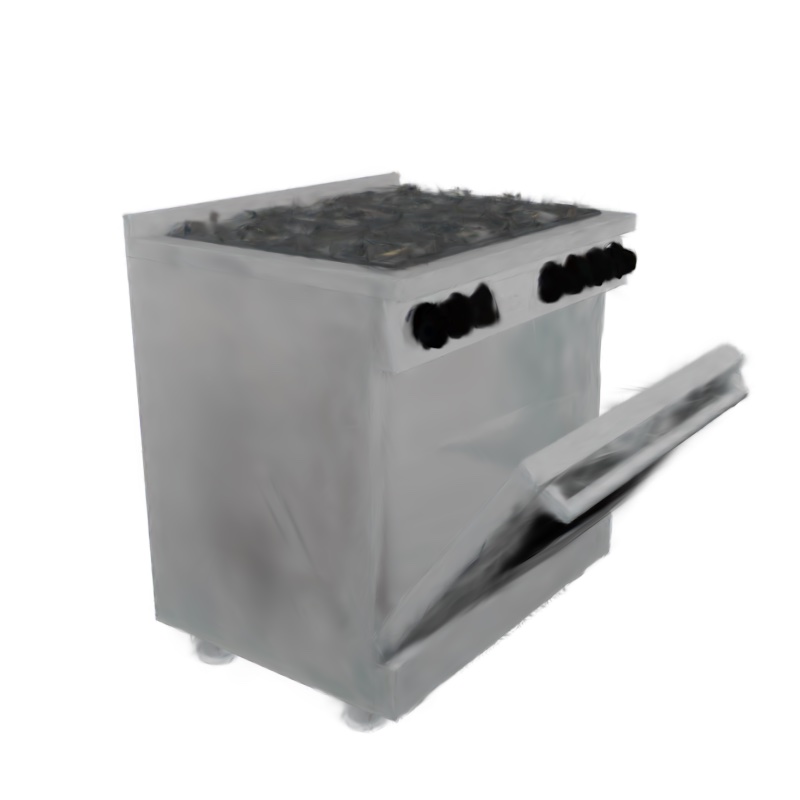}\\
        \includegraphics[valign=c, width=0.1\textwidth]{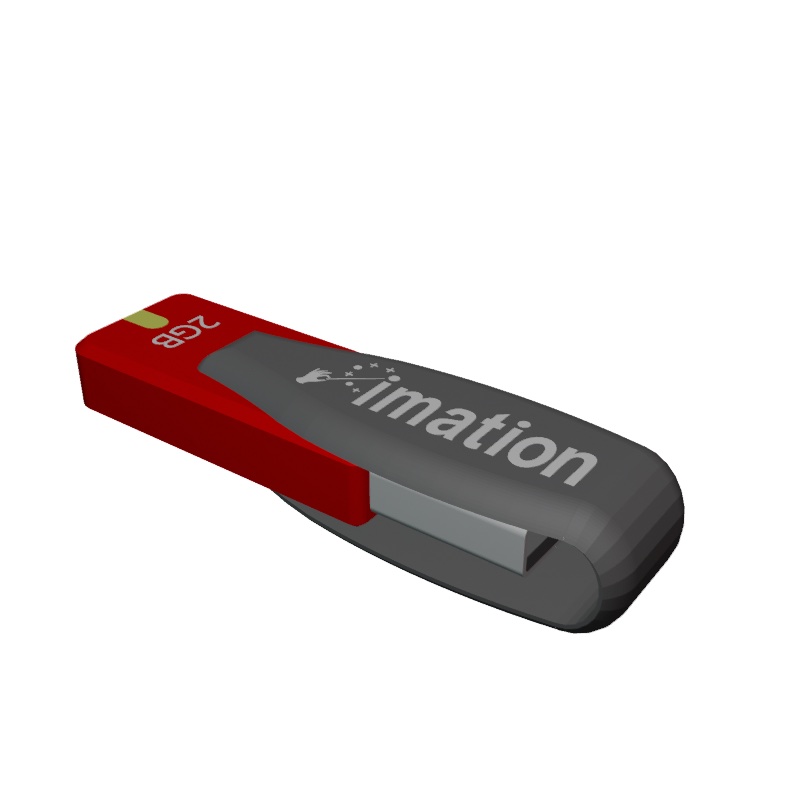}
        &\includegraphics[valign=c, width=0.1\textwidth]{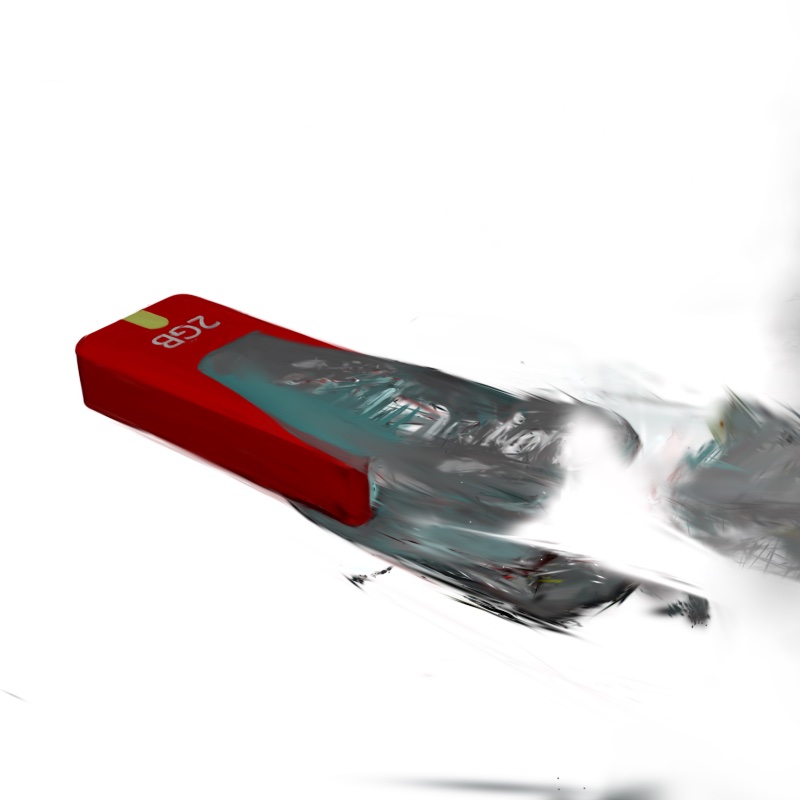}
        &\includegraphics[valign=c, width=0.1\textwidth]{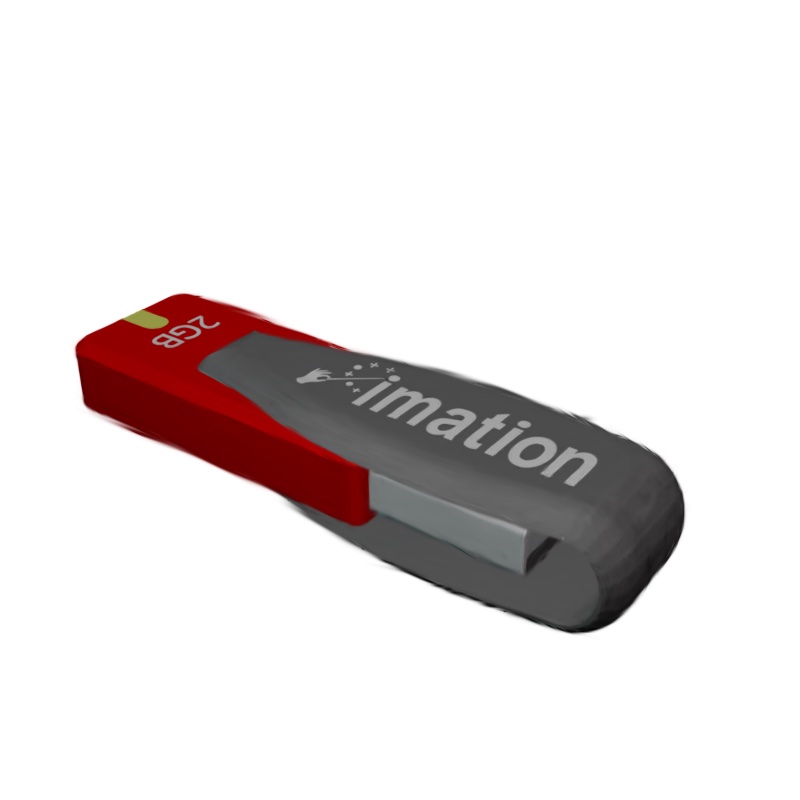}
        &\includegraphics[valign=c, width=0.1\textwidth]{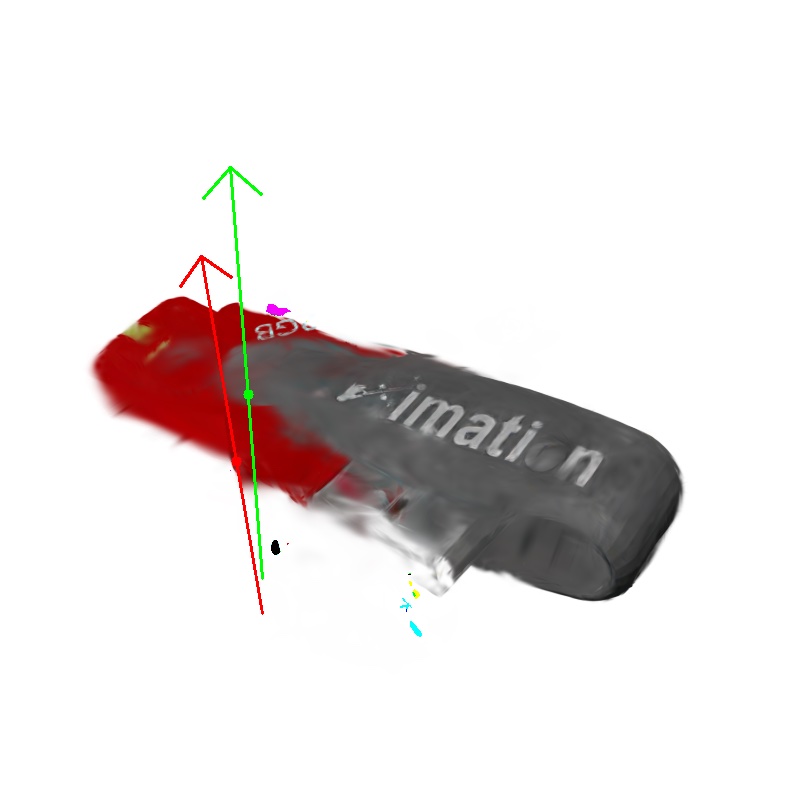}
        &\includegraphics[valign=c, width=0.1\textwidth]{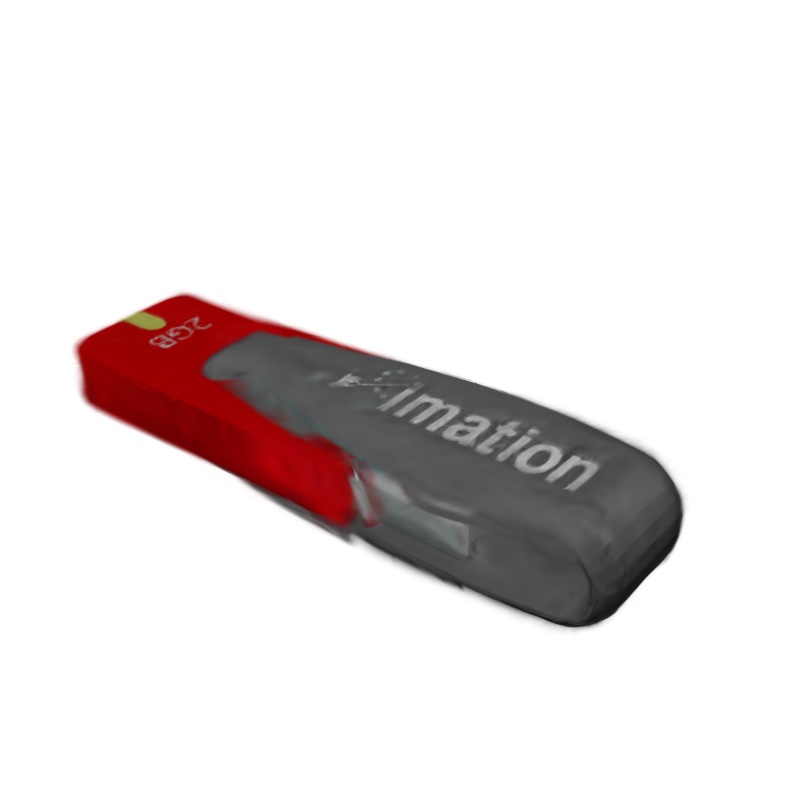}\\
        \includegraphics[valign=c, width=0.1\textwidth]{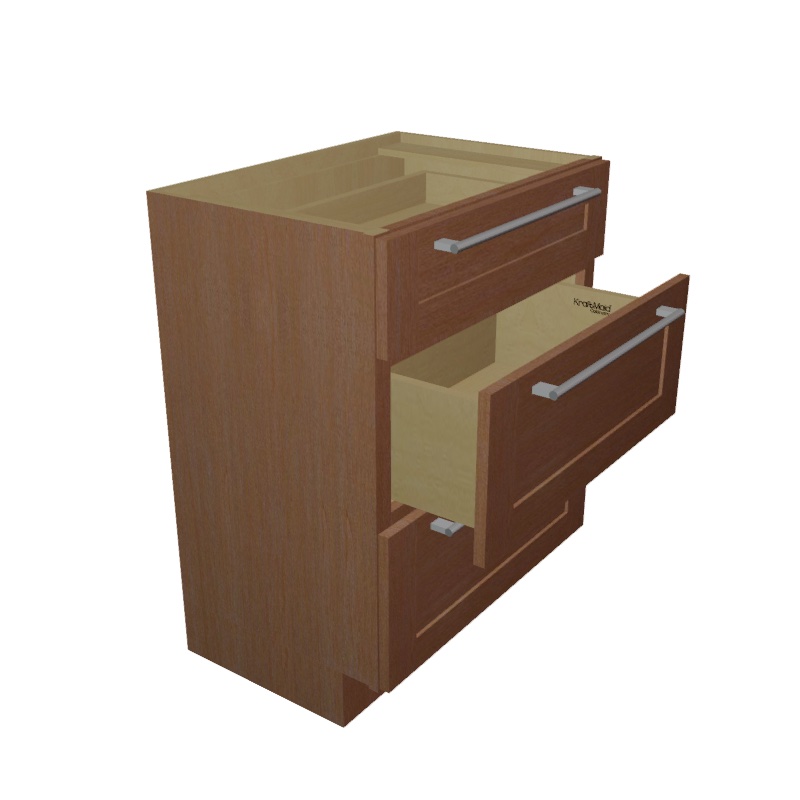}
        &\includegraphics[valign=c, width=0.1\textwidth]{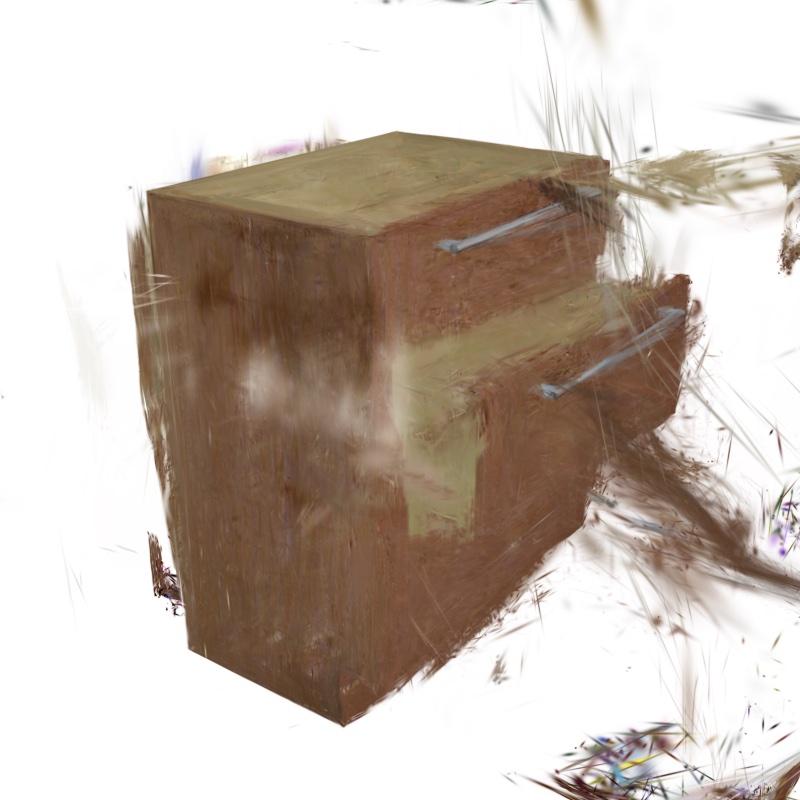}
        &\includegraphics[valign=c, width=0.1\textwidth]{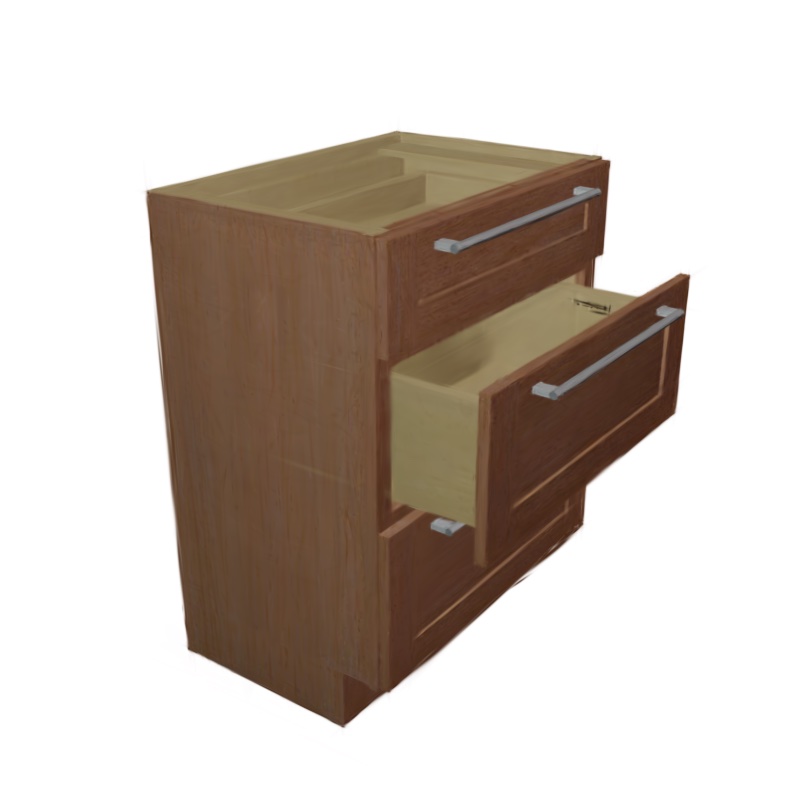}
        & failed
        &\includegraphics[valign=c, width=0.1\textwidth]{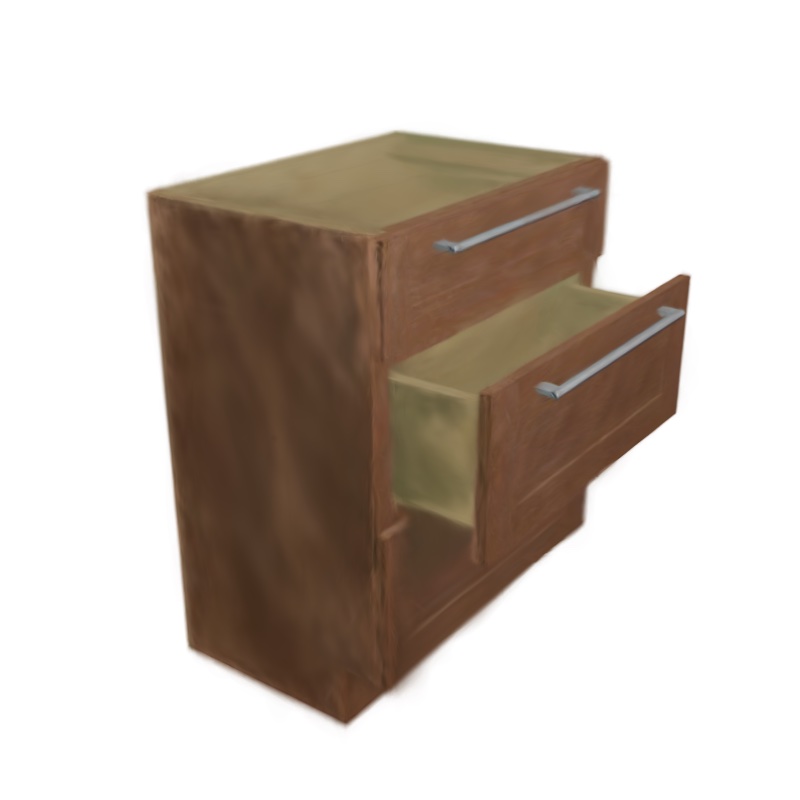}
        
        \end{tblr}
    }
    \caption{\textbf{Novel view synthesis at target state.} AGS-GT and AGS-VGGT fail with only 4 views. Results of ours matches AGS-Full (100 views + GT poses) visually.}
    \label{fig:novel_articulation_rendering}
\end{figure}

\begin{figure}[t]
    \centering
    \resizebox{\columnwidth}{!}{%
        \SetTblrInner{rowsep=0pt,colsep=1pt}
        \scriptsize
        \begin{tblr}{cc|ccccc}
        \SetCell[c=2]{c}{RGB input} & & {AGS-GT} & {AGS-Full} & {FM-Dense} & {Ours} & {GT}
            \\
        \hline
        \includegraphics[valign=c, width=0.1\textwidth]{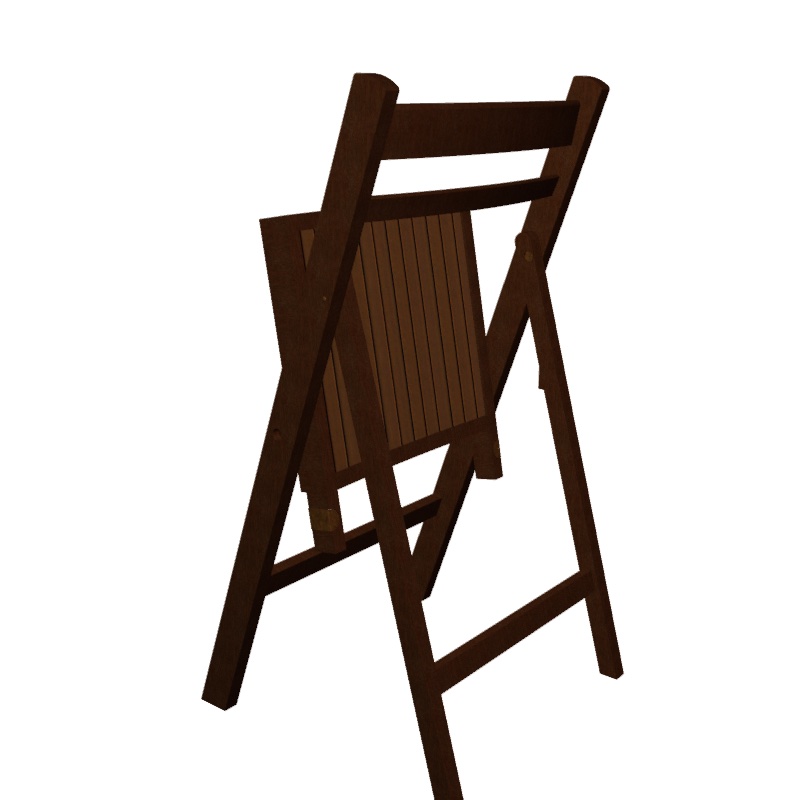}
        &\includegraphics[valign=c, width=0.1\textwidth]{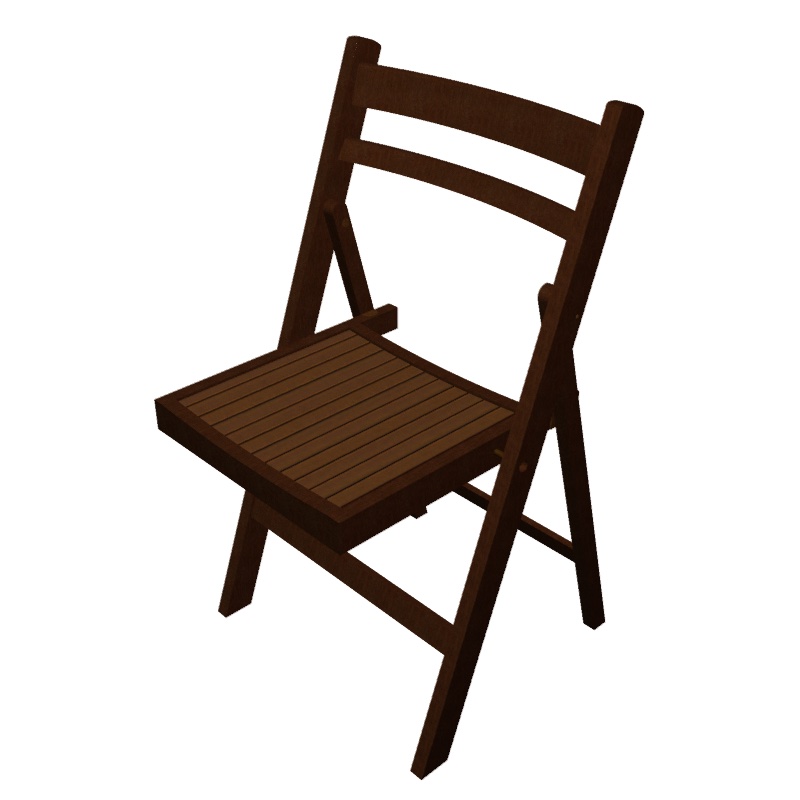}
        & \includegraphics[valign=c, width=0.1\textwidth]{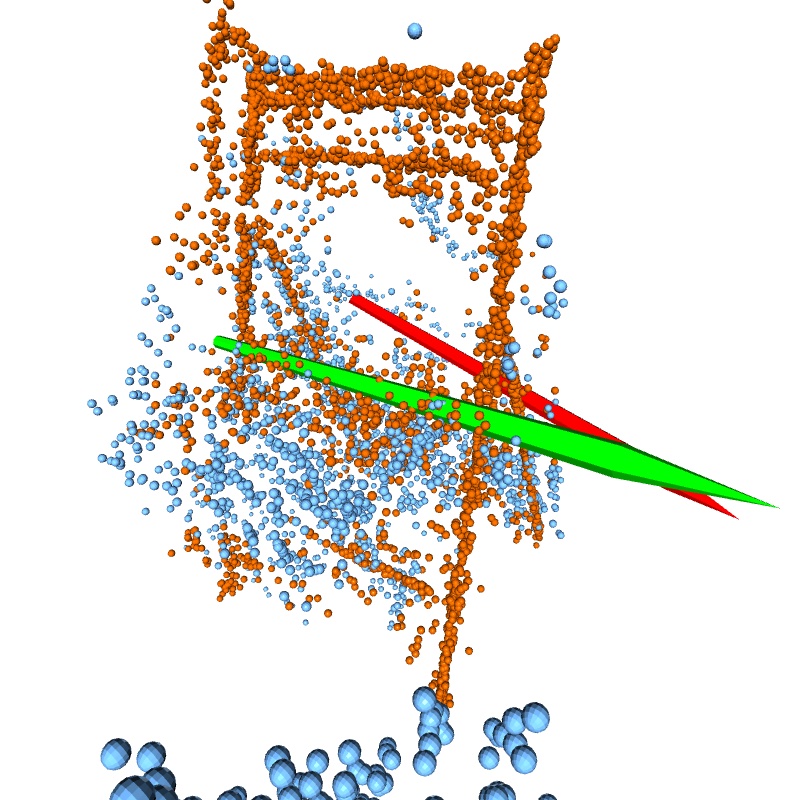}
        & \includegraphics[valign=c, width=0.1\textwidth]{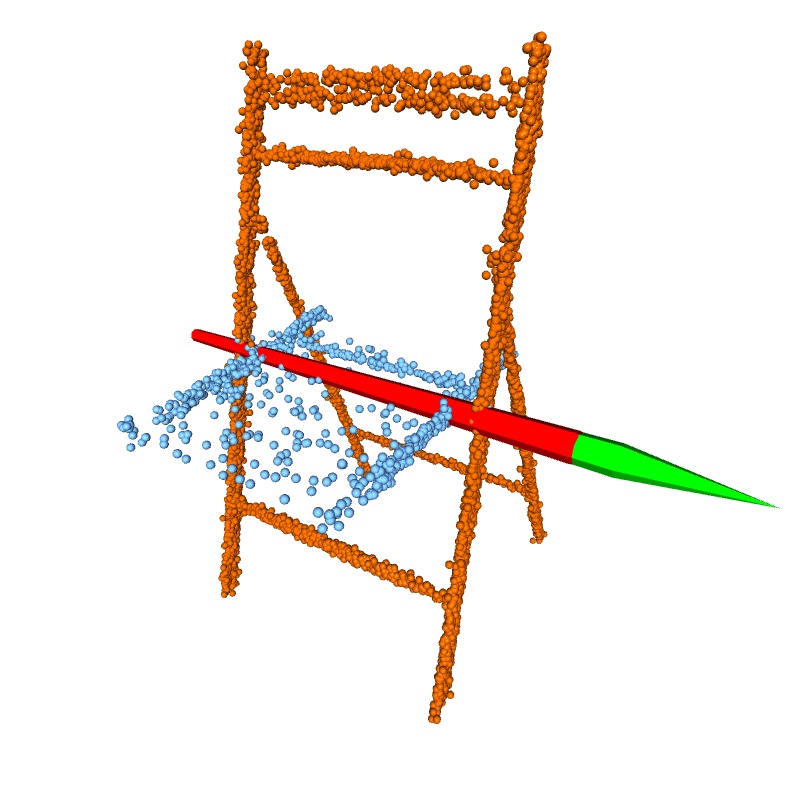}
        & \includegraphics[valign=c, width=0.1\textwidth]{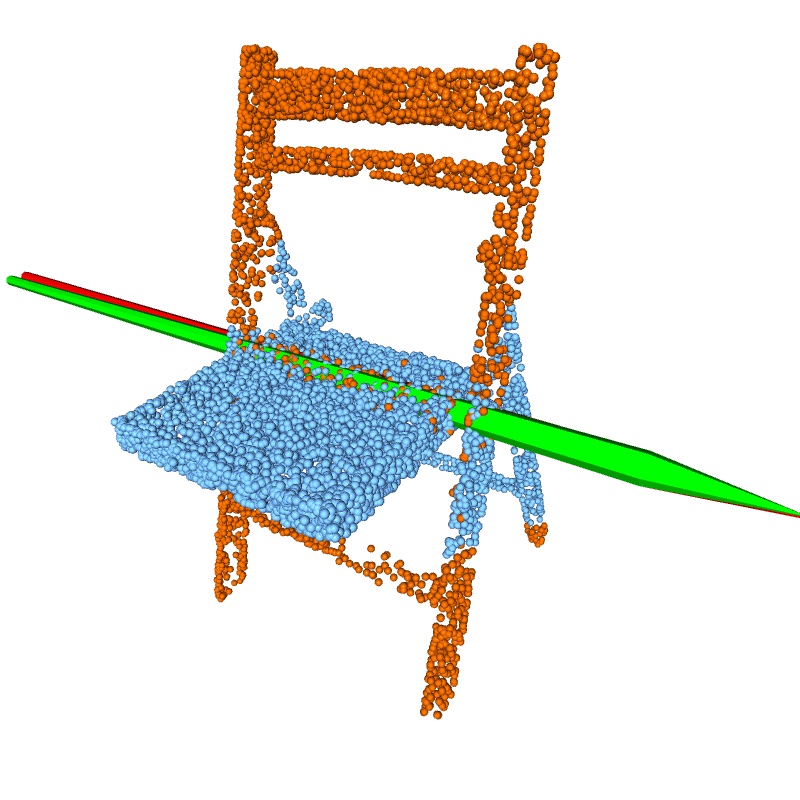}
        & \includegraphics[valign=c, width=0.1\textwidth]{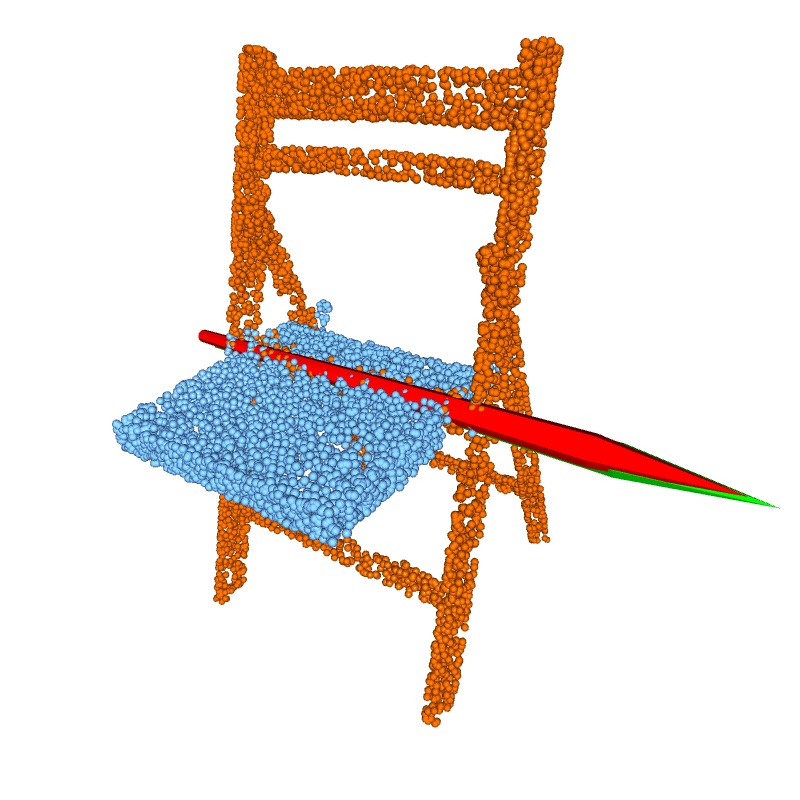}
        & \includegraphics[valign=c, width=0.1\textwidth]{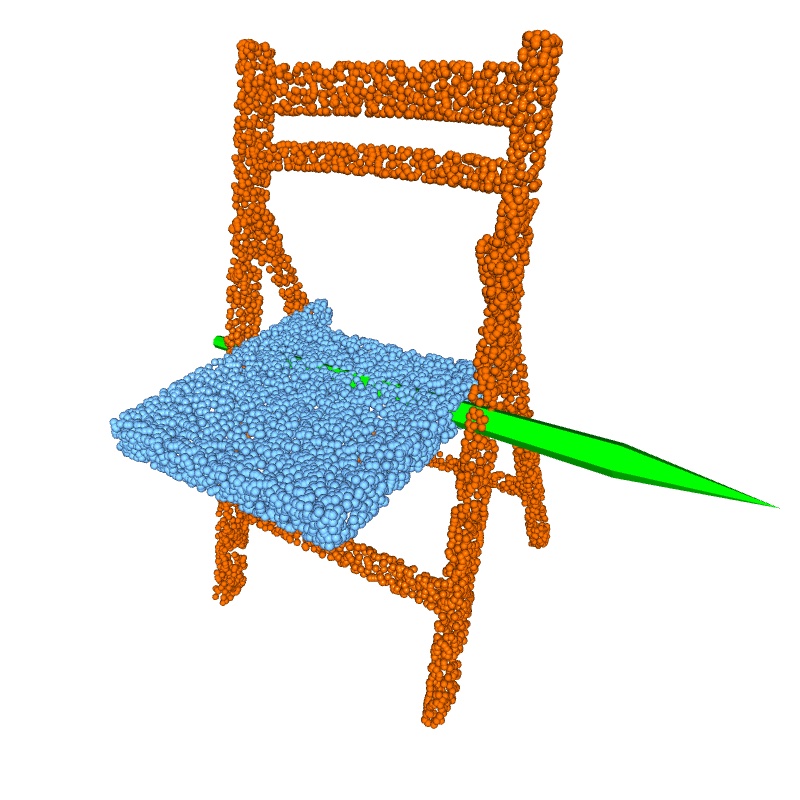} \\
        \includegraphics[valign=c, width=0.1\textwidth]{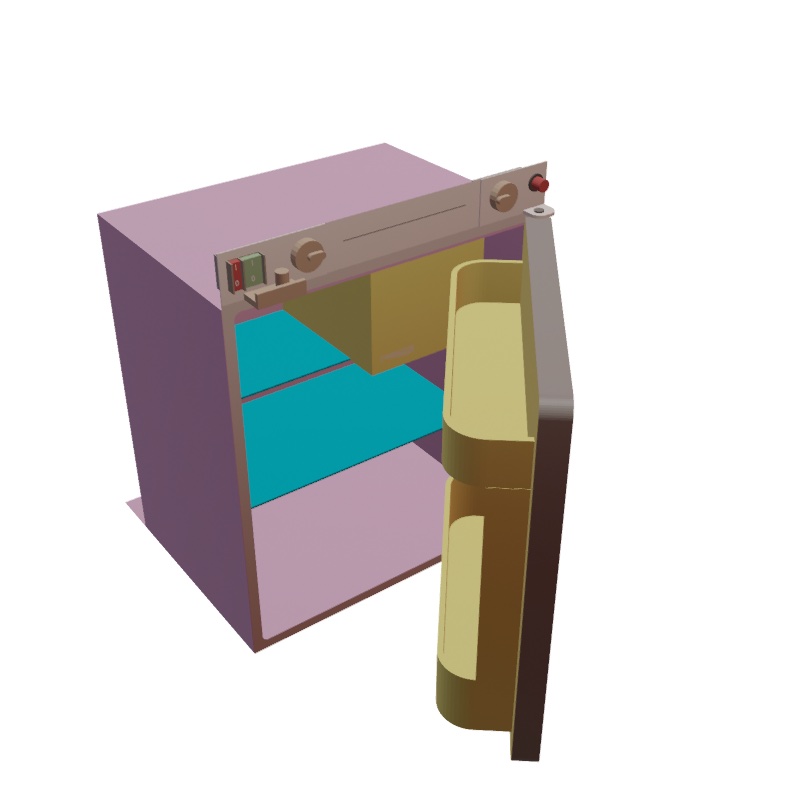}
        &\includegraphics[valign=c, width=0.1\textwidth]{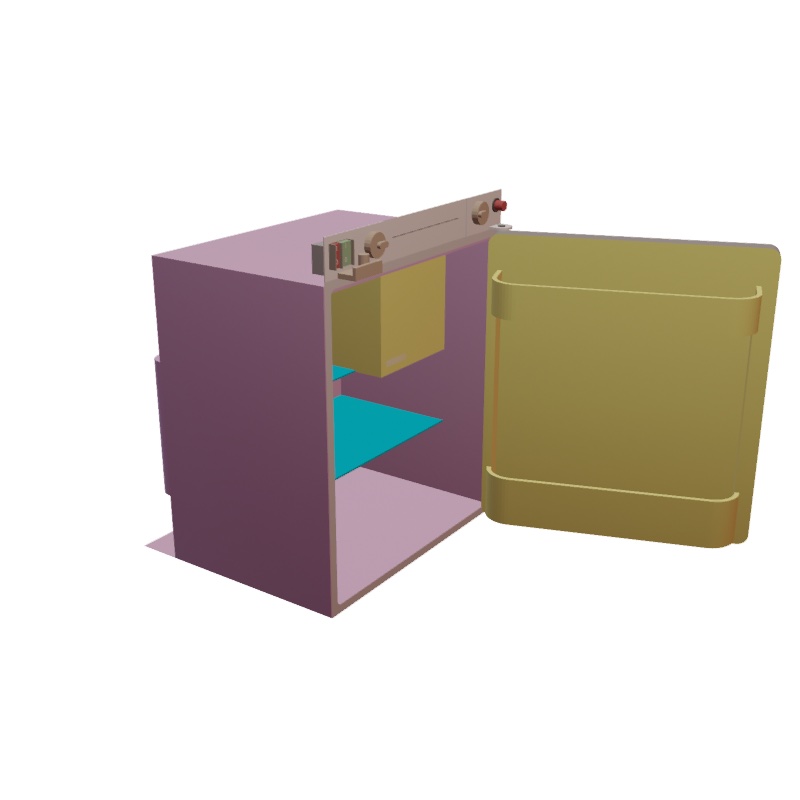}
        & \includegraphics[valign=c, width=0.1\textwidth]{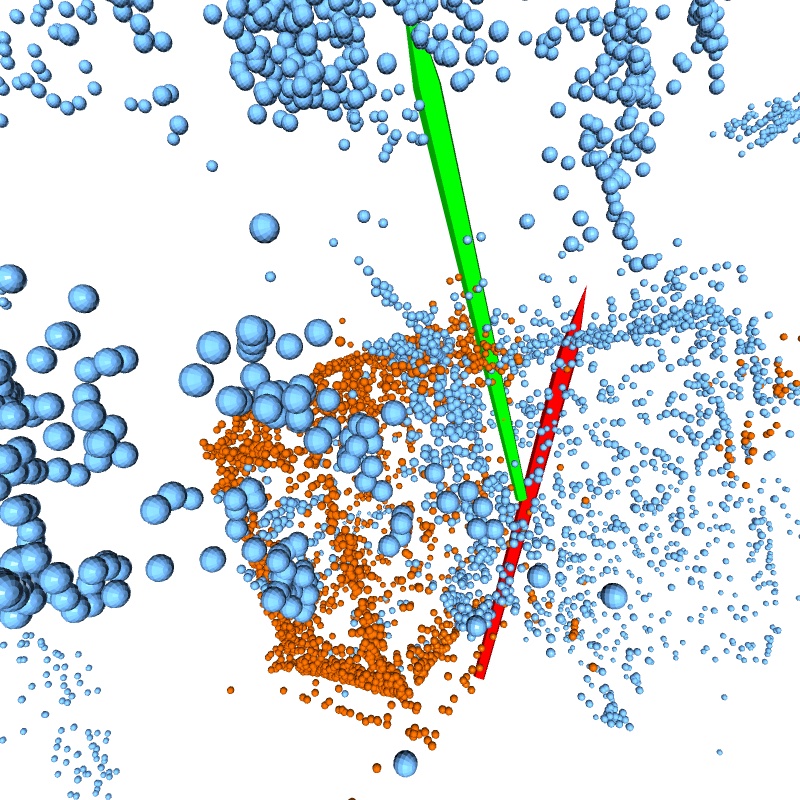}
        & \includegraphics[valign=c, width=0.1\textwidth]{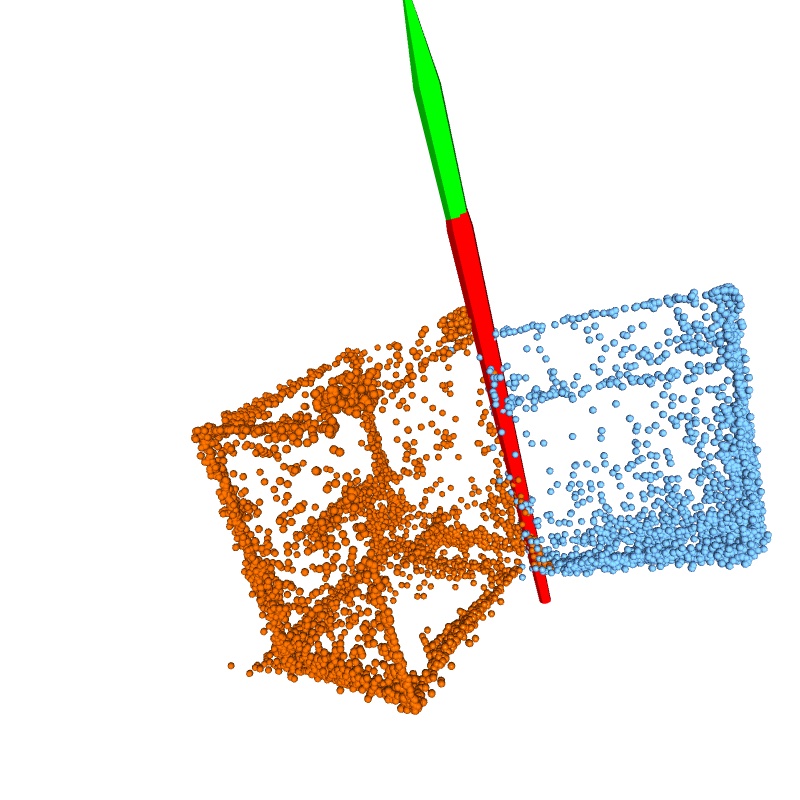}
        & \includegraphics[valign=c, width=0.1\textwidth]{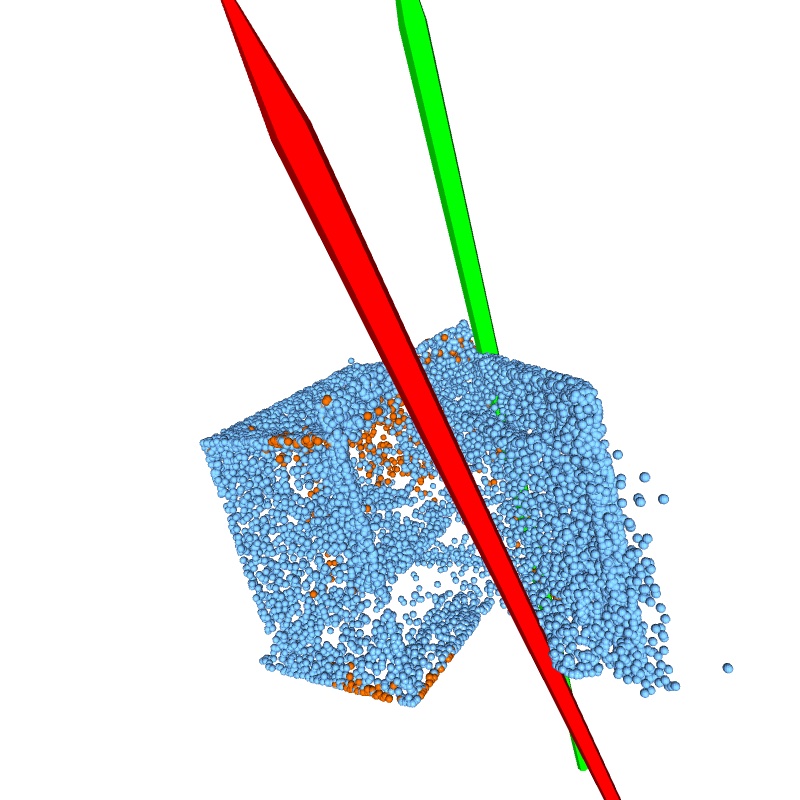}
        & \includegraphics[valign=c, width=0.1\textwidth]{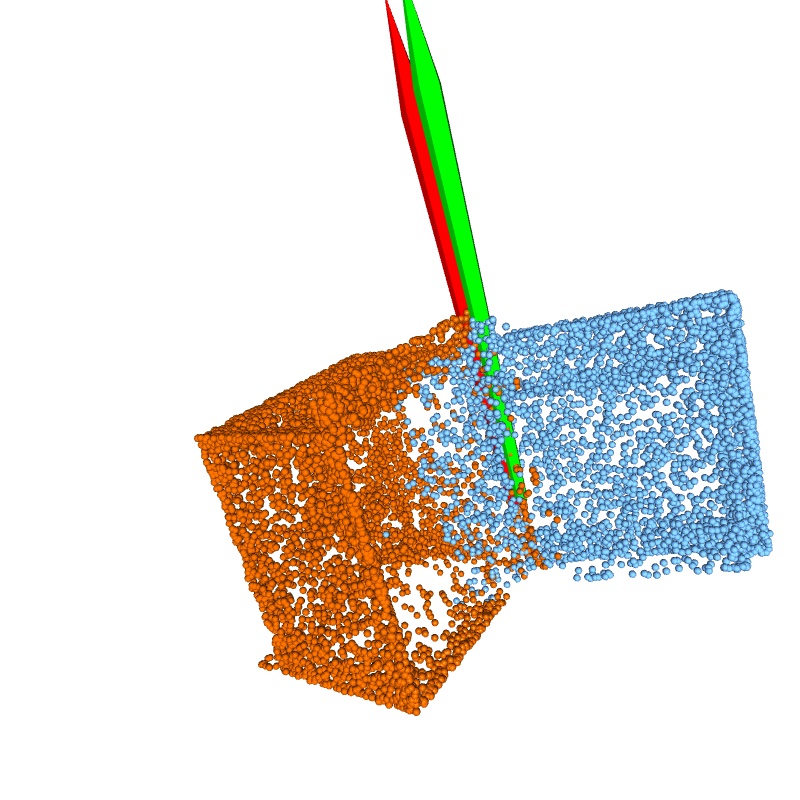}
        & \includegraphics[valign=c, width=0.1\textwidth]{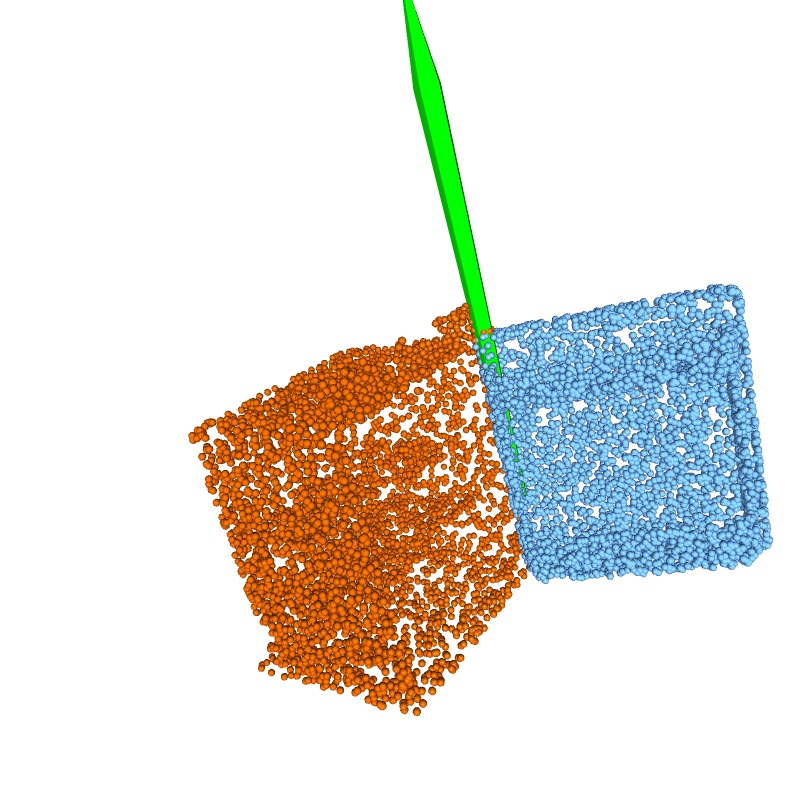} \\
        \includegraphics[valign=c, width=0.1\textwidth]{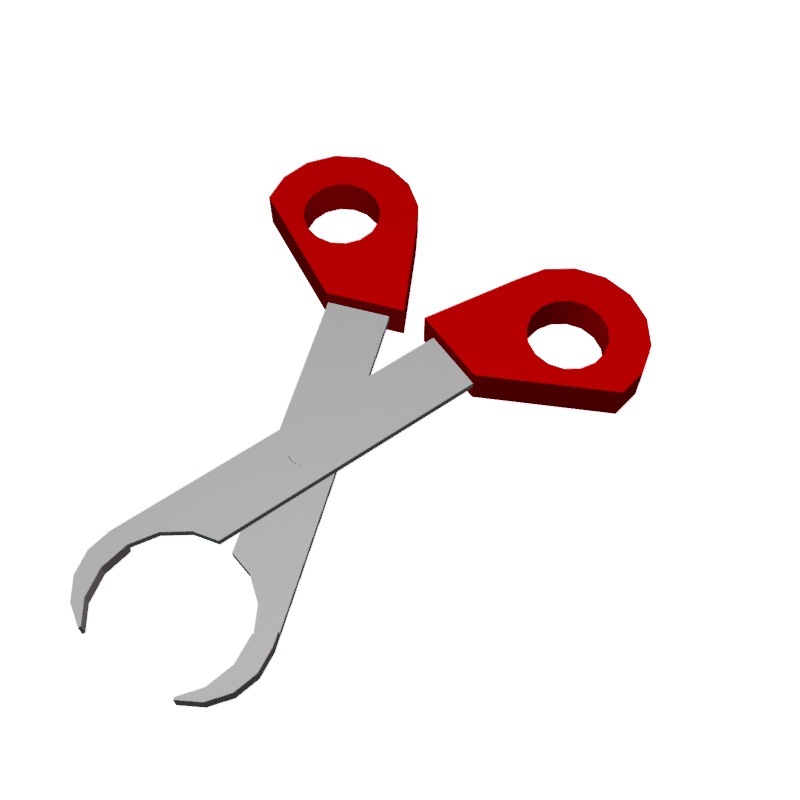}
        &\includegraphics[valign=c, width=0.1\textwidth]{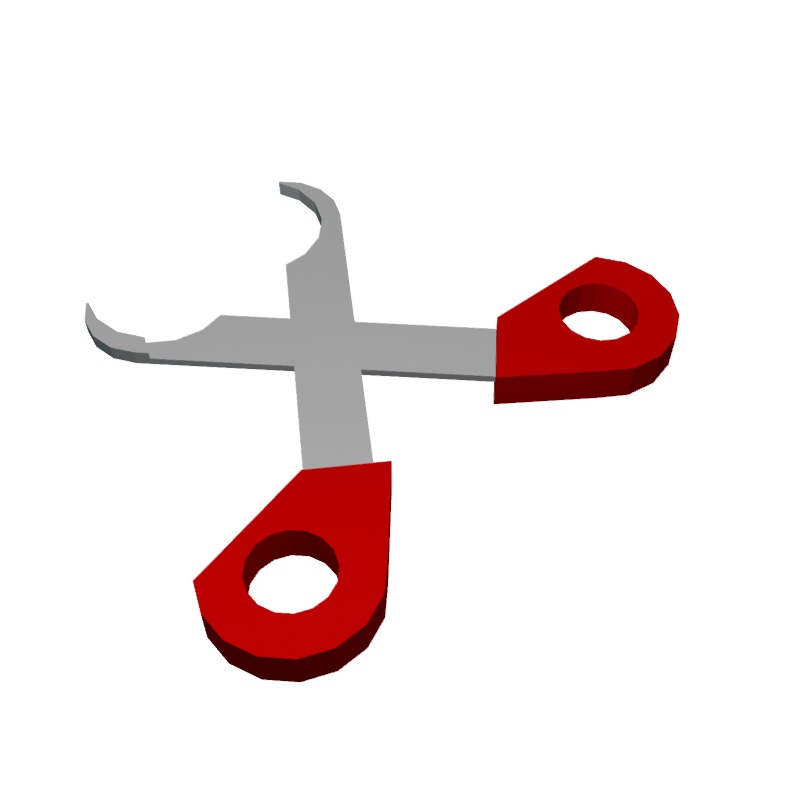}
        & \includegraphics[valign=c, width=0.1\textwidth]{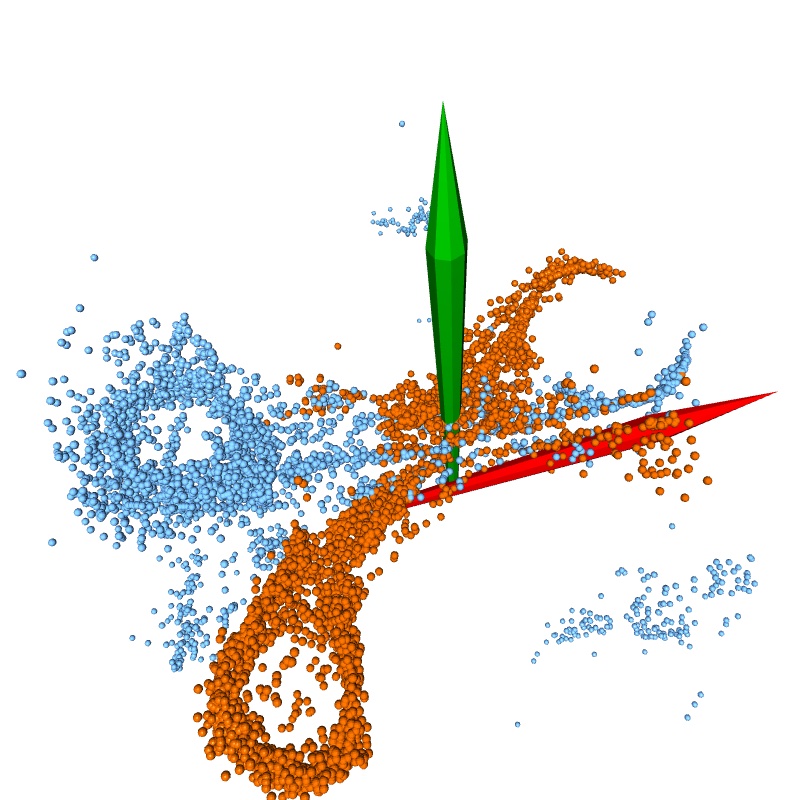}
        & \includegraphics[valign=c, width=0.1\textwidth]{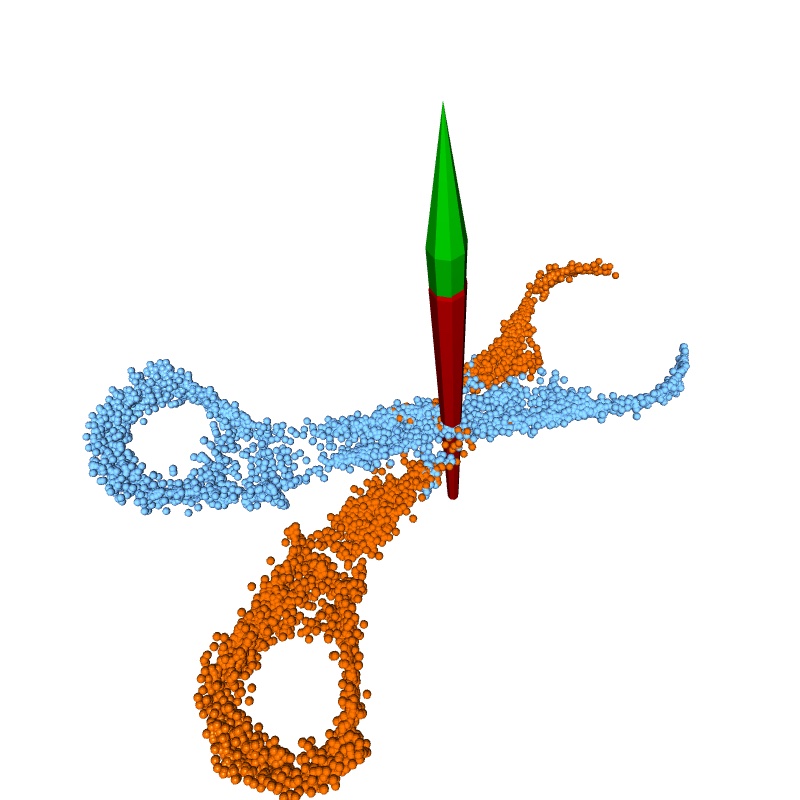}
        & \includegraphics[valign=c, width=0.1\textwidth]{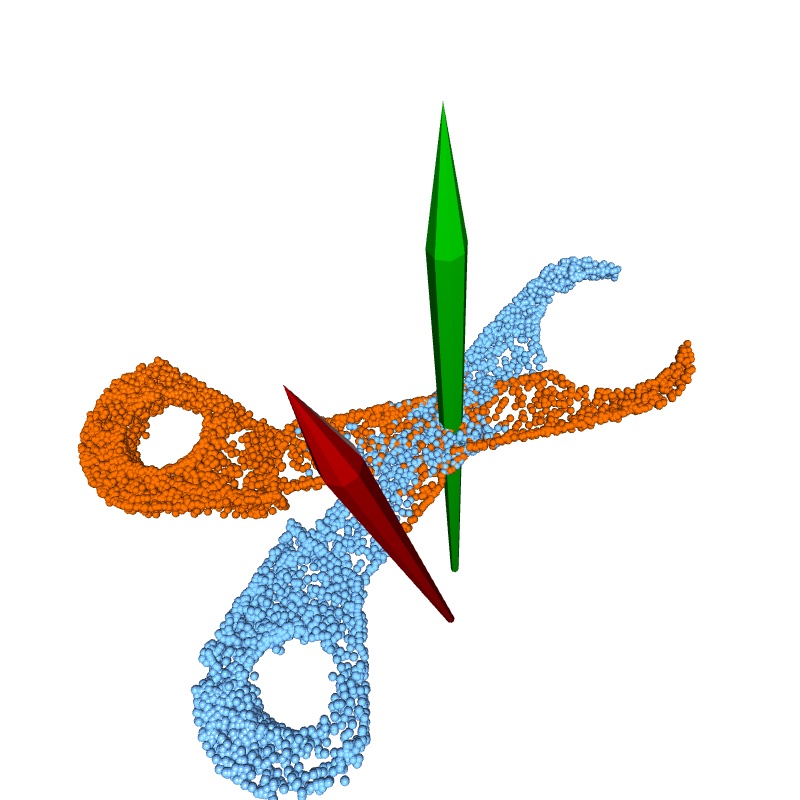}
        & \includegraphics[valign=c, width=0.1\textwidth]{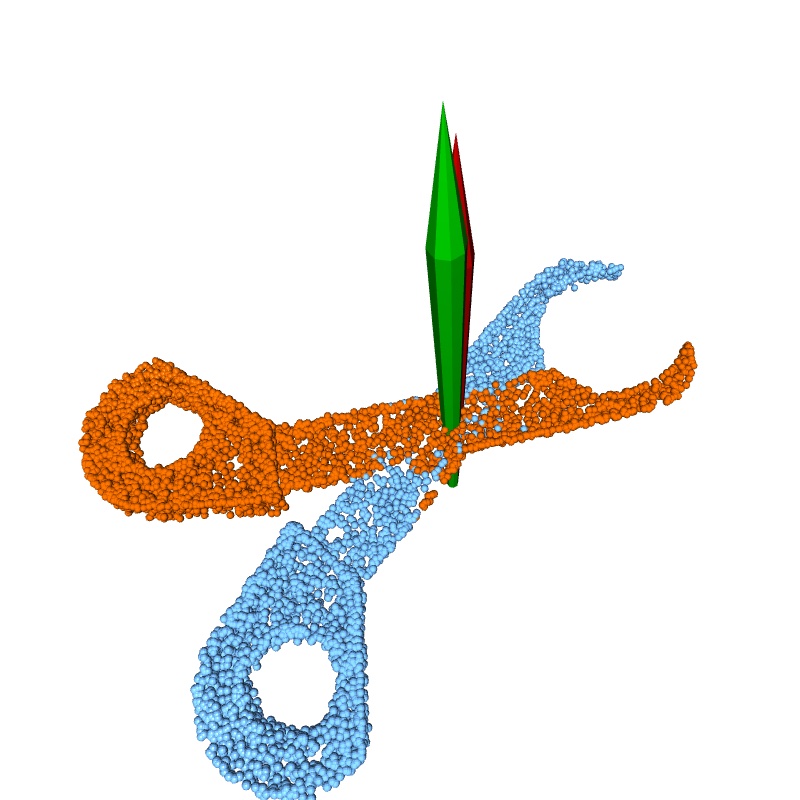}
        & \includegraphics[valign=c, width=0.1\textwidth]{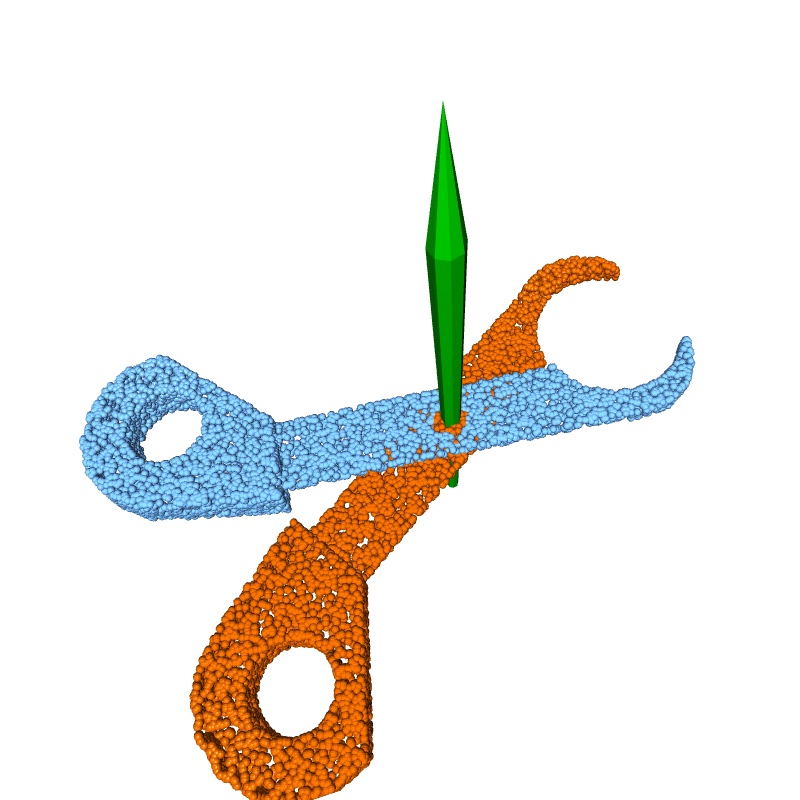} \\
        \includegraphics[valign=c, width=0.1\textwidth]{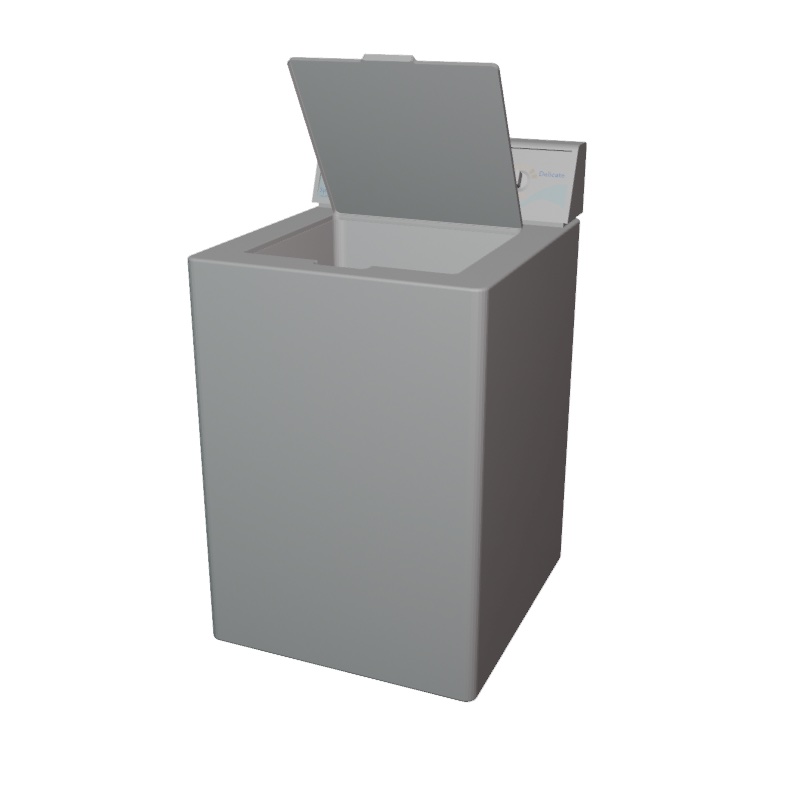}
        &\includegraphics[valign=c, width=0.1\textwidth]{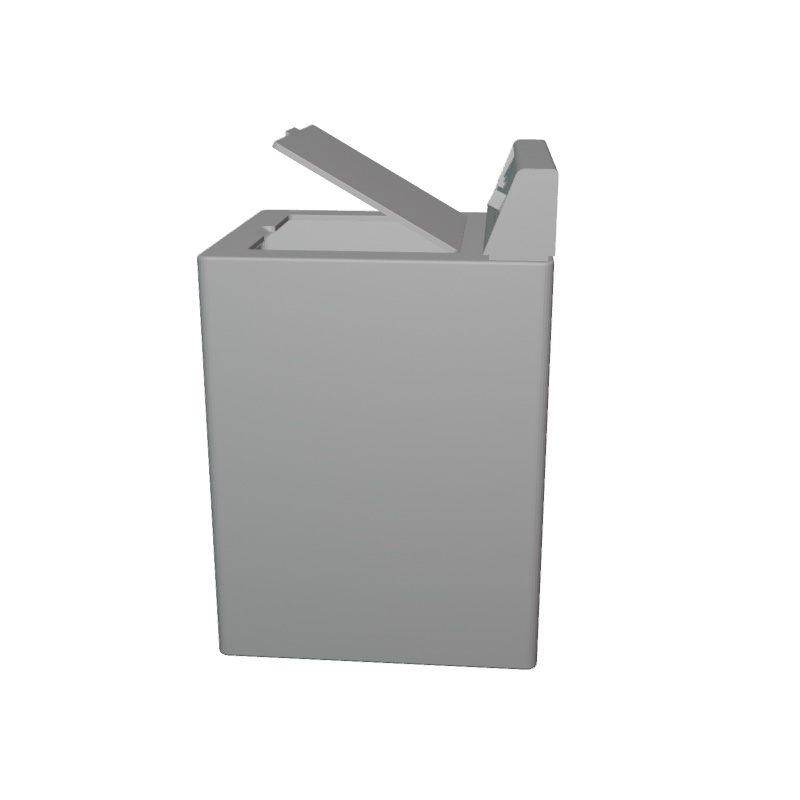}
        & \includegraphics[valign=c, width=0.1\textwidth]{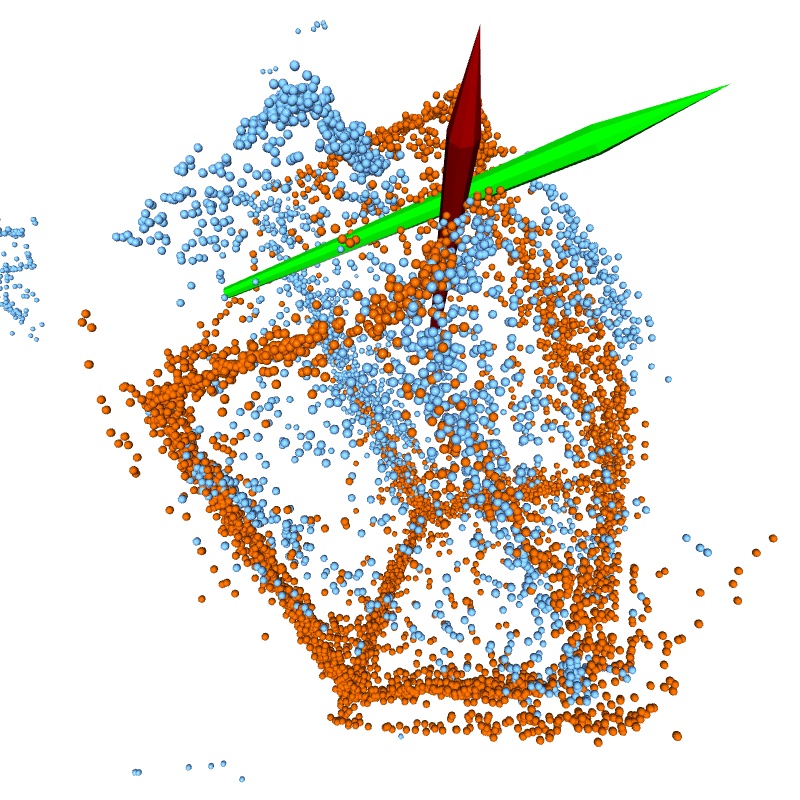}
        & \includegraphics[valign=c, width=0.1\textwidth]{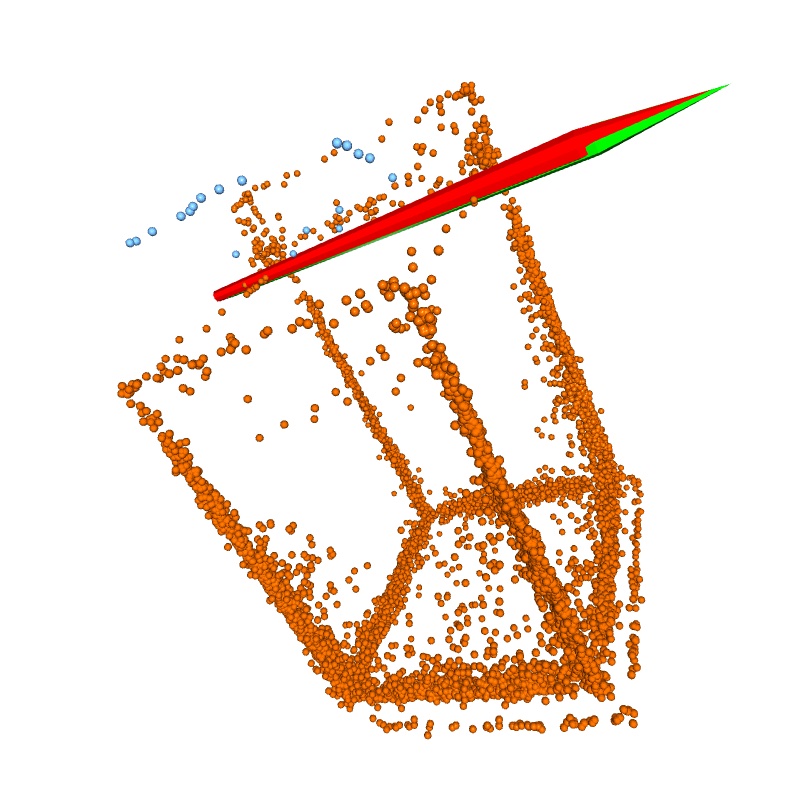}
        & \includegraphics[valign=c, width=0.1\textwidth]{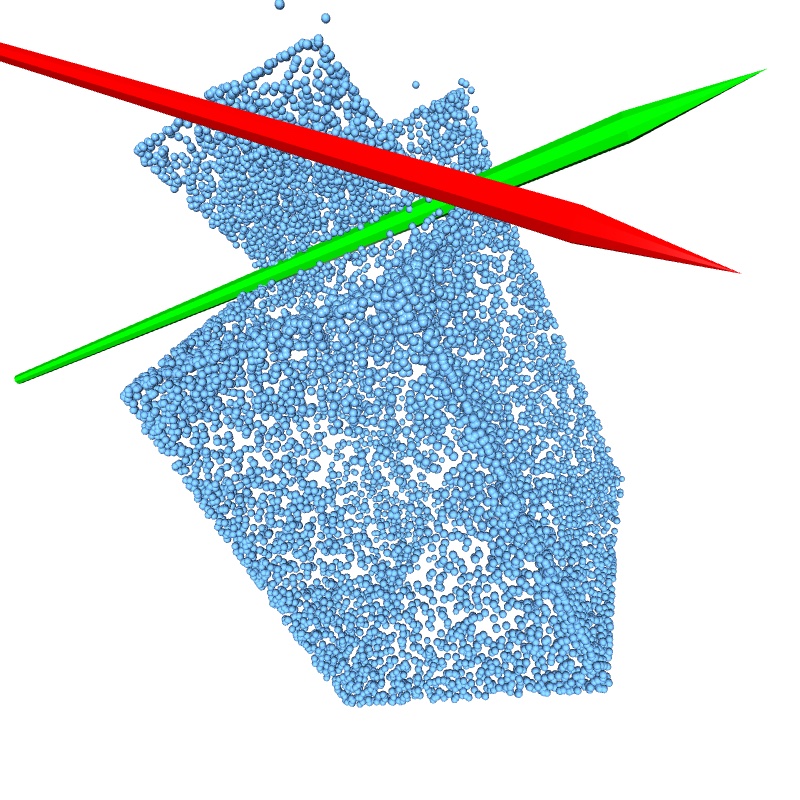}
        & \includegraphics[valign=c, width=0.1\textwidth]{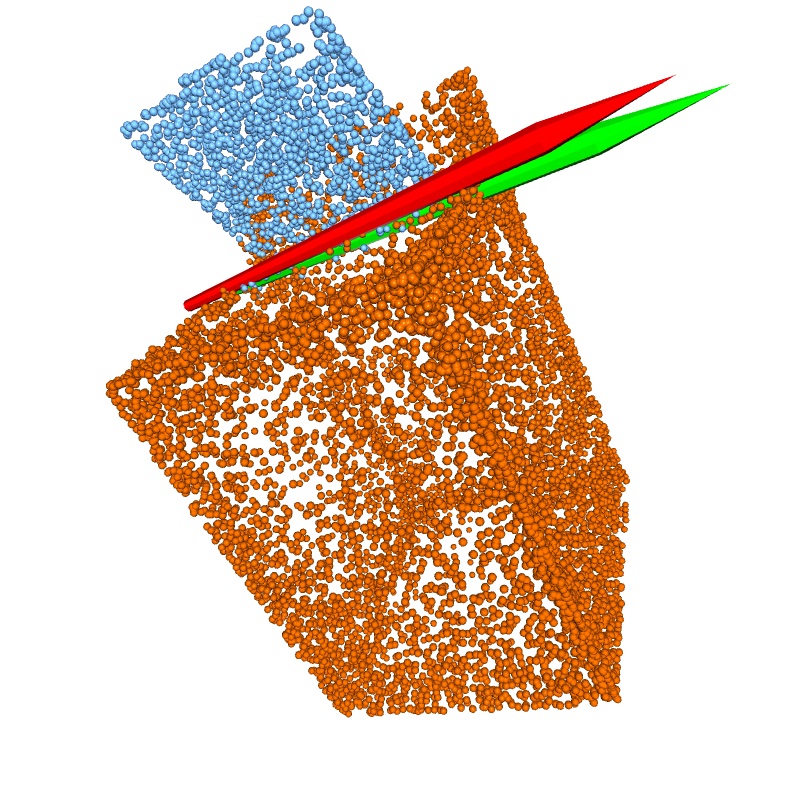}
        & \includegraphics[valign=c, width=0.1\textwidth]{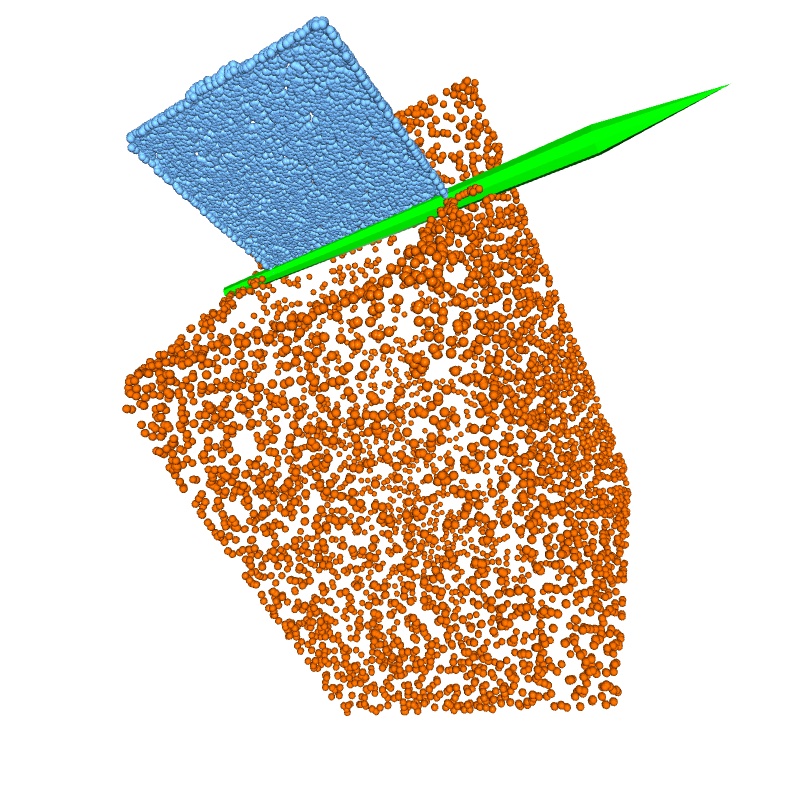} \\
        \end{tblr}
    }
    \caption{\textbf{Visualization of 3D Gaussian points with part-level segmentation and articulation axes.} Ground-truth axes are shown in green, predictions in red. AGS-Full denotes the AGS pipeline with dense and posed images input. Better view in color and zoom in. }
    \label{fig:segmentation}
\end{figure}

\begin{figure}[t]
    \centering
    \resizebox{\columnwidth}{!}{%
        \SetTblrInner{rowsep=0pt,colsep=1pt}
        \scriptsize
        \begin{tblr}{
            colspec={ccc},
            row{1}={abovesep=0pt,belowsep=0pt},
            row{2-0}={abovesep=-1pt,belowsep=-1pt}
        }
            GaussReg & FM-Dense & Ours \\
            \hline
            \includegraphics[valign=c, width=0.3\textwidth]{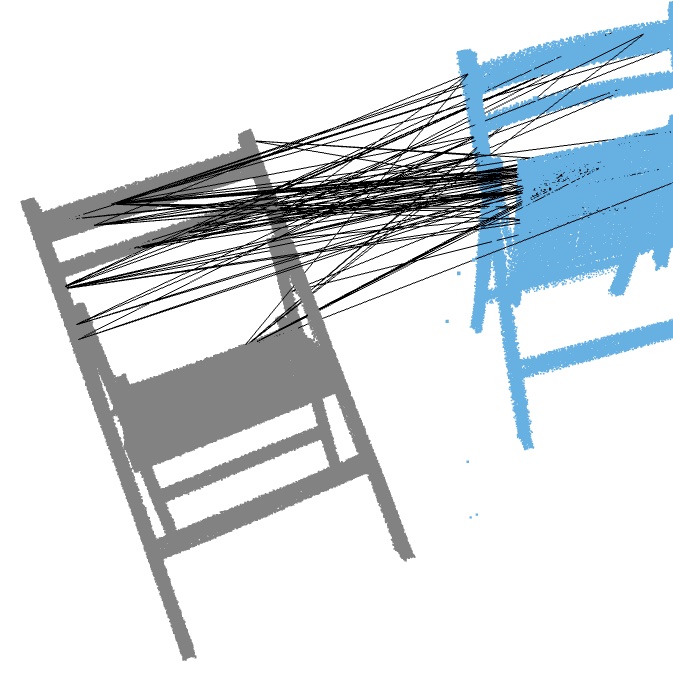}
            &\includegraphics[valign=c, width=0.3\textwidth]{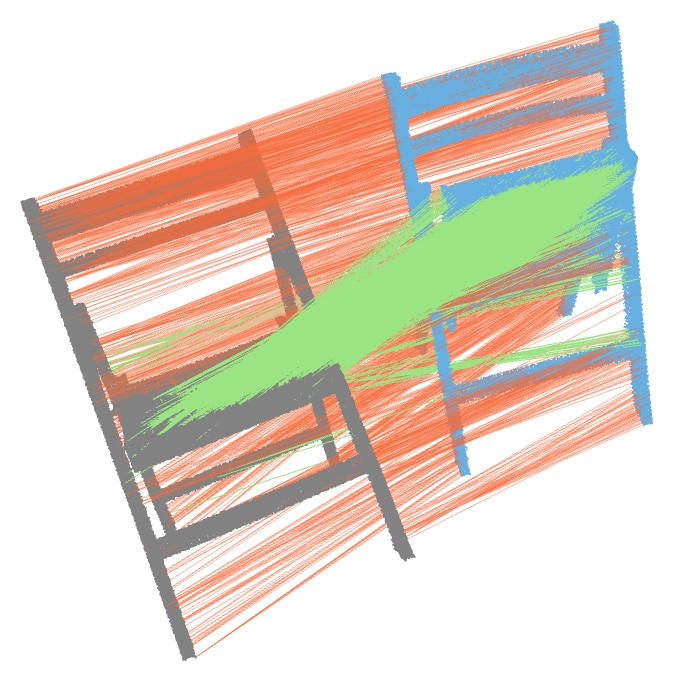}
            &\includegraphics[valign=c, width=0.3\textwidth]{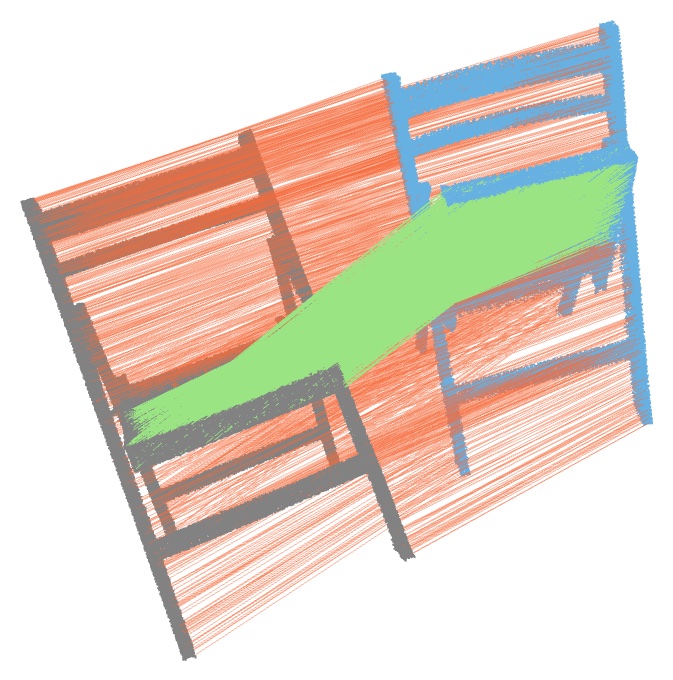}
            \\
            \includegraphics[valign=c, width=0.3\textwidth]{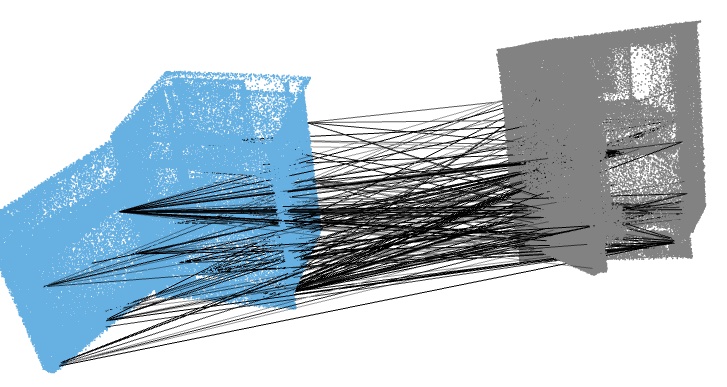}
            &\includegraphics[valign=c, width=0.3\textwidth]{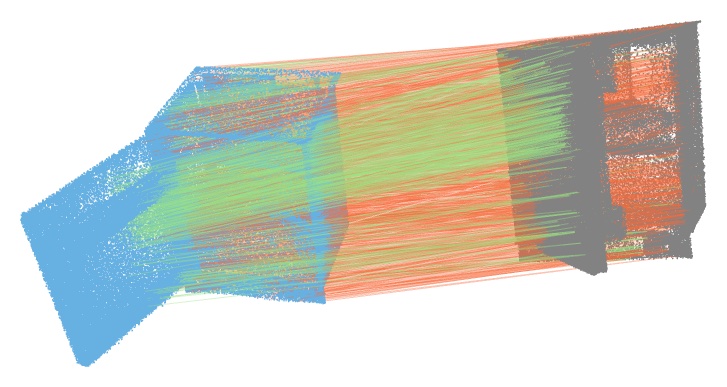}
            &\includegraphics[valign=c, width=0.3\textwidth]{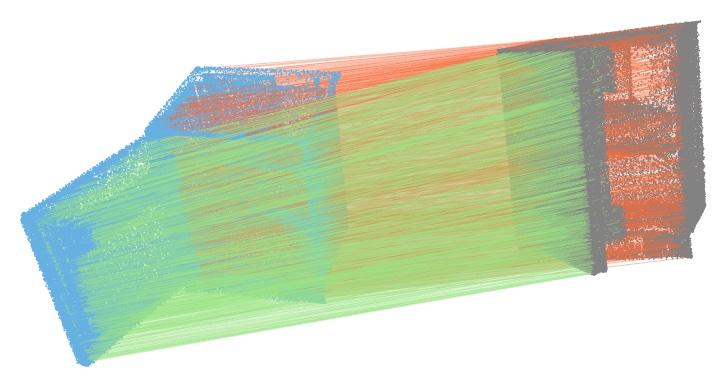}
            \\
            \includegraphics[valign=c, width=0.3\textwidth]{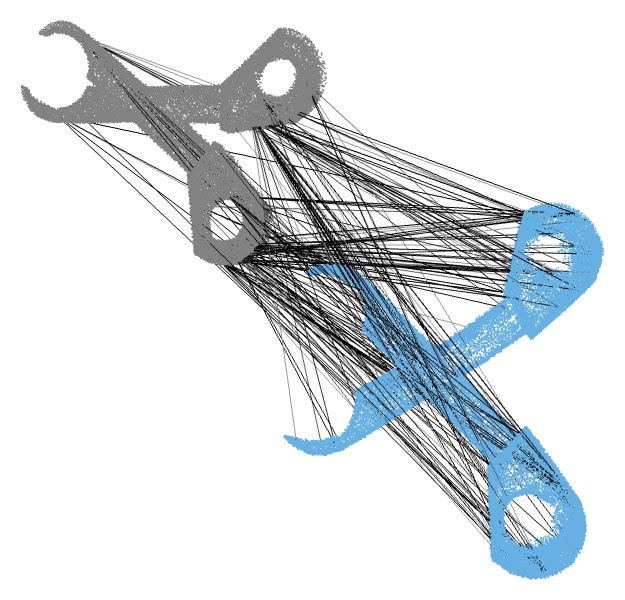}
            &\includegraphics[valign=c, width=0.3\textwidth]{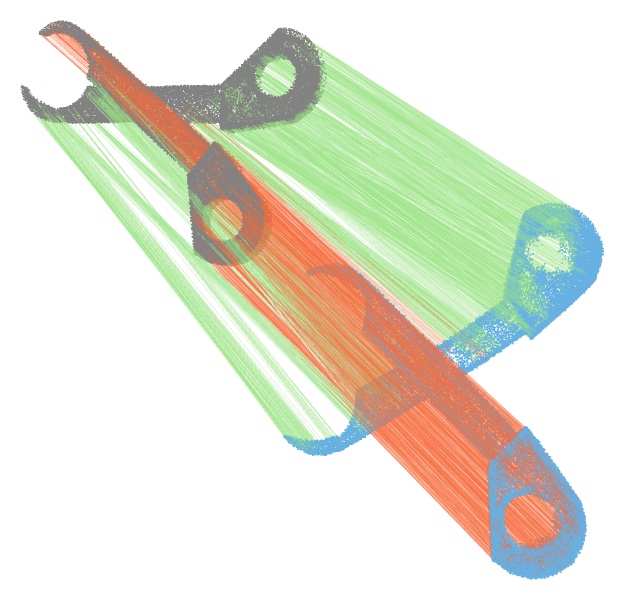}
            &\includegraphics[valign=c, width=0.3\textwidth]{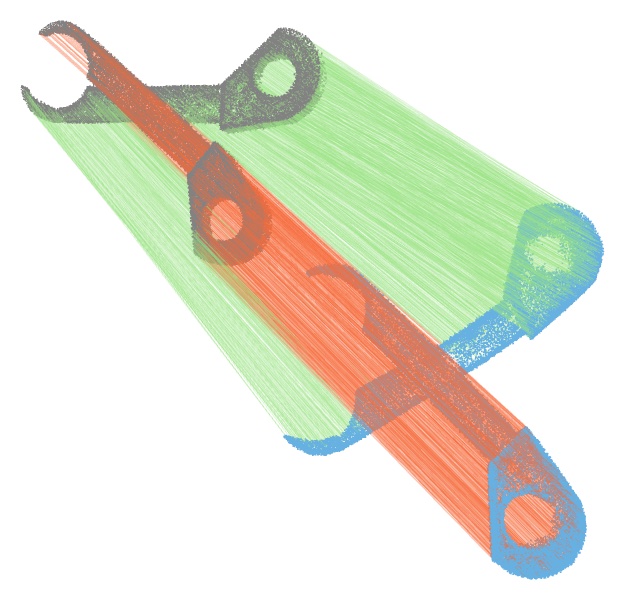}
            \\
            \includegraphics[valign=c, width=0.3\textwidth]{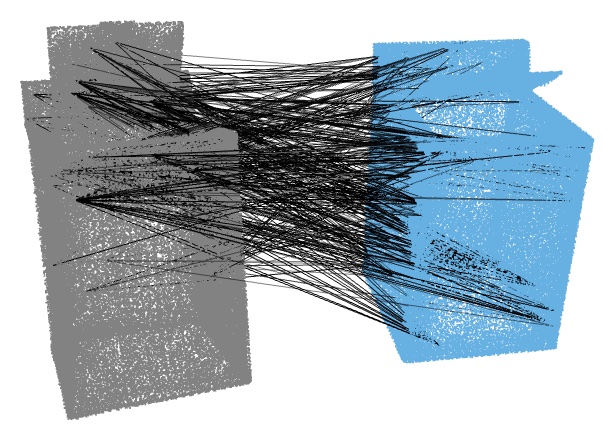}
            &\includegraphics[valign=c, width=0.3\textwidth]{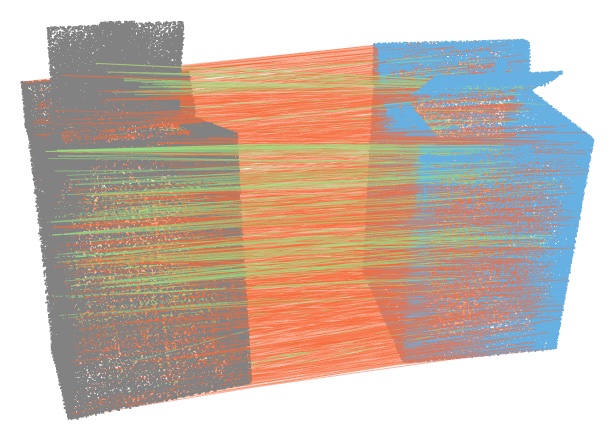}
            &\includegraphics[valign=c, width=0.3\textwidth]{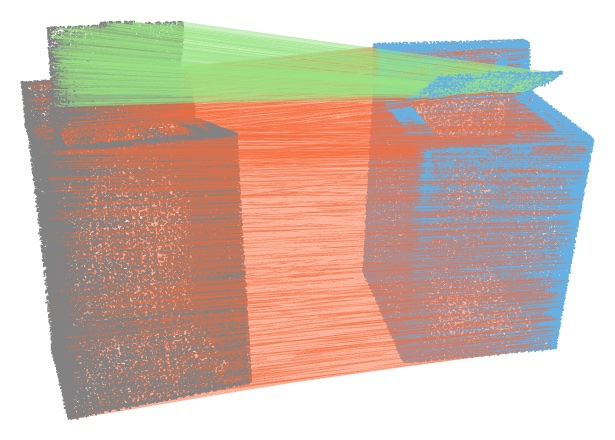}
            \\
        \end{tblr}
    }
    \caption{\textbf{Visualization of correspondences.} Inlier correspondences are shown as green and orange lines, while rejected or inconsistent correspondences are shown as black lines.}
    \label{fig:correspondences}
\end{figure}

\paragraph{Qualitative Evaluation}
\cref{fig:qual_corrs} qualitatively compares correspondence methods by visualizing their effect on downstream part-motion initialization. 
For each method, we feed its correspondences into the same TEASER~\cite{yang2020teaser} solver and visualize the resulting rigid inlier clusters. 
Inliers assigned to the two dominant parts are shown in blue and orange, while rejected correspondences and outliers are shown in gray. GaussReg~\cite{chang2024gaussreg} fails to produce usable part clusters: most points are rejected, indicating that its 1024 sparse keypoint correspondences are too noisy and insufficient for initializing part segmentation. 
FM-Dense~\cite{cao2023self} provides denser correspondences but remains unstable; most matches are rejected as outliers, and even when TEASER returns a solution, as in the laptop example, the resulting segmentation can assign both inlier clusters to the same physical part. 
In contrast, our deformation-based correspondence method produces coherent inlier clusters on distinct rigid parts, with only limited outliers near part boundaries. These residual errors are further refined during joint optimization.

\cref{fig:correspondences} further visualizes the dense correspondence maps produced by our method and the two baselines. Inlier correspondences are shown as green and orange lines, while rejected or inconsistent correspondences are shown as black lines. Compared with sparse keypoint matching and pretrained dense matching, our method produces more spatially coherent correspondences across articulation states, leading to more reliable part segmentation and motion initialization.

\begin{figure}[t]
    \centering
    \SetTblrInner{rowsep=0pt,colsep=0pt}
    {
    \begin{tblr}{*{3}{Q[c,m,wd=0.33\linewidth]}}

    \includegraphics[width=\linewidth]{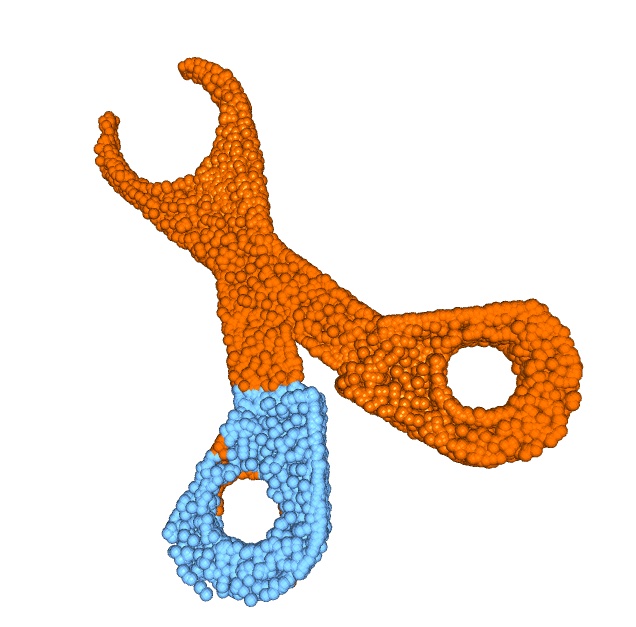}
    &\includegraphics[width=\linewidth]{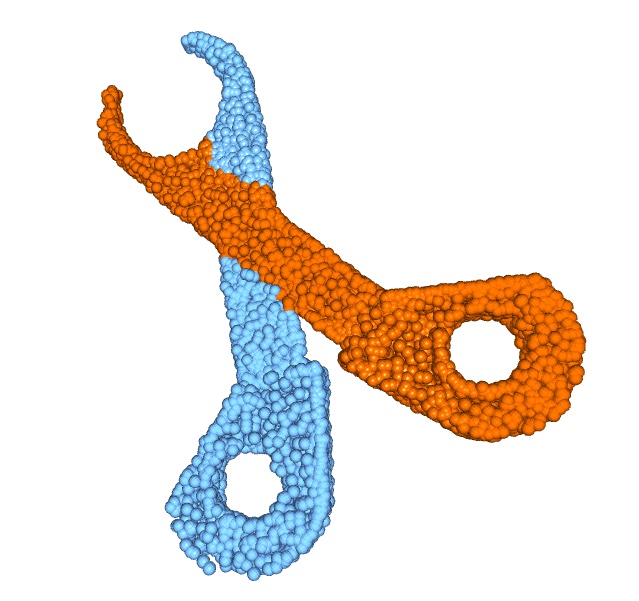}
    &\includegraphics[width=\linewidth]{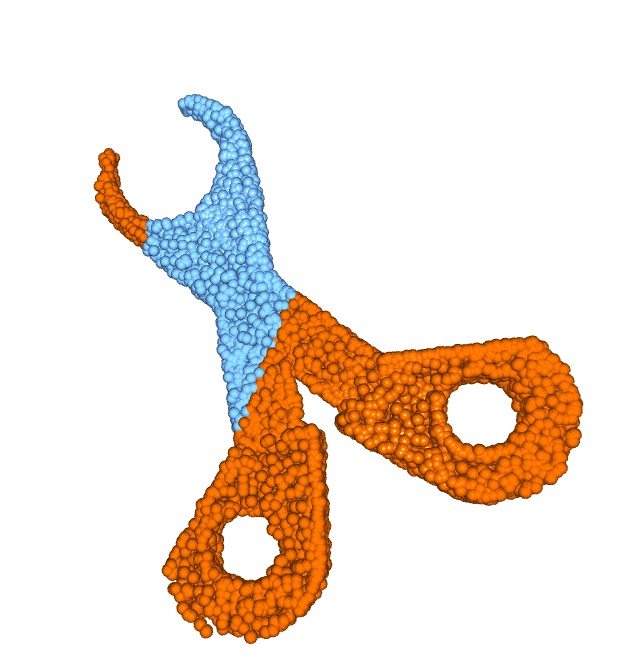}\\
    \includegraphics[width=\linewidth]{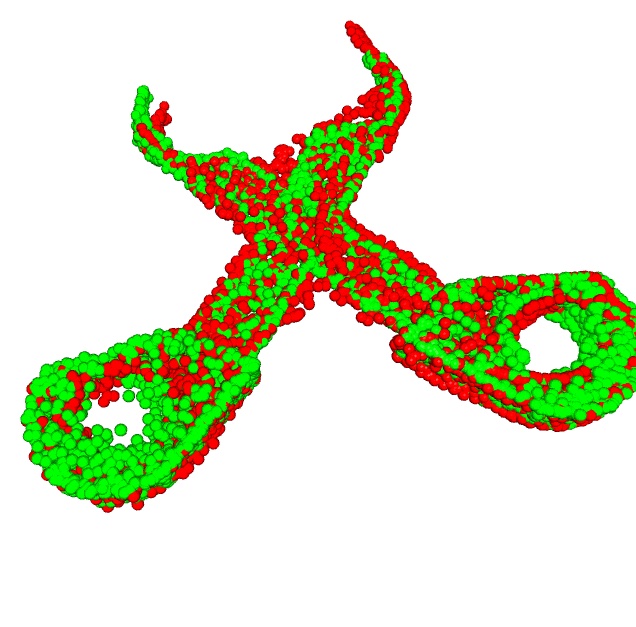}
    &\includegraphics[width=\linewidth]{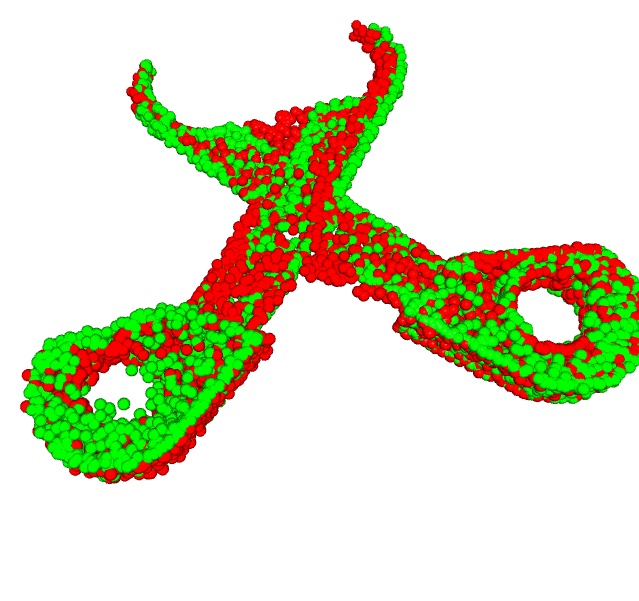}
    &\includegraphics[width=\linewidth]{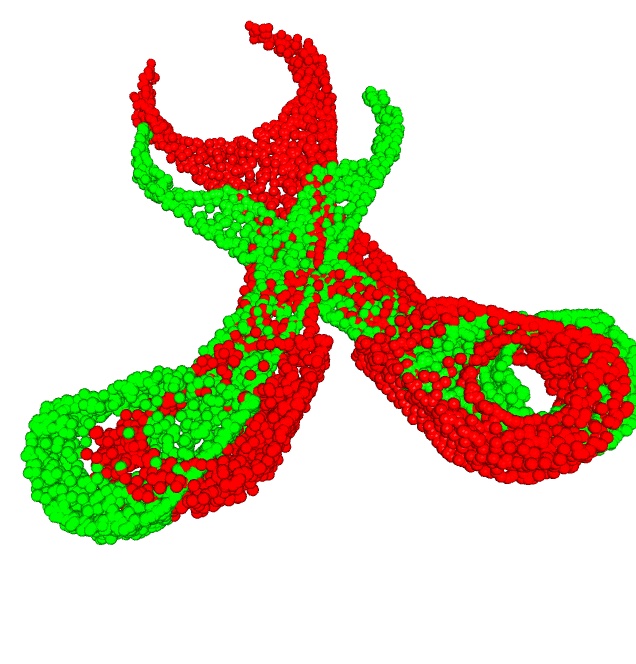}\\
    $k_0 = -5$ & $k_0 = -8$ & $k_0 = -11$
    \end{tblr}}
    \caption{\textbf{Ablation on $k_0$.} First row: coarse segmentation, different colors for different parts. Second row: deformation overlay, red for deformed $\hat{G}^t$, green for target $G^t$.}
    \label{fig:ablation_k0}
\end{figure}

\cref{fig:segmentation} visualizes the final part segmentation and estimated joint axes. 
For reference, we include AGS~\cite{guo2025articulatedgs} in its standard dense-view (i.e., 100 views for state), posed-image setting as a strong baseline, and denote it as AGS-FULL. 
Despite using substantially sparser and unposed inputs, our method produces segmentation and joint-axis estimates that are visually comparable to AGS. 
In contrast, the FM-Dense~\cite{cao2023self} baseline shows clear degradation, with errors in both part segmentation and joint estimation. 
These results demonstrate that our full pipeline can recover accurate part structure and kinematics from sparse, unposed observations. 
We further show qualitative results for novel articulation synthesis in \cref{fig:novel_articulation_rendering}, with additional visualizations provided in the supplementary material.

\subsection{Analysis and Ablations}

\paragraph{Varying $k_0$}
\Cref{fig:ablation_k0} qualitatively analyzes the effect of the deformation-field frequency parameter $k_0$ defined in \cref{eq:pe}. 
Large values of $k_0$ (e.g., $-5$) make the deformation field overly flexible, leading to over-deformation and poor part separation. 
In contrast, very small values (e.g., $-11$) bias the deformation toward nearly global rigid motion, causing under-deformation and preventing the source reconstruction from aligning with the target state. 
Both extremes degrade the initial part segmentation and downstream optimization. 
Empirically, we find that values between $-7$ and $-9$ provide a good balance between rigidity and flexibility, enabling coherent alignment while preserving part-level structure.


\paragraph{Ablation on key pipeline components}

\begin{table*}[t]
    \centering
    \begin{tblr}{
        colspec={ccc|cccccc},
        row{1} = {font=\bfseries},
        row{2} = {font=\bfseries},
    }
        \hline
        \SetCell[c=3]{c} Components & & & \SetCell[c=6]{c} Metrics & & & & & \\
        \cline{1-3} \cline{4-9}
        cam\_opt & $\mathcal{L}_{\text{CD}}$ & $\mathcal{L}_{\text{photo}}$ & Ang Err $\downarrow$ & Pos Err $\downarrow$ & Motion Dist $\downarrow$ & PSNR $\uparrow$& SSIM $\uparrow$  & CD $\downarrow$ \\
        \hline
        $\checkmark$ & $\checkmark$ & $\checkmark$ & 1.160 & 0.105 & 1.318& 26.573 & 0.953 & 2.272\\
        \hline
        $\color{red}\times$ & $\checkmark$ & $\checkmark$ & 1.858 & 0.581 & 2.175 & 24.101 & 0.880 & 3.323\\
        $\checkmark$ & $\color{red}\times$ & $\checkmark$ & 1.341 & 0.270 & 1.649 & 25.489 & 0.912 & 2.855 \\
        $\checkmark$ & $\checkmark$ & $\color{red}\times$ & 1.635 & 0.315 & 1.947 & 24.678 & 0.905 & 3.042\\
        \hline
    \end{tblr}
    \caption{Ablation study on the key components of our pipeline. We report the average metrics across all objects.}
    \label{tab:ablation_components}
\end{table*}

\begin{table}[t]
    \centering
    \SetTblrInner{rowsep=0pt,colsep=1pt}
    \begin{tblr}{*{5}{Q[c,m,wd=0.19\linewidth]}}
    \hline
    \textbf{Pert. level} &\textbf{1\%} & \textbf{2\%} & \textbf{5\%} & \textbf{10\%}\\
    \hline
    \textbf{RGB}
    &\includegraphics[valign=c, width=\linewidth]{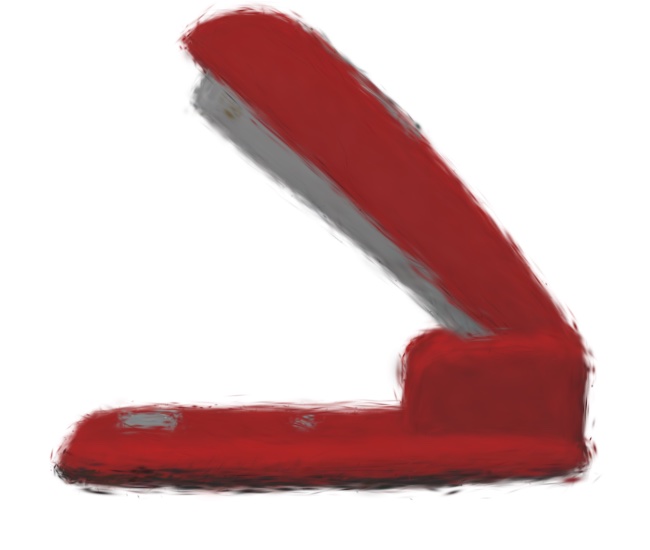}
    & \includegraphics[valign=c, width=\linewidth]{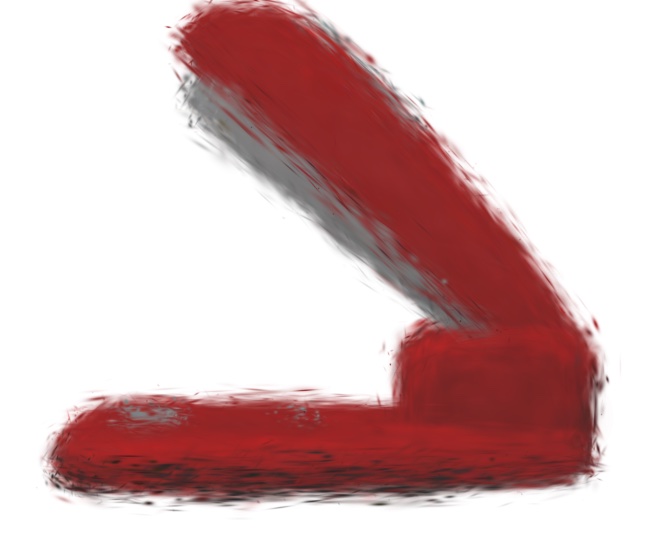}
    & \includegraphics[valign=c, width=\linewidth]{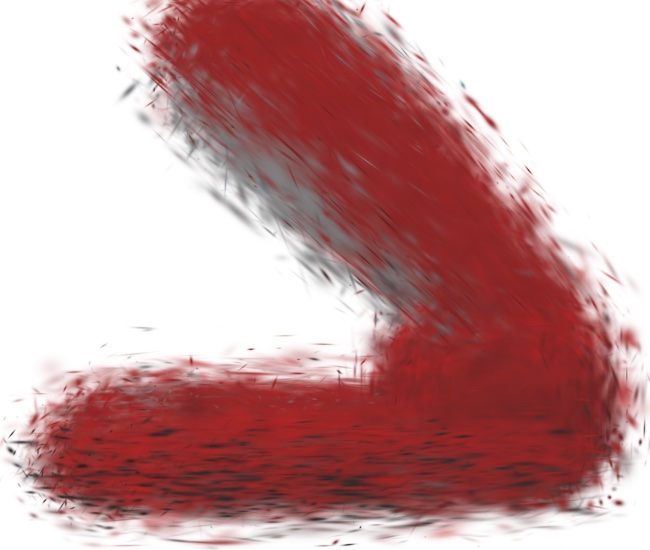}
    & \includegraphics[valign=c, width=\linewidth]{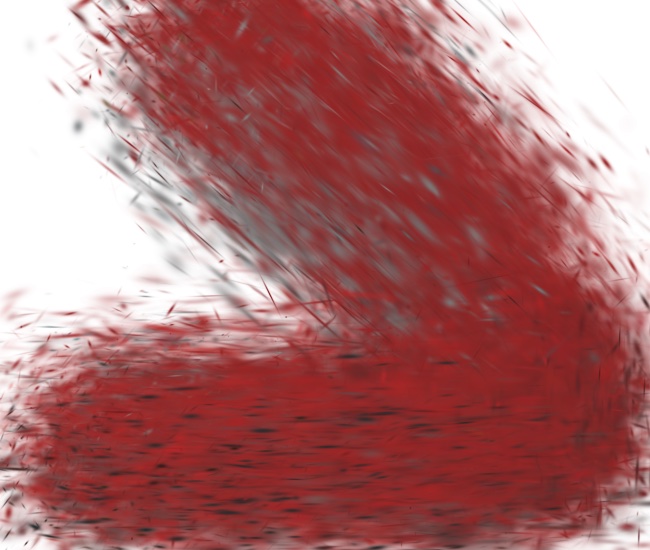}\\
    \textbf{Seg.}
    &\includegraphics[valign=c, width=\linewidth]{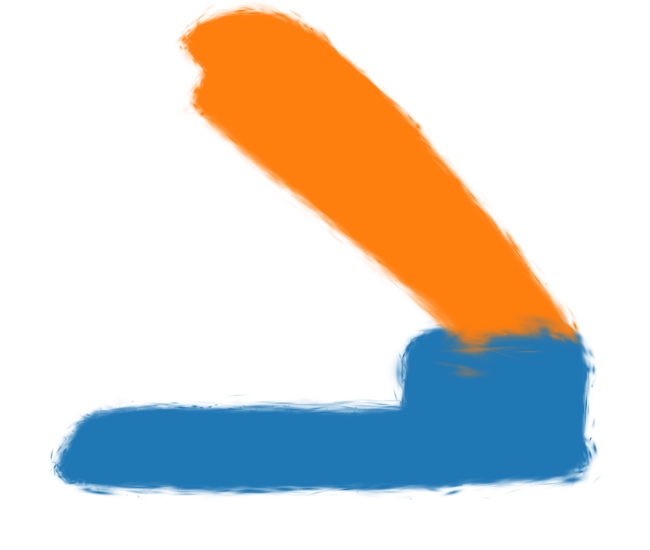}
    & \includegraphics[valign=c, width=\linewidth]{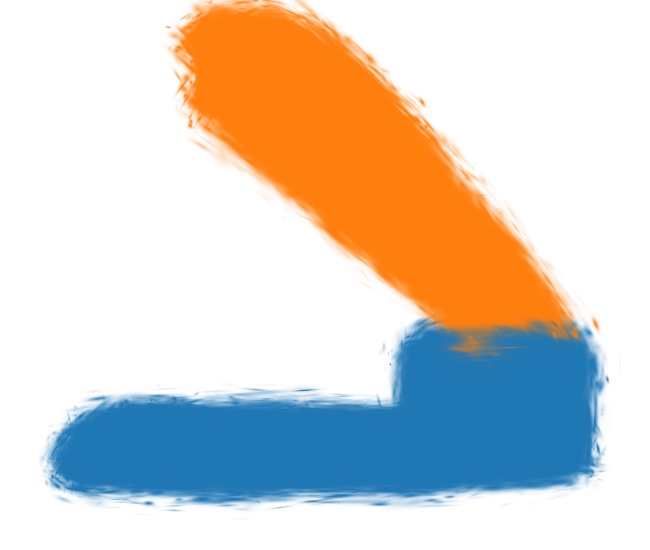}
    & \includegraphics[valign=c, width=\linewidth]{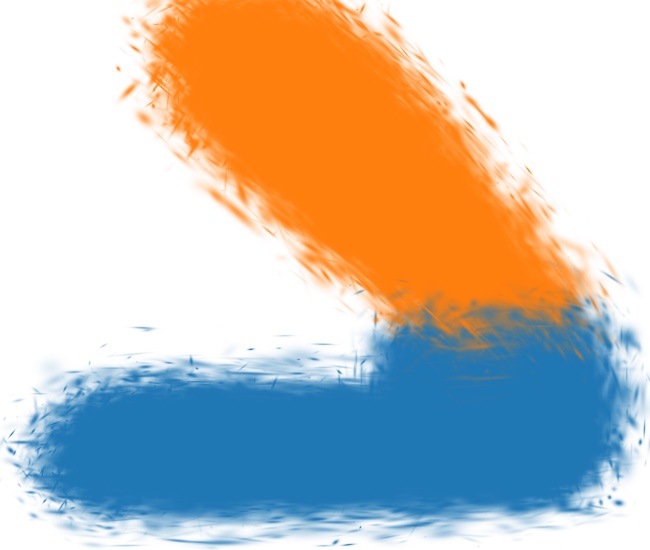}
    & \includegraphics[valign=c, width=\linewidth]{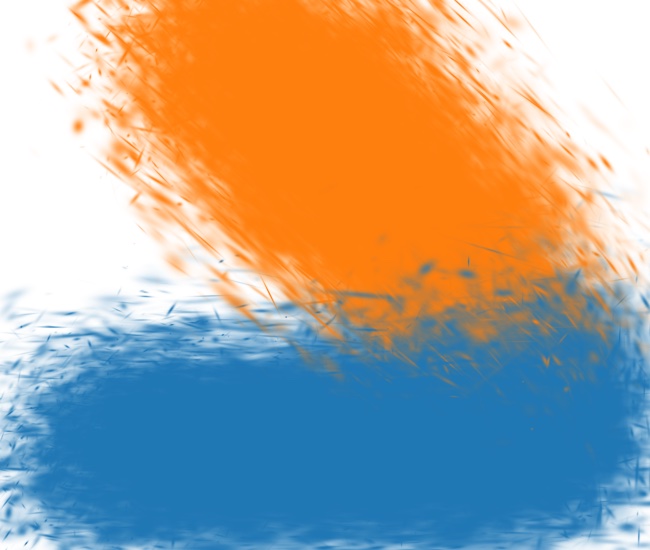}\\
    \textbf{mIoU} & $0.888$ & $0.779$ & $0.545$ & $0.345$\\
    \hline
    \end{tblr}
    \caption{Sensitivity Analysis: perturbation level vs.\ mIoU.}
    \label{tab:ablation_sensitivity}
\end{table}

We conducted an ablation study to analyze the impact of key components in our pipeline: camera pose refinement in step 1 and the losses used for deformation network optimization ($\mathcal{L}_{\text{CD}}$ and $\mathcal{L}_{\text{photo}}$) in step 2. As shown in \cref{tab:ablation_components}, removing any of these components leads to a degradation in performance. These components are crucial for initializing the part segmentation and pose estimation in Step 3. Omitting these components leads to poorer initializations, which subsequently hampers the final joint optimization stage. Further qualitative evidence can be found in the supplementary material.


\paragraph{Robustness to initialization quality}
\label{para:initialization_robustness}
We evaluate the sensitivity of our method to the quality of FreeSplatter's~\cite{xu2025freesplatter} initialization in \cref{tab:ablation_sensitivity}. Specifically, we perturb Gaussian point positions ($\bm{\mu}$) by amounts proportional to the object's bounding box. Our method is reasonably robust to the reconstruction noise, maintaining a mIoU of 0.545 even with a perturbation level of 5\%, where the object starts to loose its original shape.

\begin{figure}[t]
    \centering
    \includegraphics[width=\columnwidth]{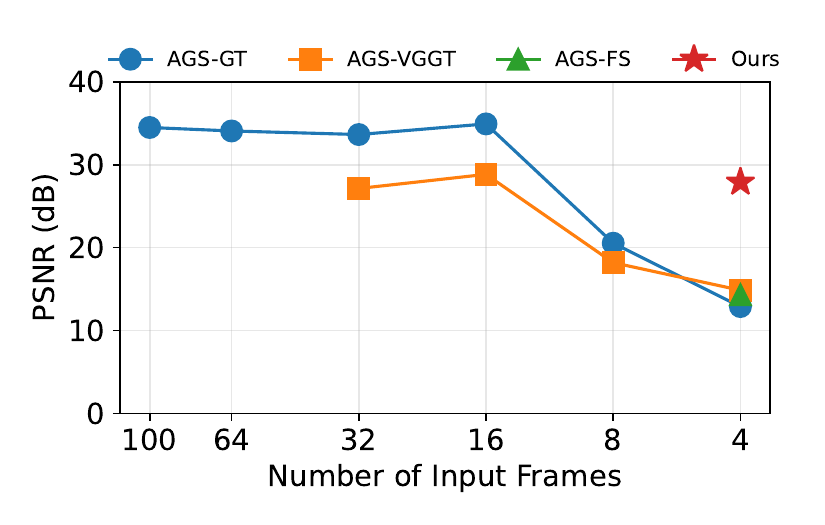}
    \caption{\textbf{PSNR \wrt number of views per articulation status.} Our method with 4-view input matches the performance of AGS-GT with 16 views.}
    \label{fig:ablation_frames}
\end{figure}

\begin{figure}[t]
    \centering
    \resizebox{\columnwidth}{!}{%
        \SetTblrInner{rowsep=1pt,colsep=1pt}
        \scriptsize
        \begin{tblr}{Q[c,wd=0.1\linewidth]|*{6}{Q[c,wd=0.14\linewidth]|}}
        &\SetCell[c=2]{c}{4-view} & & \SetCell[c=2]{c}{8-view} & & \SetCell[c=2]{c}{16-view} \\
        \hline
        AGS-GT
        &\includegraphics[valign=c, width=\linewidth]{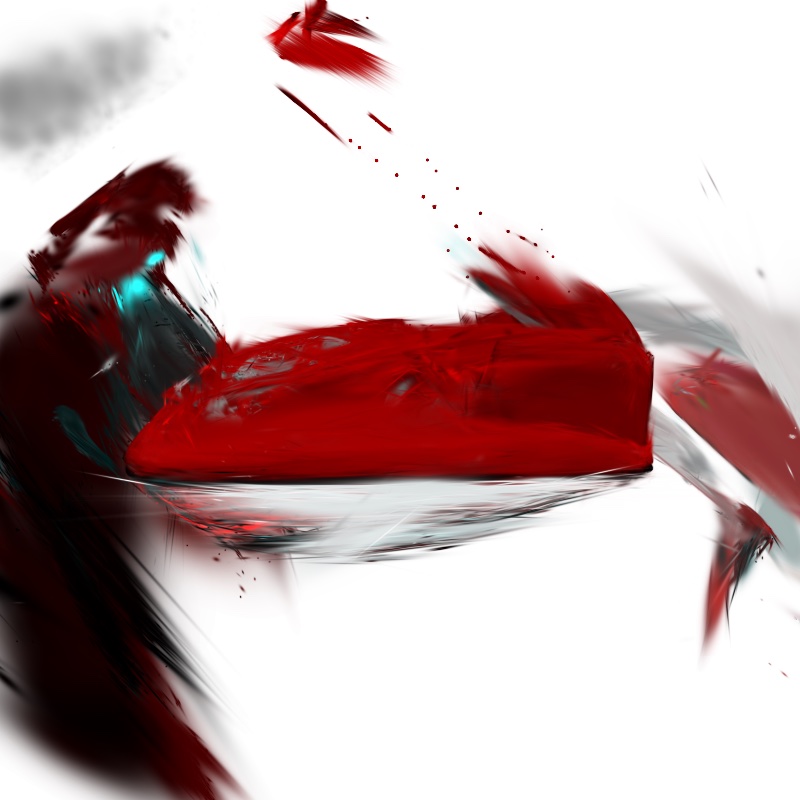}
        & \includegraphics[valign=c, width=\linewidth]{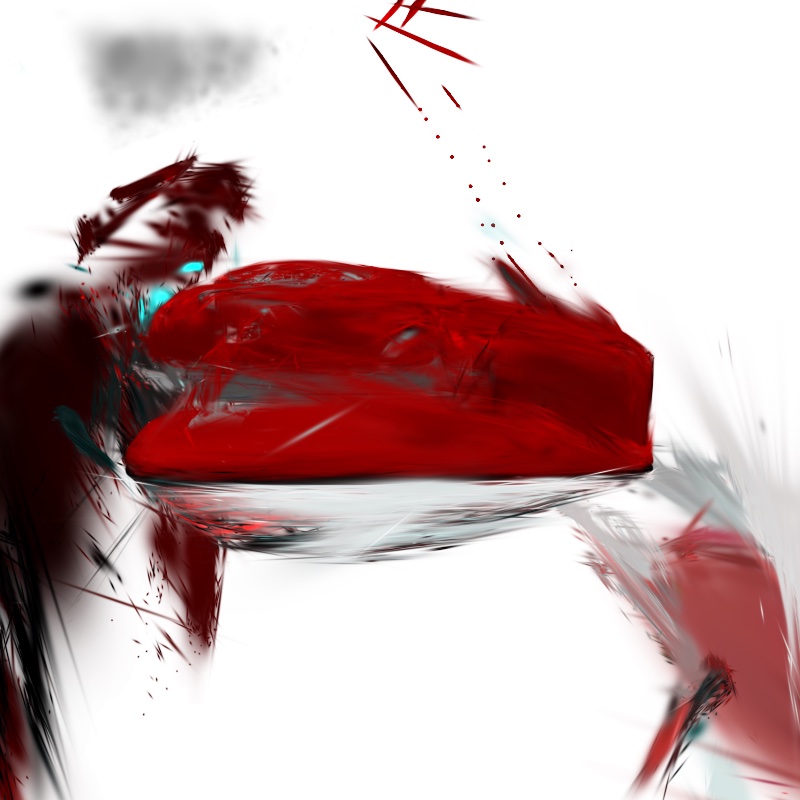}
        & \includegraphics[valign=c, width=\linewidth]{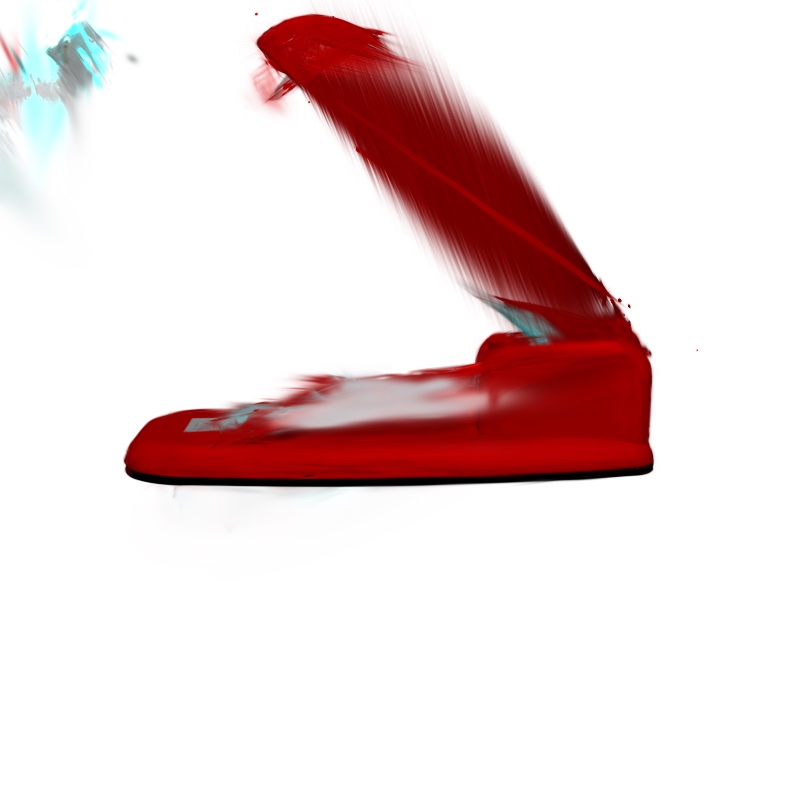}
        &\includegraphics[valign=c, width=\linewidth]{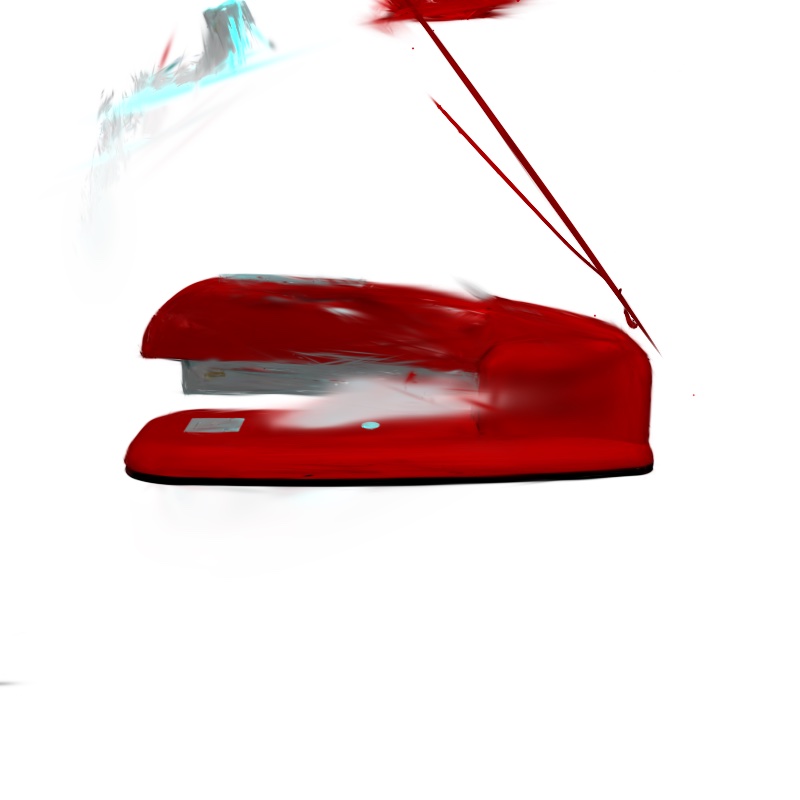}
        & \includegraphics[valign=c, width=\linewidth]{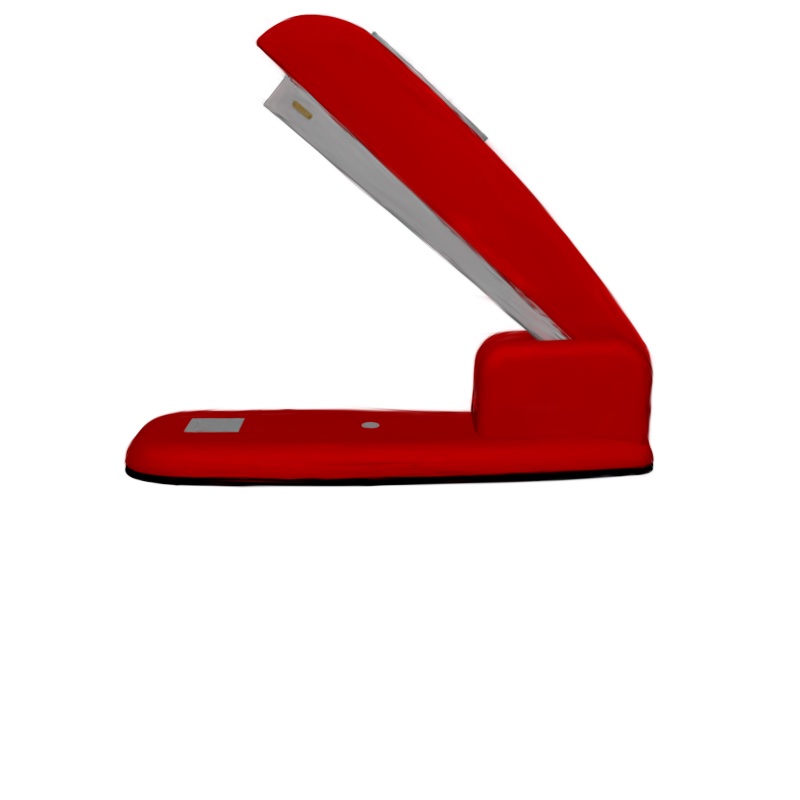}
        &\includegraphics[valign=c, width=\linewidth]{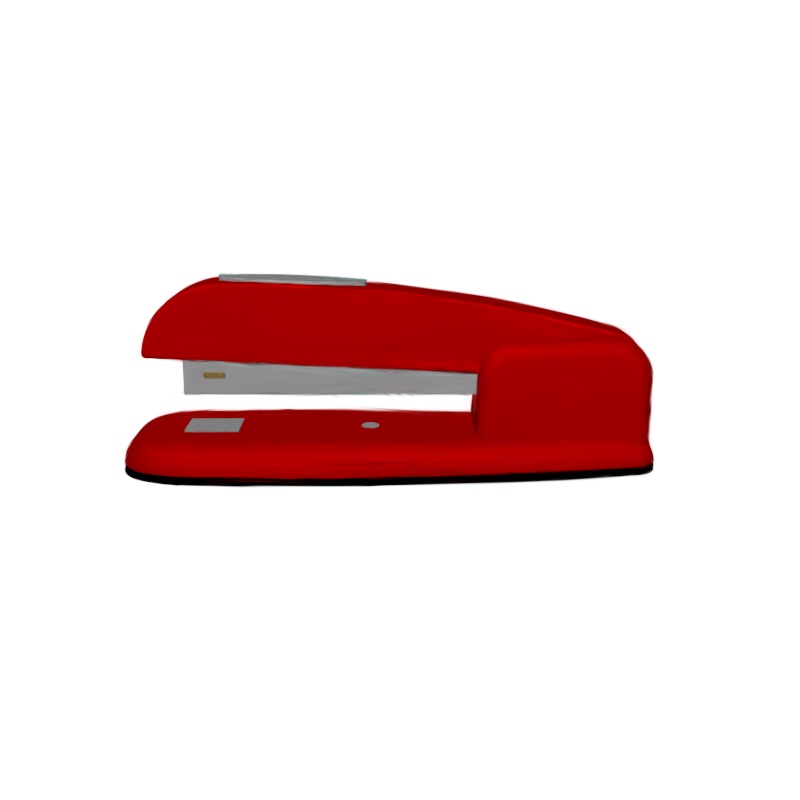} \\
        AGS-VGGT
        &\includegraphics[valign=c, width=\linewidth]{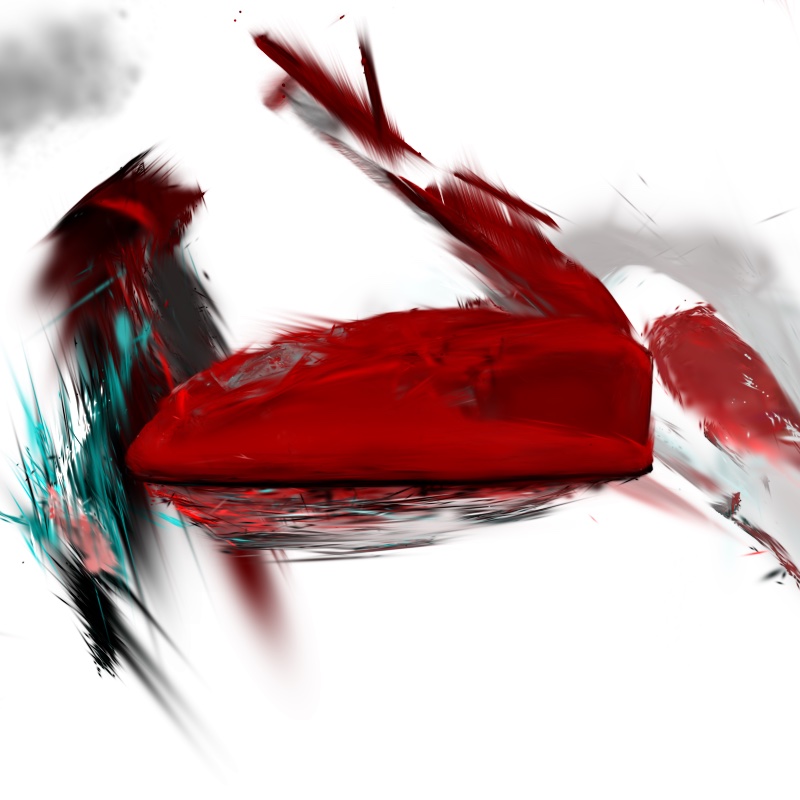}
        & \includegraphics[valign=c, width=\linewidth]{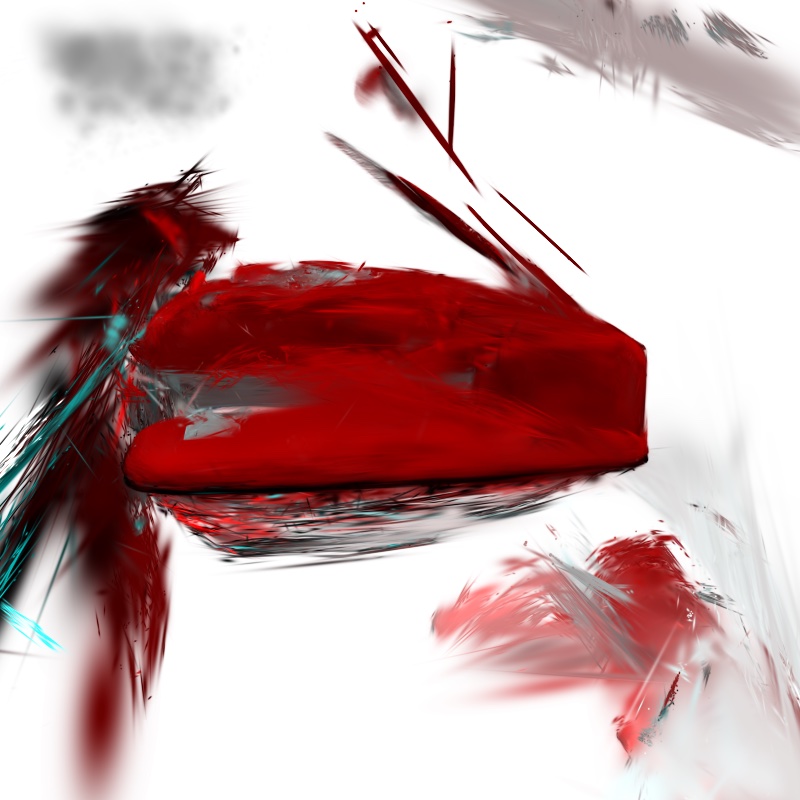}
        & \includegraphics[valign=c, width=\linewidth]{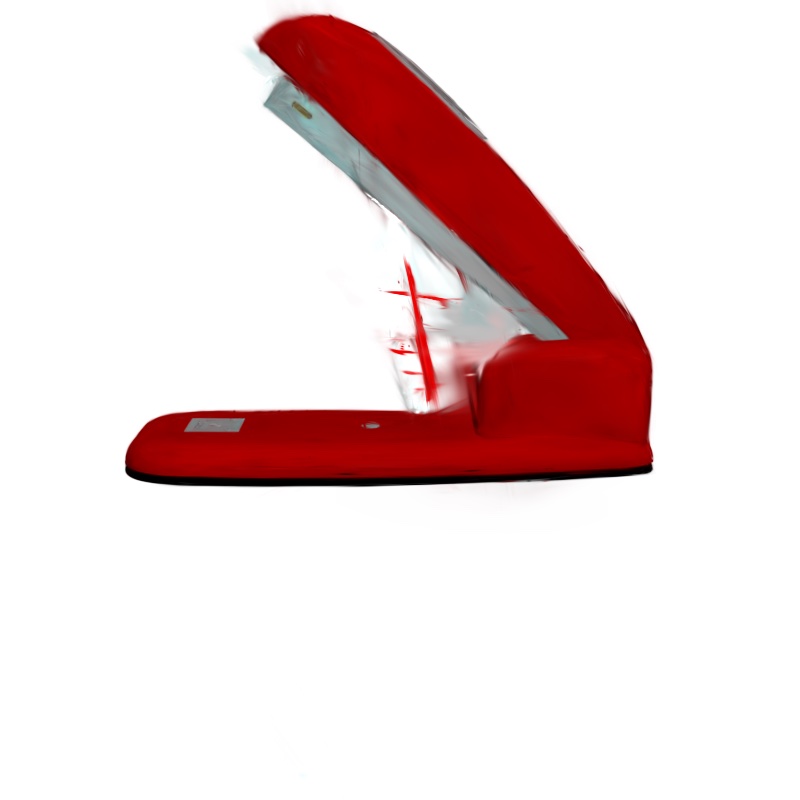}
        &\includegraphics[valign=c, width=\linewidth]{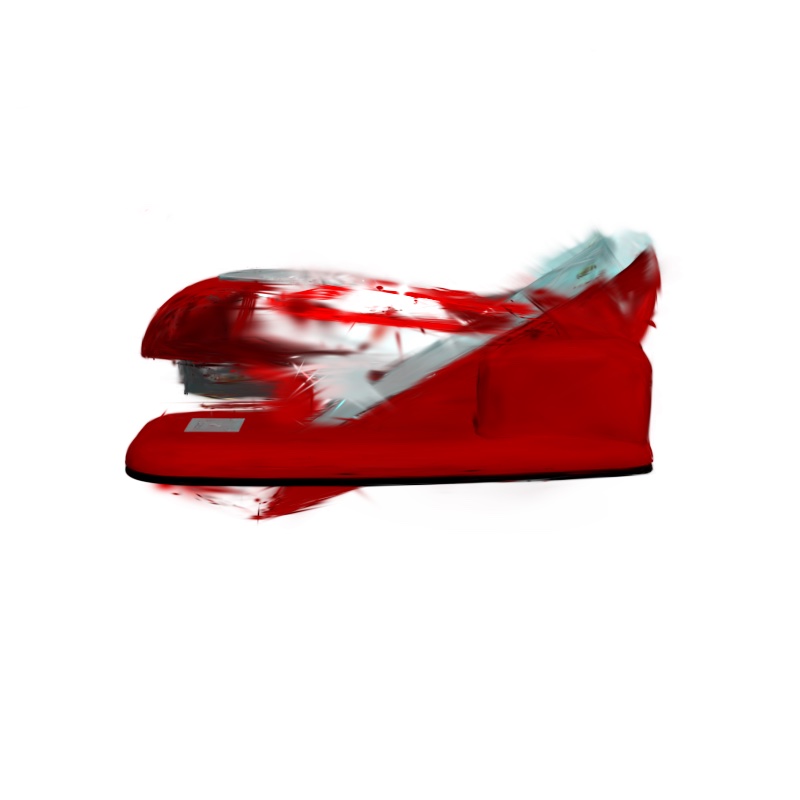}
        & \includegraphics[valign=c, width=\linewidth]{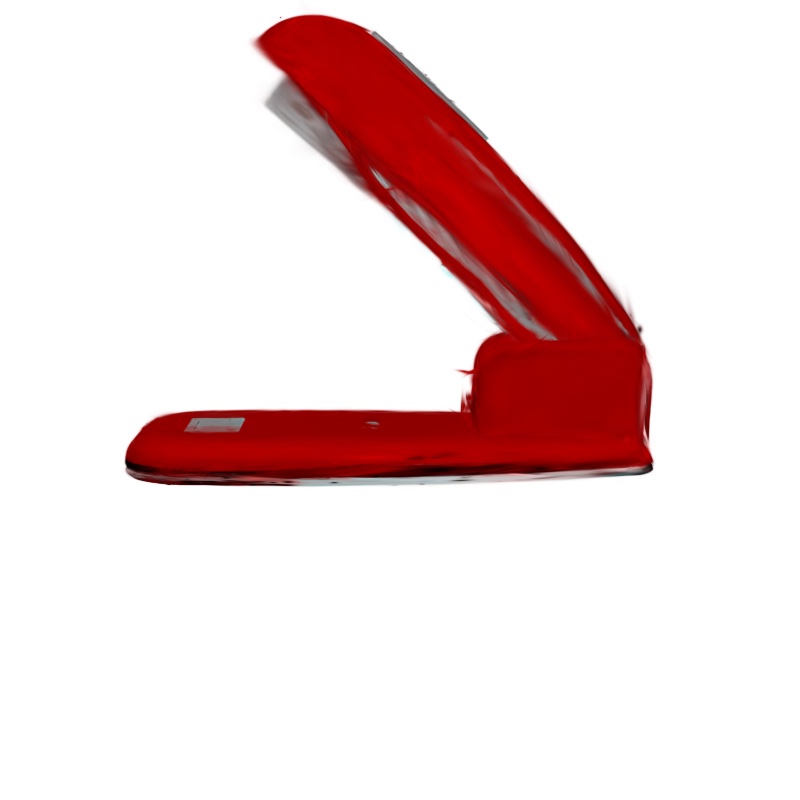}
        &\includegraphics[valign=c, width=\linewidth]{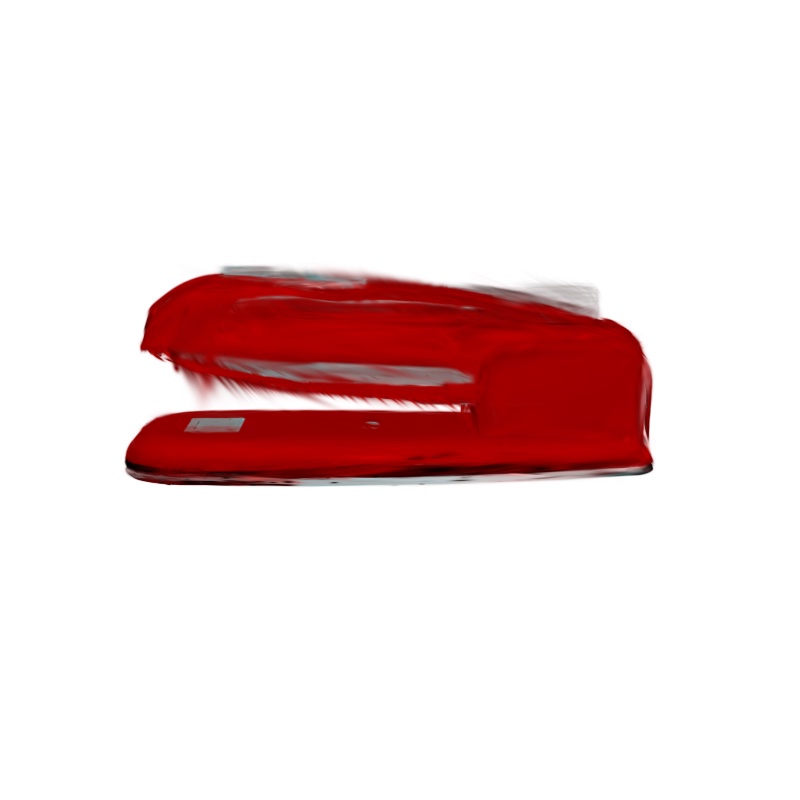} \\
        Ours
        &\includegraphics[valign=c, width=\linewidth]{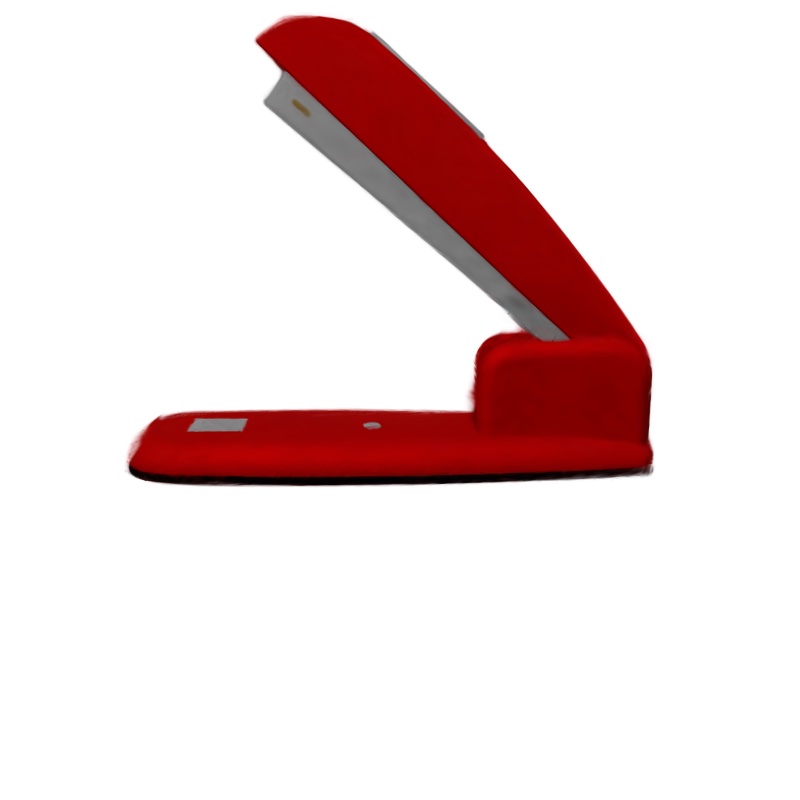}
        & \includegraphics[valign=c, width=\linewidth]{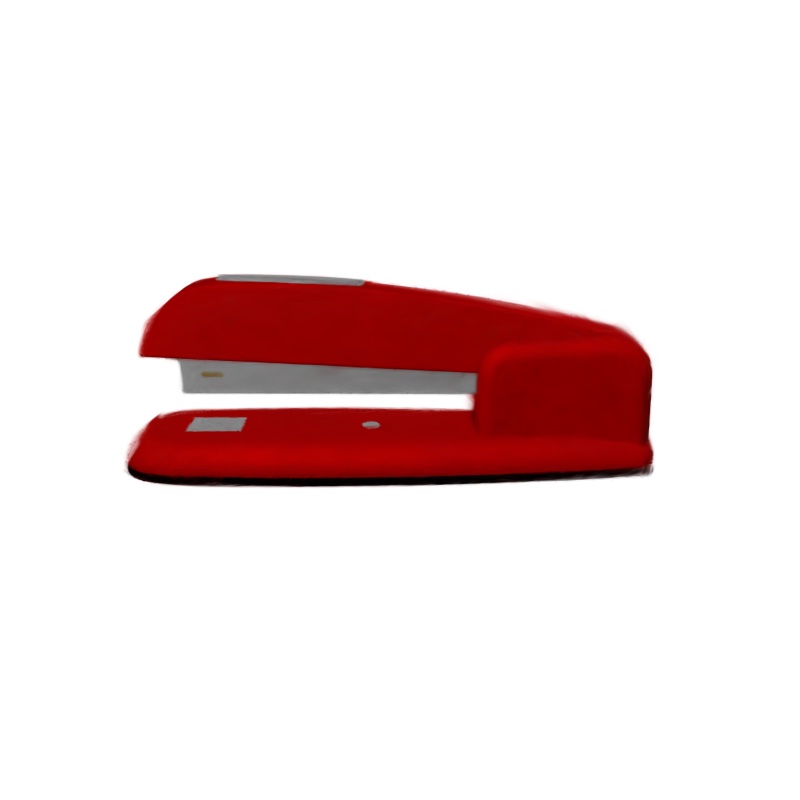}
        & \includegraphics[valign=c, width=\linewidth]{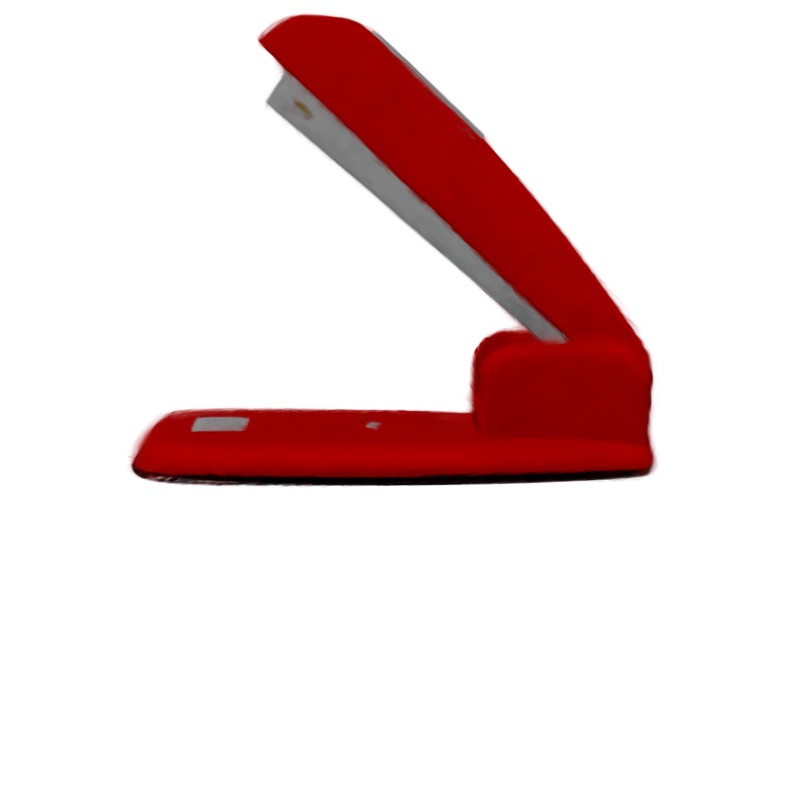}
        &\includegraphics[valign=c, width=\linewidth]{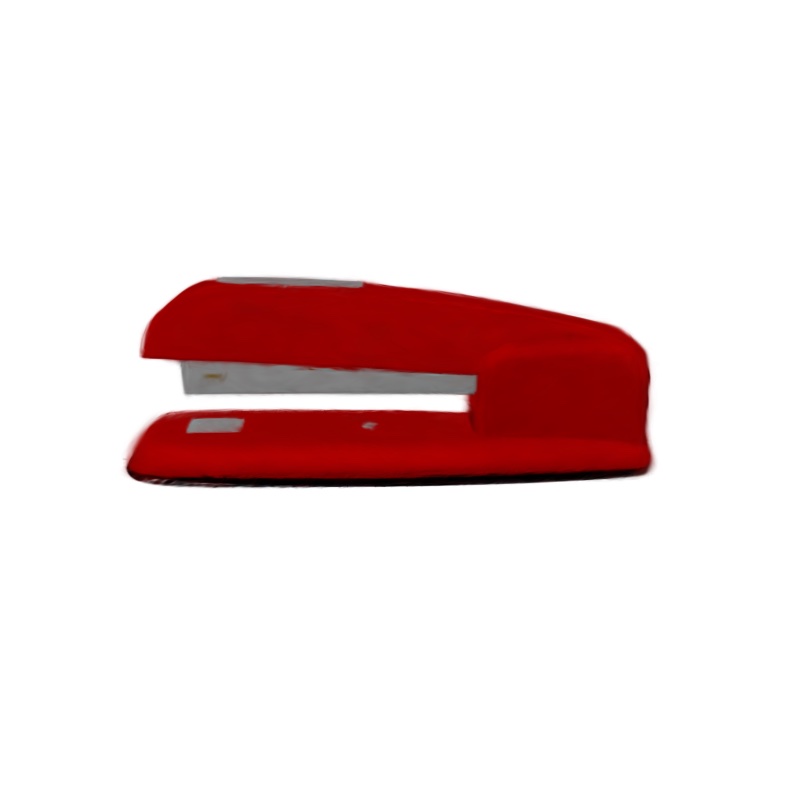}
        & \includegraphics[valign=c, width=\linewidth]{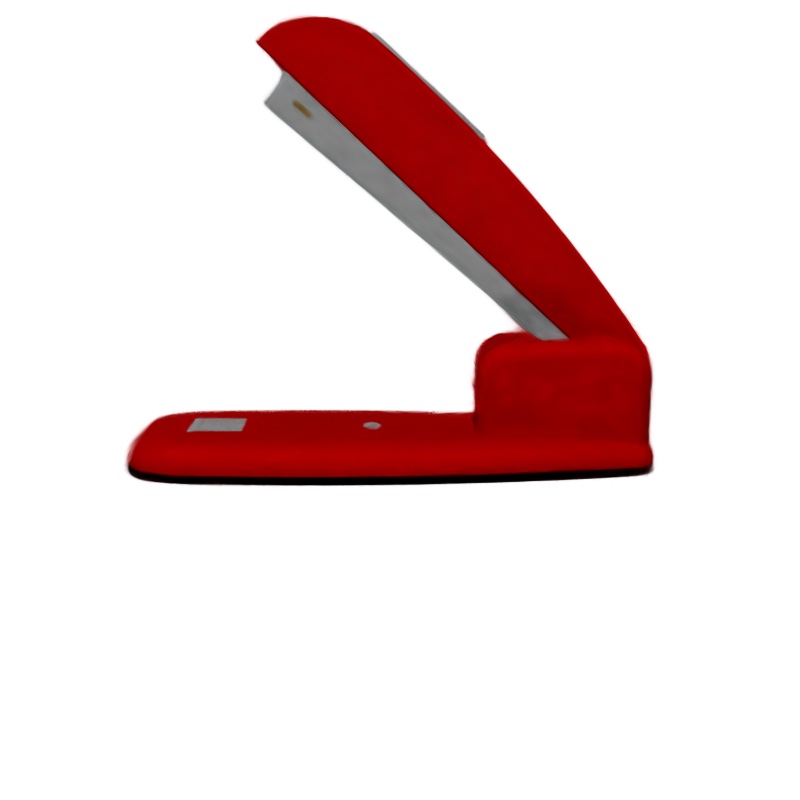}
        &\includegraphics[valign=c, width=\linewidth]{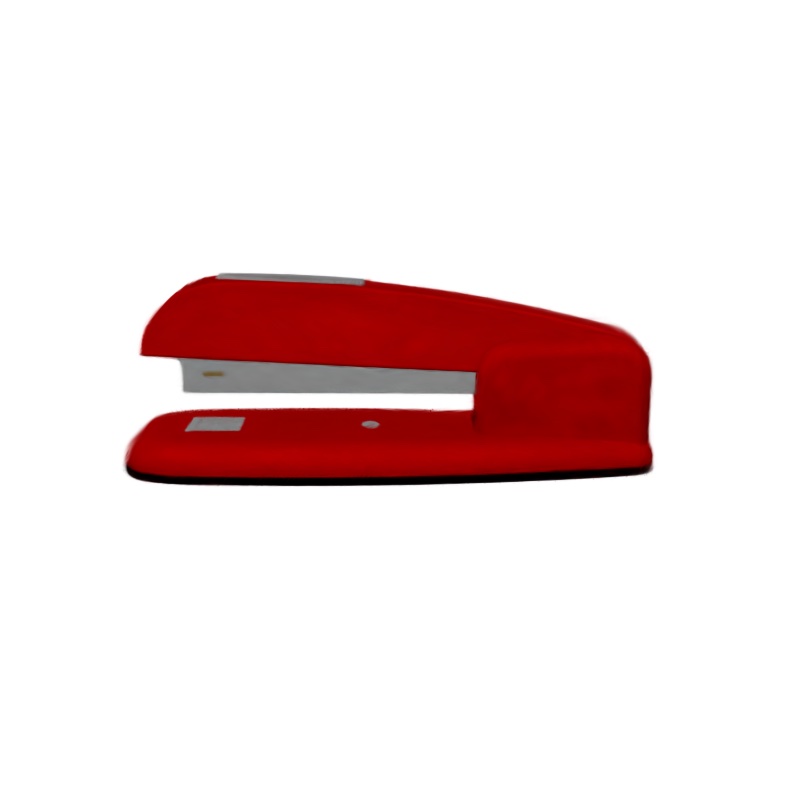}
        \end{tblr}
    }
    \caption{\textbf{Qualitative evaluation for the effect of the number of input images.} We show novel view synthesis for source state (left) and target state (right) at each settings. From top to bottom: AGS-GT, AGS-VGGT, Ours. }
    \label{fig:qualitative_num_view}
\end{figure}

\paragraph{Performance w.r.t. number of input views}
With just 4-view images, baseline methods struggle to reconstruct objects even when in static state. We investigate how the number of input views affects performance by varying the views per articulation state. \cref{fig:ablation_frames} shows that PSNR for baseline methods improves with more input images, saturating after 16 images (VGGT runs out of memory with 64 images, 32 per status on our 48GB VRAM GPU), even with VGGT-estimated poses. Our method matches the saturated performance without ground truth poses with only 4 input images and stays consistent across different view counts as shown in \cref{fig:qualitative_num_view}.
\vspace{1cm}

\paragraph{Failure Cases}

\begin{figure}[t]
    \centering
    \SetTblrInner{rowsep=0pt,colsep=0pt}
    {
    \begin{tblr}{*{4}{Q[c,m,wd=0.25\linewidth]}}
    
    Source view &Target view & GT & Prediction \\
    \includegraphics[width=\linewidth]{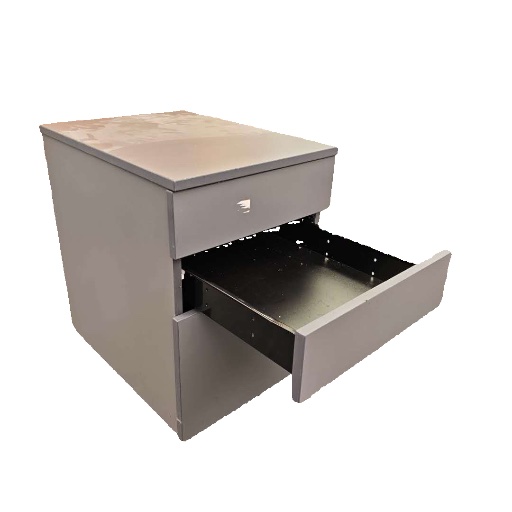}
    &\includegraphics[width=\linewidth]{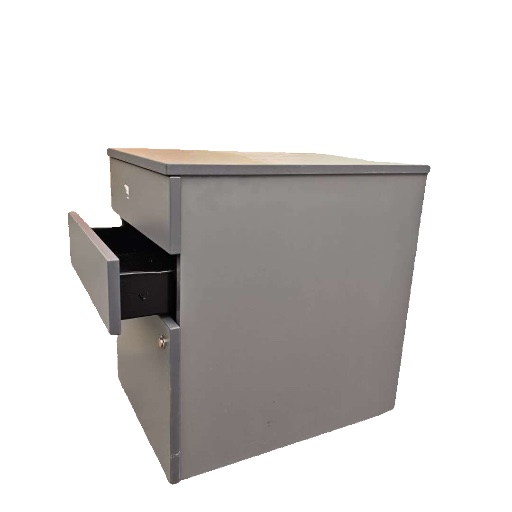}
    &\includegraphics[width=\linewidth]{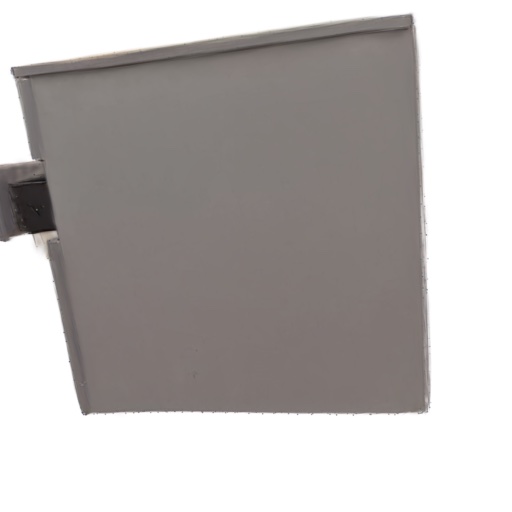}
    &\includegraphics[width=\linewidth]{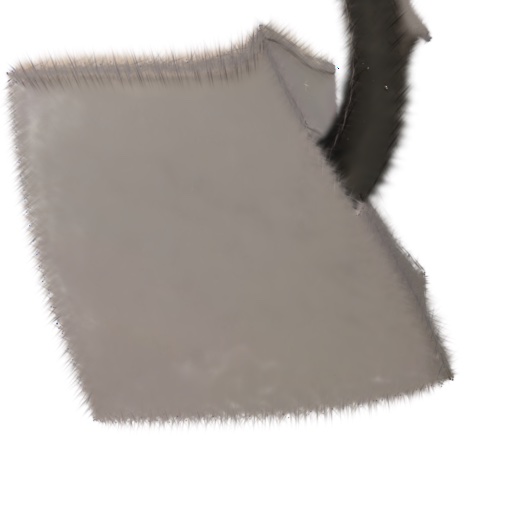}\\
    \includegraphics[width=\linewidth]{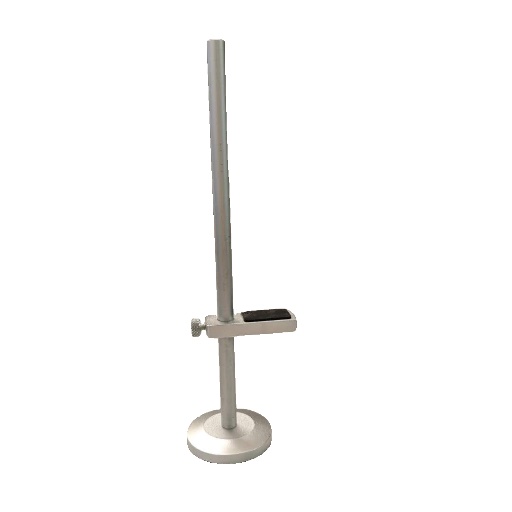}
    &\includegraphics[width=\linewidth]{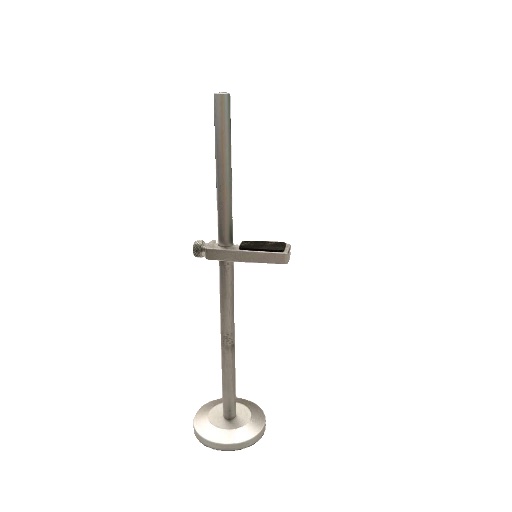}
    &\includegraphics[width=\linewidth]{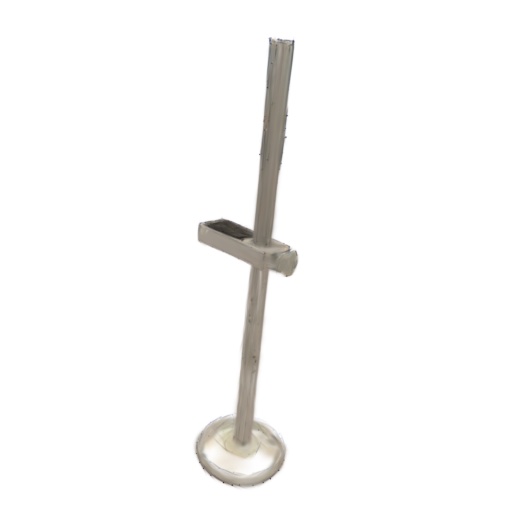}
    &\includegraphics[width=\linewidth]{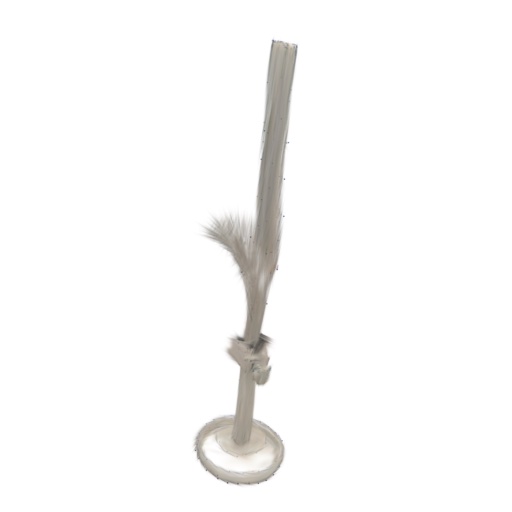}\\

    \end{tblr}
    }
    \caption{Visualization of the failure cases in extreme misalignment case (top) and thin structures with large motion (bottom). The GT and prediction are rendered at the same viewpoint. We can see that the pose estimation for storage is wrong and the deformation incorrectly bent the drawer. As for the rod with the holder, the holder part is also wrongly deformed and lost its shape.}
    \label{fig:supp_failure_cases}
\end{figure}

We show representative failure cases in \cref{fig:supp_failure_cases}. 
Our method relies on FreeSplatter~\cite{xu2025freesplatter} for initial 3D reconstruction, which establishes a coordinate system based on the first input image of a set. 

When the first images of the source and target sets are captured from drastically different viewpoints (e.g., front vs. back), the resulting 3D models can be severely misaligned. 
This large misalignment can cause our deformation network to converge to a poor local minimum, especially for objects with geometric symmetries. 
For instance, in the real-world storage-drawer case shown in \cref{fig:supp_failure_cases}, using opposing side views as initial frames causes the network to favor a non-rigid deformation of the drawer rather than finding the correct global alignment. To mitigate this, we recommend selecting initial views from similar perspectives for each image set.

\begin{figure}[t]
    \centering
    \SetTblrInner{rowsep=1pt,colsep=1pt}
    \small
    \begin{tblr}{*{4}{Q[c,m,wd=0.25\linewidth]}}
    TEASER init. & Phase 1 & Phase 2 & Phase 3 \\
    \hline
    \includegraphics[valign=c, width=\linewidth]{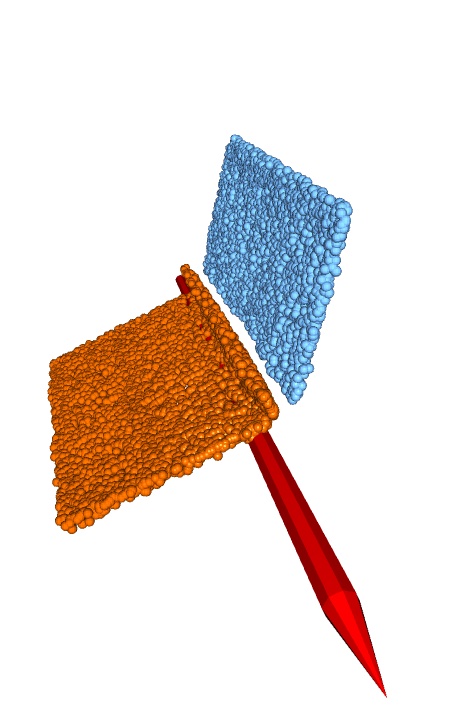}
    &\includegraphics[valign=c, width=\linewidth]{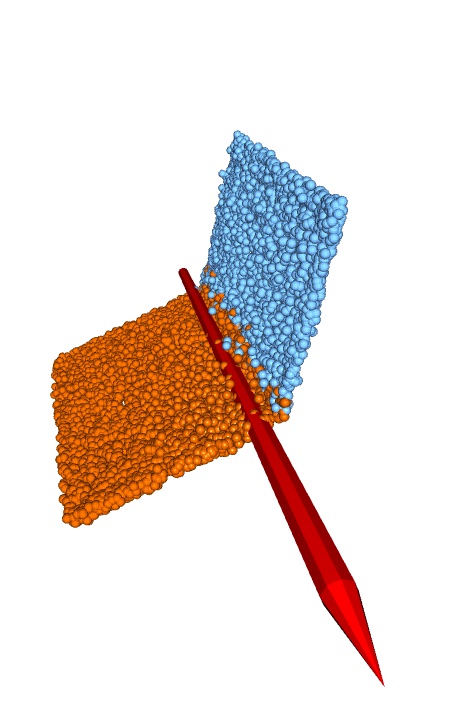} 
    & \includegraphics[valign=c, width=\linewidth]{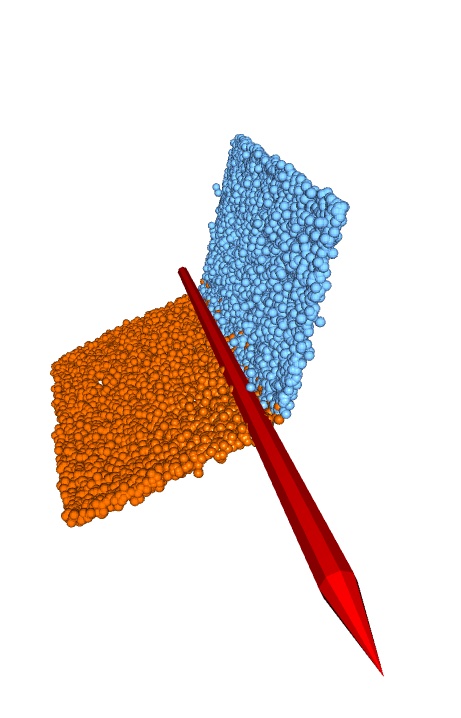}
    &\includegraphics[valign=c, width=\linewidth]{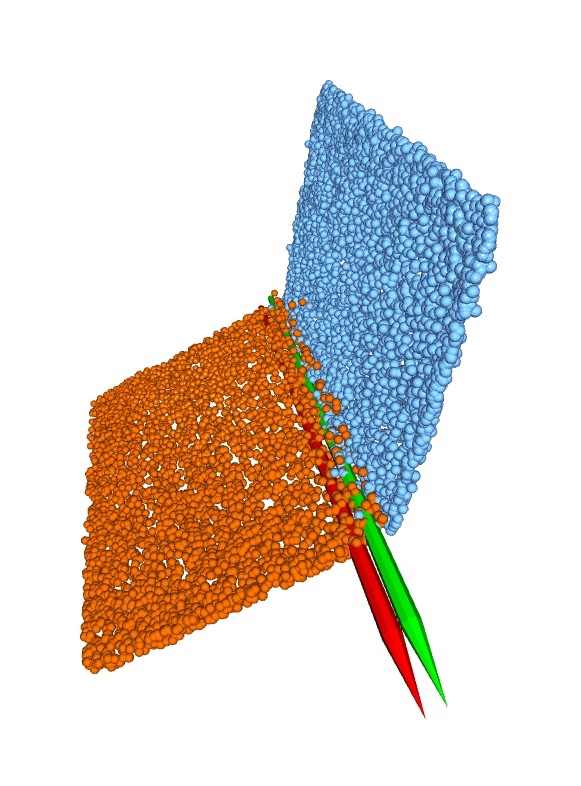} 

    \end{tblr}
    \caption{Visualization of the joint optimization process, showing the progression from the initial state to the final refined output for the laptop object. The lid in \textcolor{cyan}{cyan} is detached and incomplete after TEASER initialization. And its position and shape gets refined after each phase. Better view in color and zoom in.}
    \label{fig:supp_joint_opt}
\end{figure}

\paragraph{Effectiveness of the Joint Optimization}
The effectiveness of our joint optimization process is demonstrated in \cref{fig:supp_joint_opt}. The figure shows how the process corrects initial errors: despite dense matching, the state after TEASER initialization still has spatial discrepancies between parts. The lid in cyan is incompleted and detached from the keyboard of the laptop in orange. Phase 1 recovers connectivity between the lid and the keyboard, Phase 2 refines part segmentation, and Phase 3 produces the final, rigid, joint-optimized output.

\paragraph{Computation Efficiency}

Our pipeline takes $\sim$15 minutes per object on an NVIDIA RTX 4090 GPU ($\sim$1\,min for reconstruction, $\sim$4\,min for deformation, $\sim$10\,min for joint refinement), with a peak VRAM usage of 15\,GB.

\subsection{Real-world Experiments} 
\begin{figure}[h!]
    \centering
    \resizebox{0.95\linewidth}{!}{%
    \SetTblrInner{rowsep=0pt,colsep=1pt}
    \scriptsize
    \begin{tblr}{cc|ccccc|c}
    \SetCell[c=2]{c}{RGB input} & & \SetCell[c=5]{c}{Novel articulation synthesis} & &  & & & Part-level segmentation
        \\
    \hline
    \includegraphics[valign=c, width=0.1\textwidth]{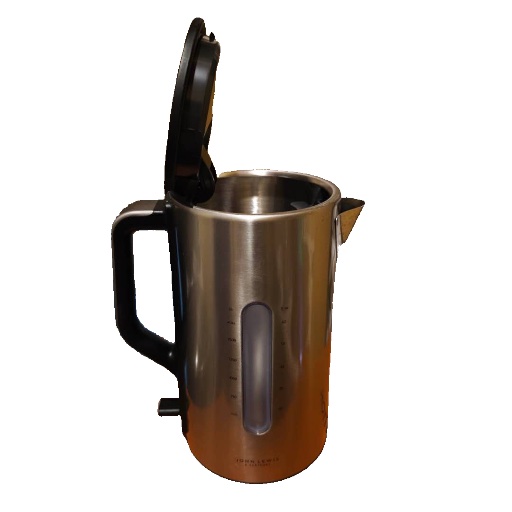} 
    &\includegraphics[valign=c, width=0.1\textwidth]{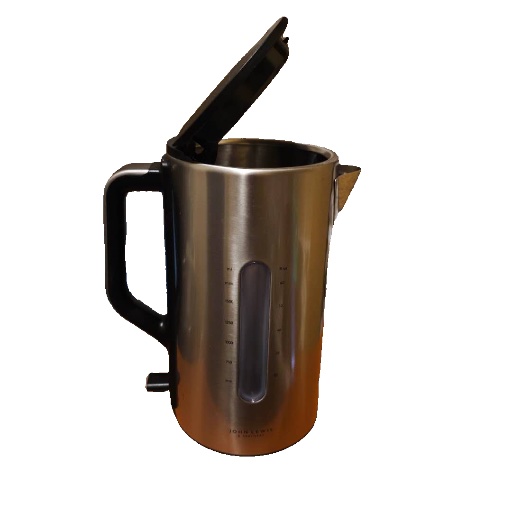}
    & \includegraphics[valign=c, width=0.1\textwidth]{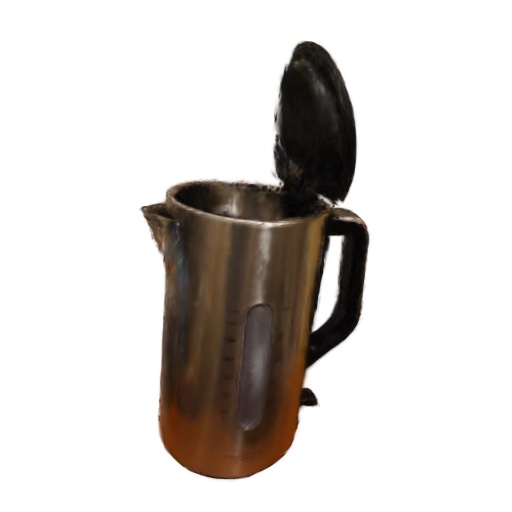} 
    & \includegraphics[valign=c, width=0.1\textwidth]{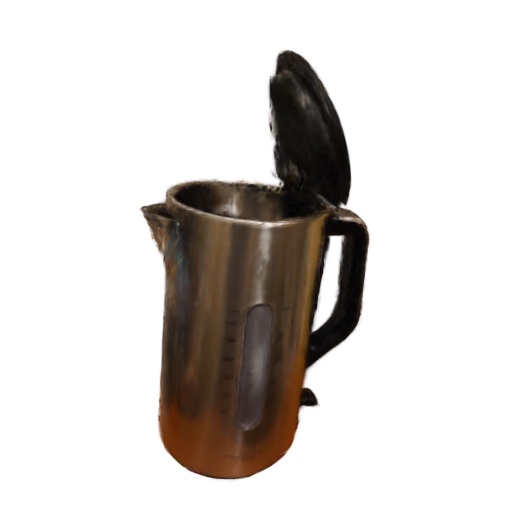}  
    & \includegraphics[valign=c, width=0.1\textwidth]{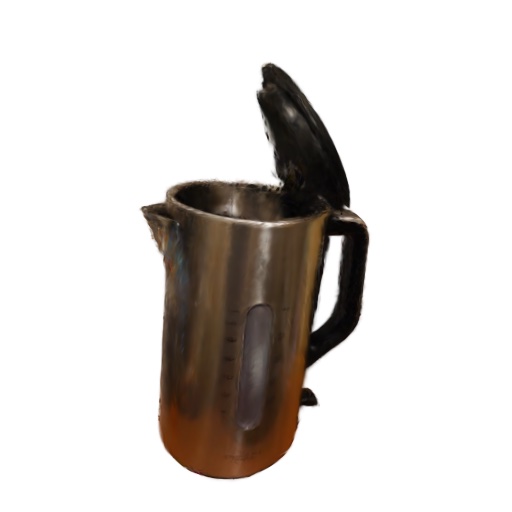} 
    & \includegraphics[valign=c, width=0.1\textwidth]{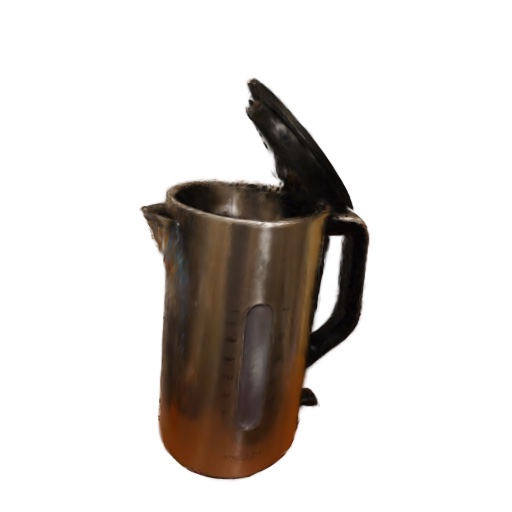} 
    & \includegraphics[valign=c, width=0.1\textwidth]{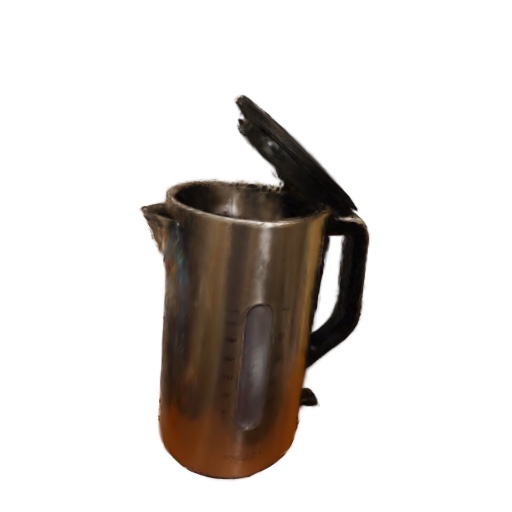}
    & \includegraphics[valign=c, width=0.1\textwidth]{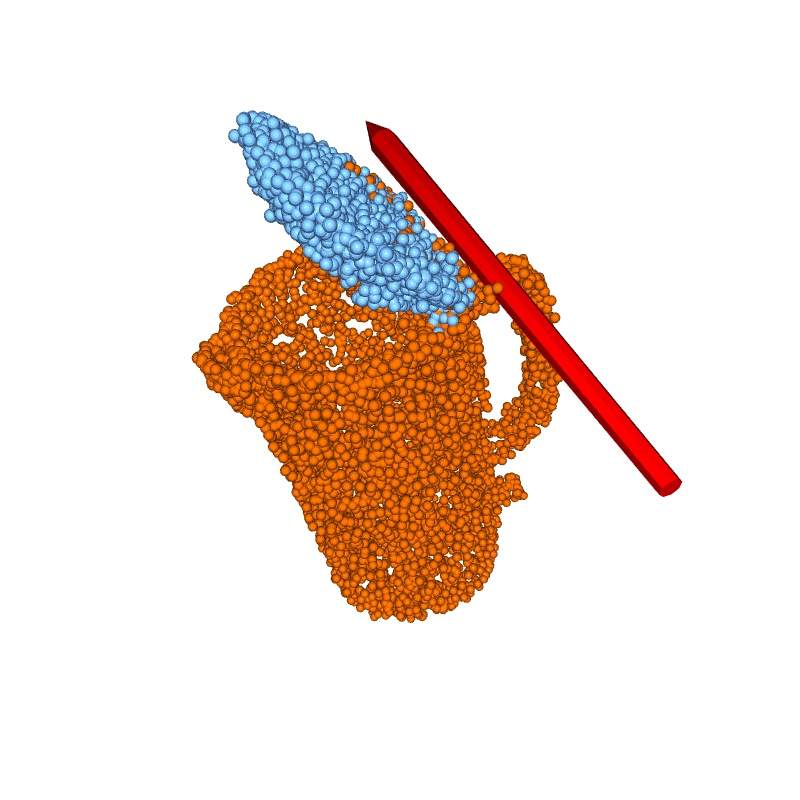} \\
    \includegraphics[valign=c, width=0.1\textwidth]{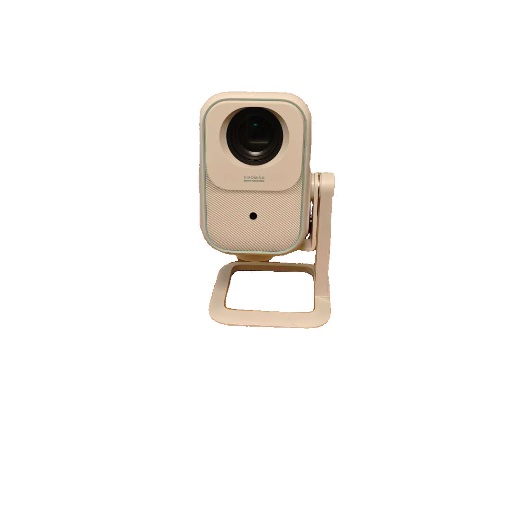} 
    &\includegraphics[valign=c, width=0.1\textwidth]{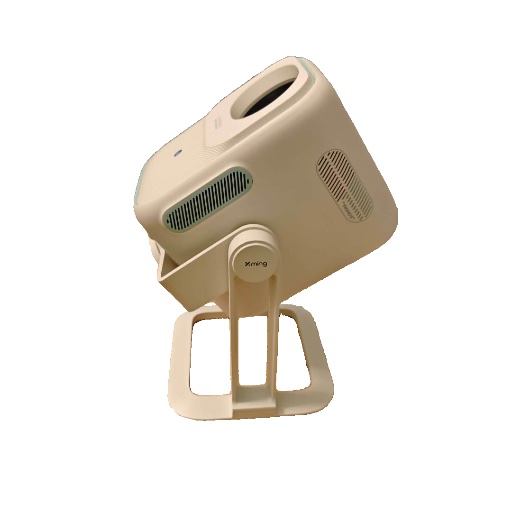}
    & \includegraphics[valign=c, width=0.1\textwidth]{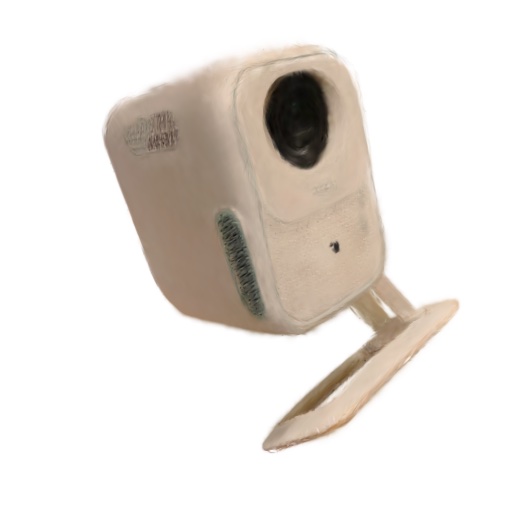} 
    & \includegraphics[valign=c, width=0.1\textwidth]{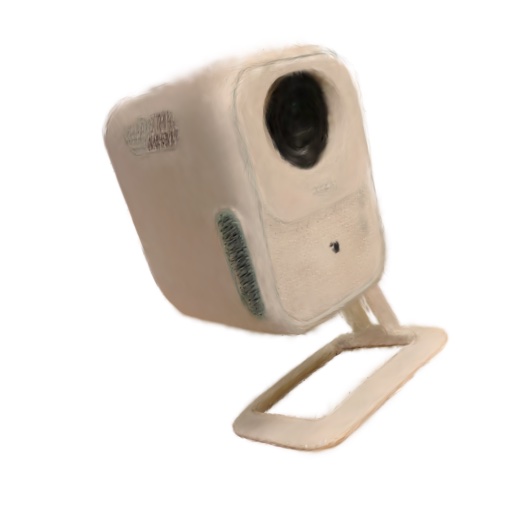}  
    & \includegraphics[valign=c, width=0.1\textwidth]{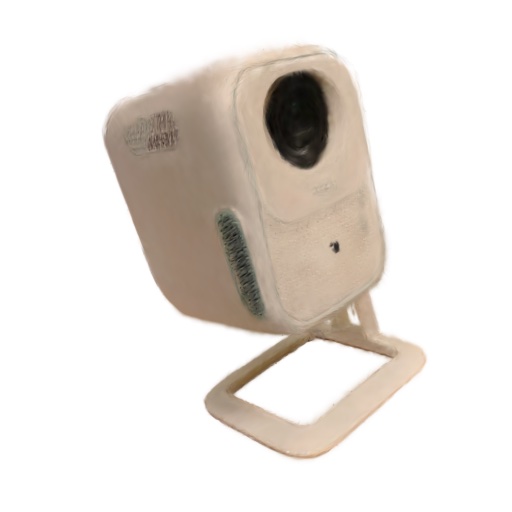} 
    & \includegraphics[valign=c, width=0.1\textwidth]{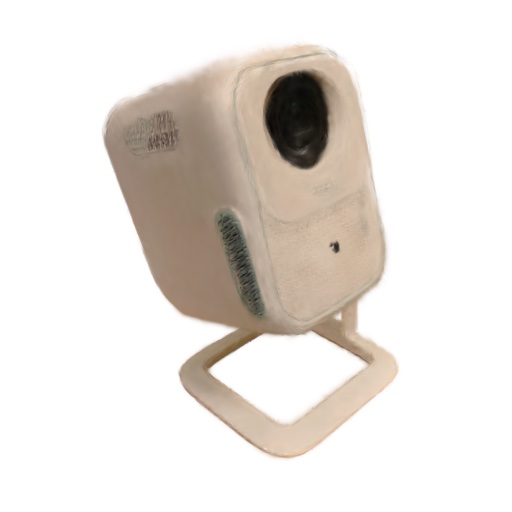} 
    & \includegraphics[valign=c, width=0.1\textwidth]{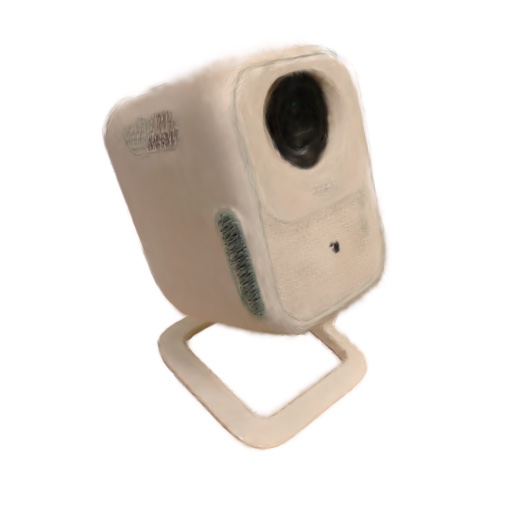} 
    & \includegraphics[valign=c, width=0.1\textwidth]{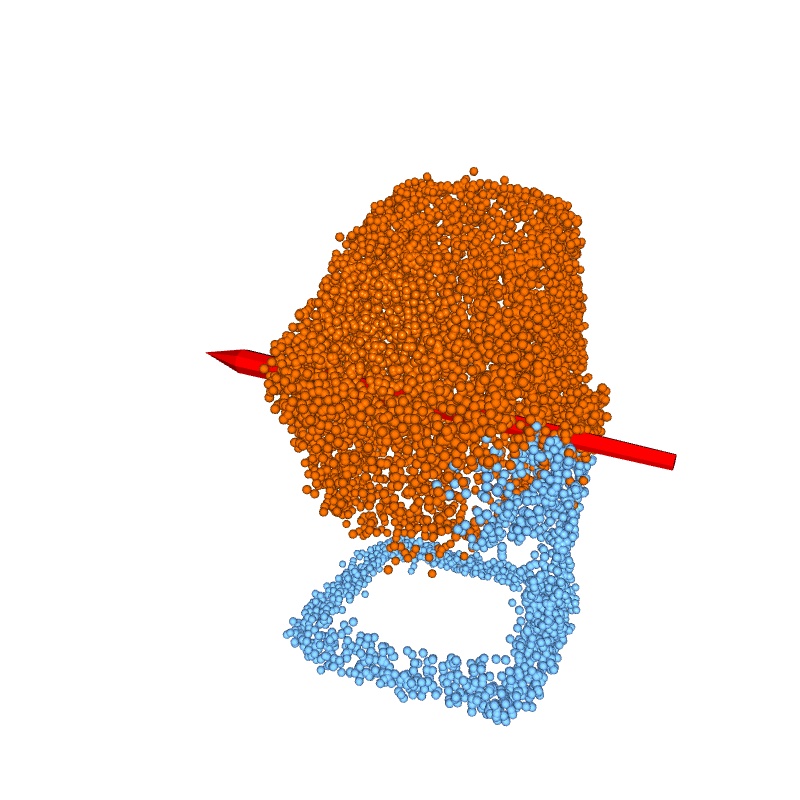}\\
    \includegraphics[valign=c, width=0.1\textwidth]{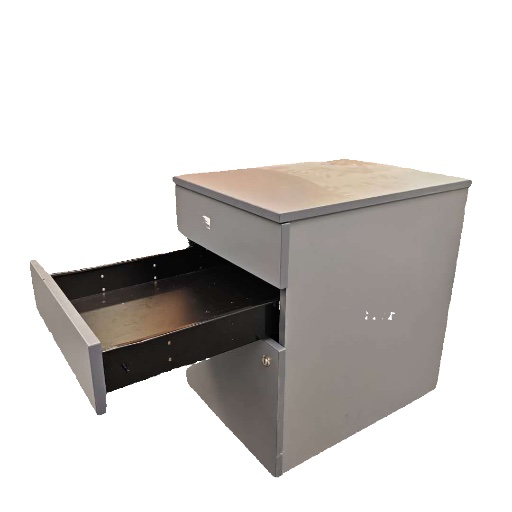} 
    &\includegraphics[valign=c, width=0.1\textwidth]{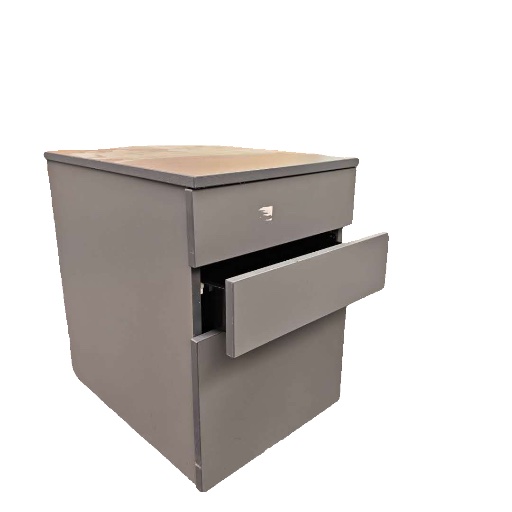}
    & \includegraphics[valign=c, width=0.1\textwidth]{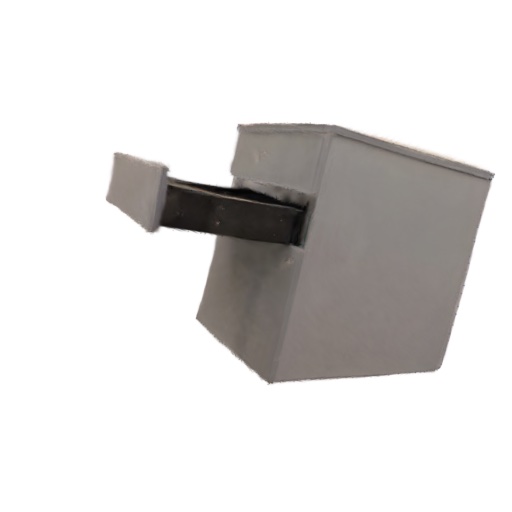} 
    & \includegraphics[valign=c, width=0.1\textwidth]{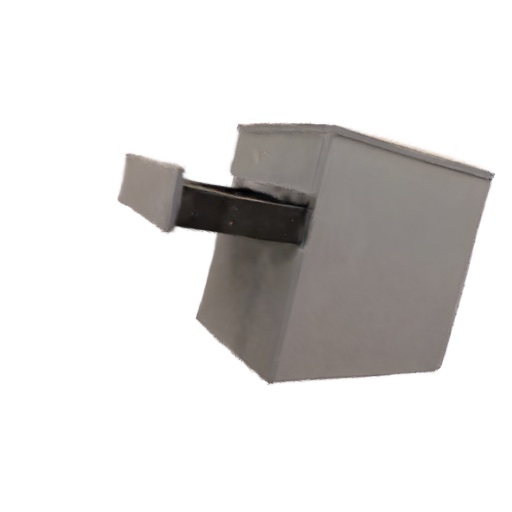}  
    & \includegraphics[valign=c, width=0.1\textwidth]{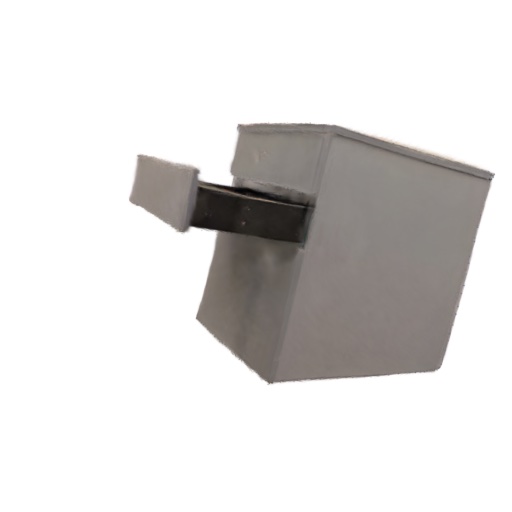} 
    & \includegraphics[valign=c, width=0.1\textwidth]{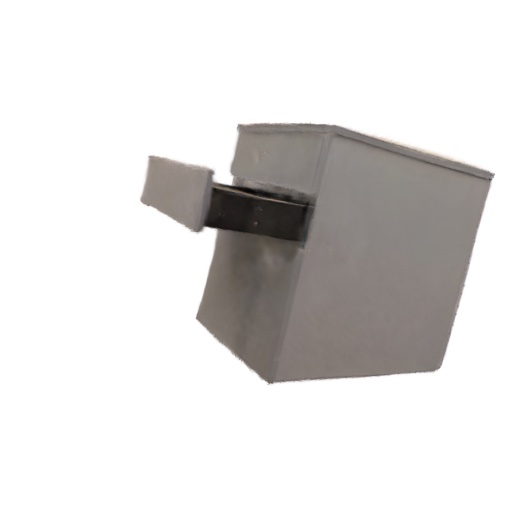} 
    & \includegraphics[valign=c, width=0.1\textwidth]{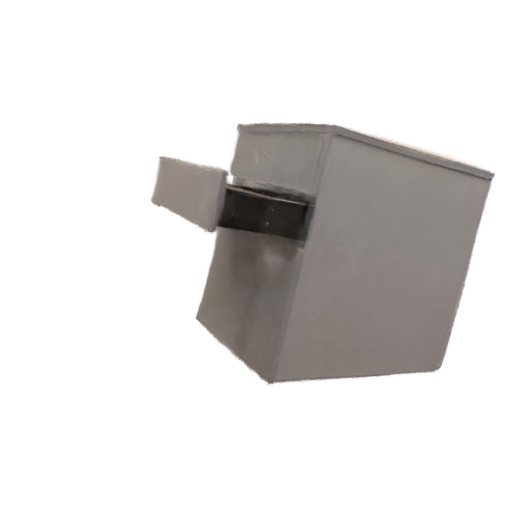} 
    & \includegraphics[valign=c, width=0.1\textwidth]{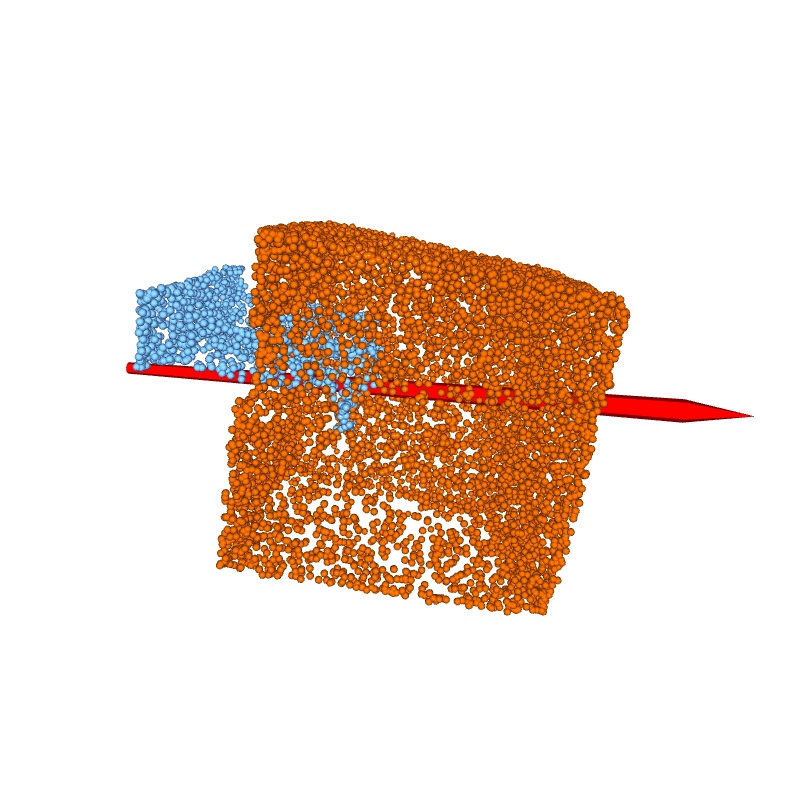}\\
    \includegraphics[valign=c, width=0.1\textwidth]{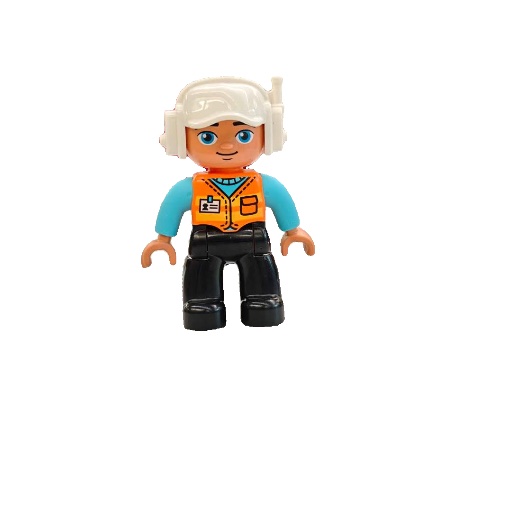} 
    &\includegraphics[valign=c, width=0.1\textwidth]{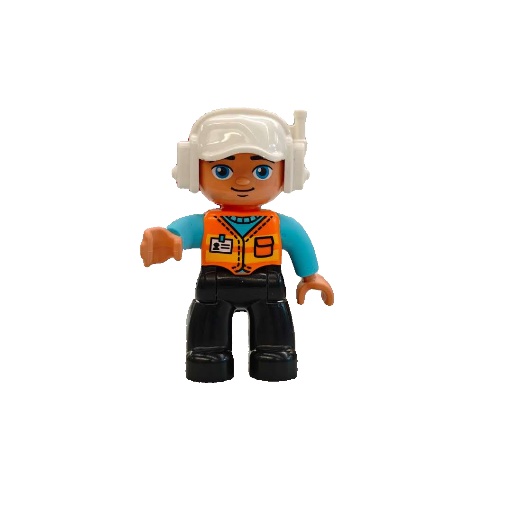}
    & \includegraphics[valign=c, width=0.1\textwidth]{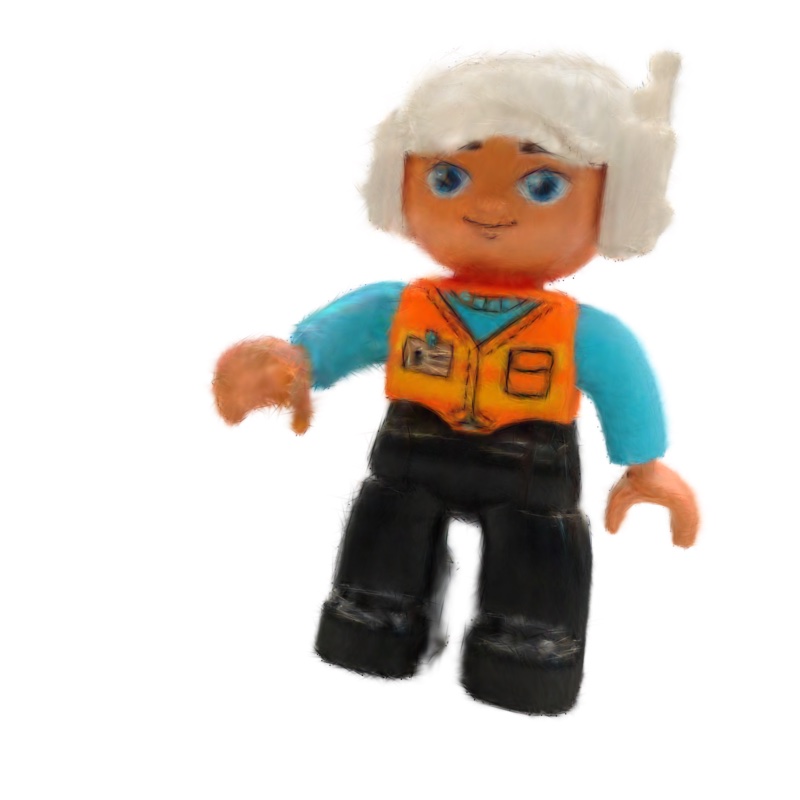} 
    & \includegraphics[valign=c, width=0.1\textwidth]{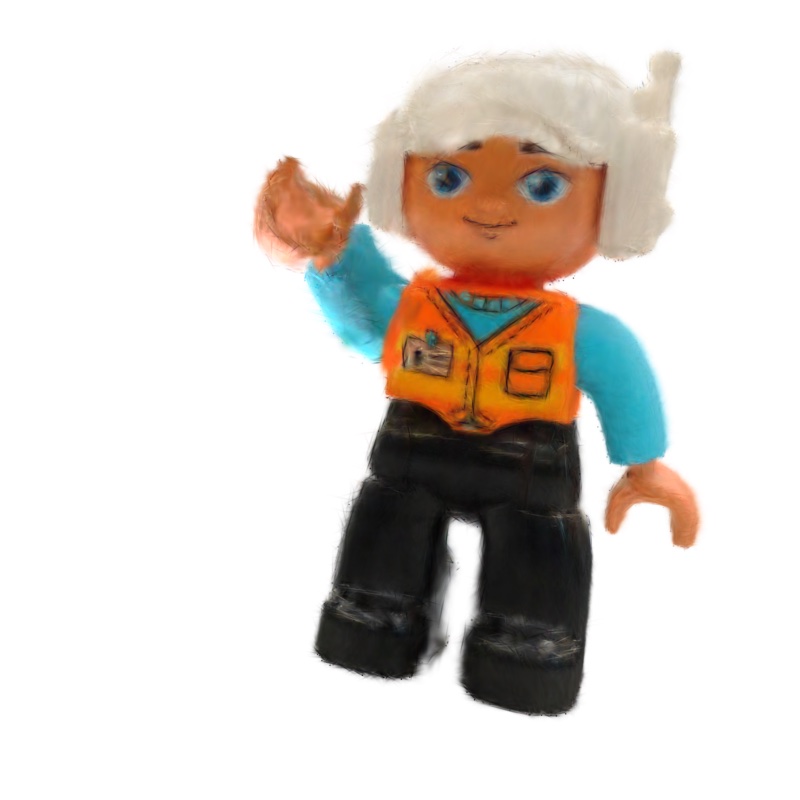}  
    & \includegraphics[valign=c, width=0.1\textwidth]{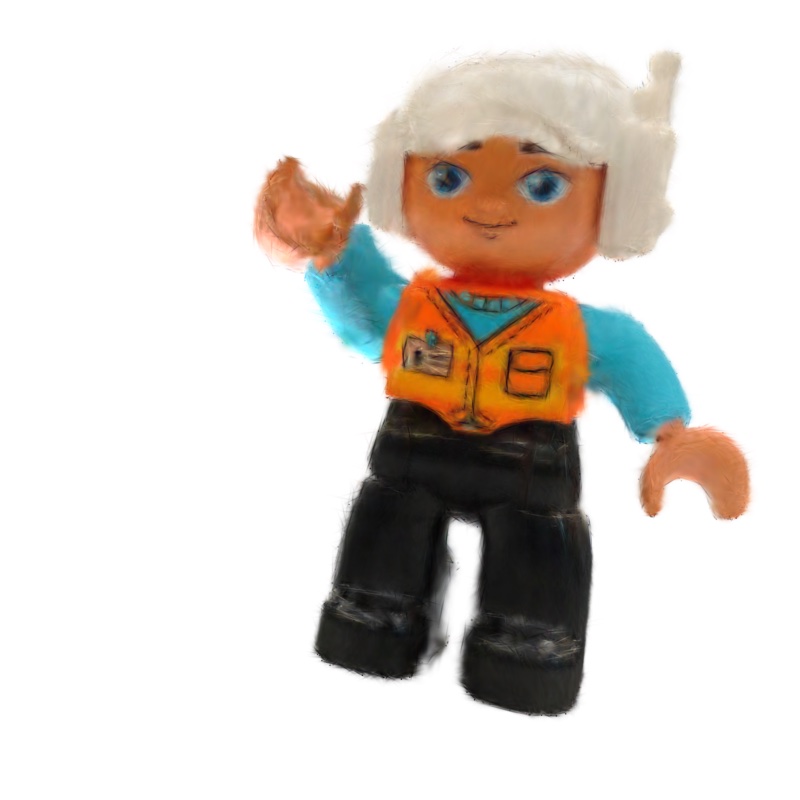} 
    & \includegraphics[valign=c, width=0.1\textwidth]{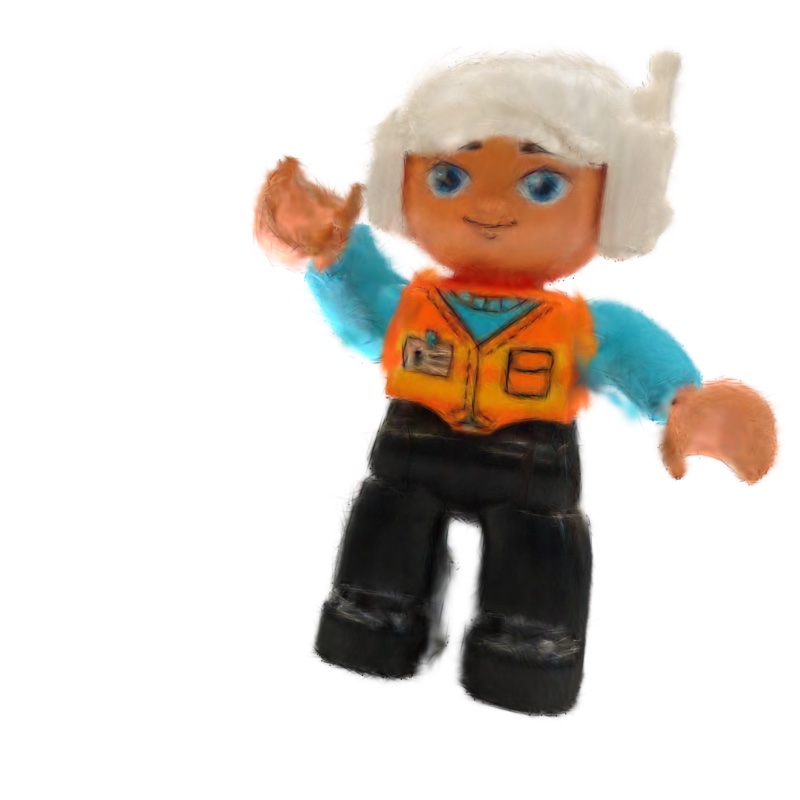} 
    & \includegraphics[valign=c, width=0.1\textwidth]{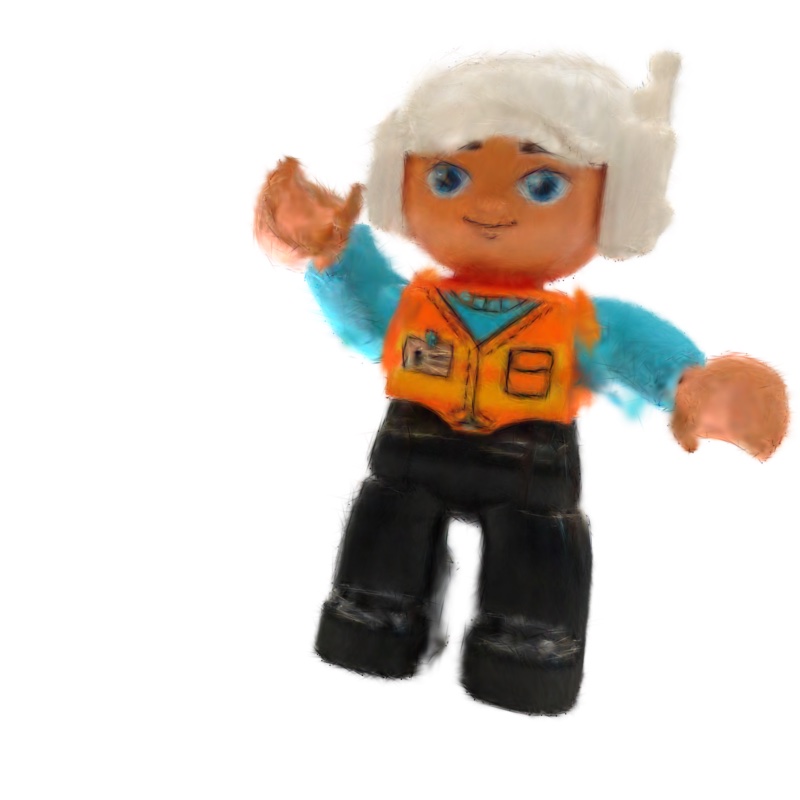} 
    & \includegraphics[valign=c, width=0.1\textwidth]{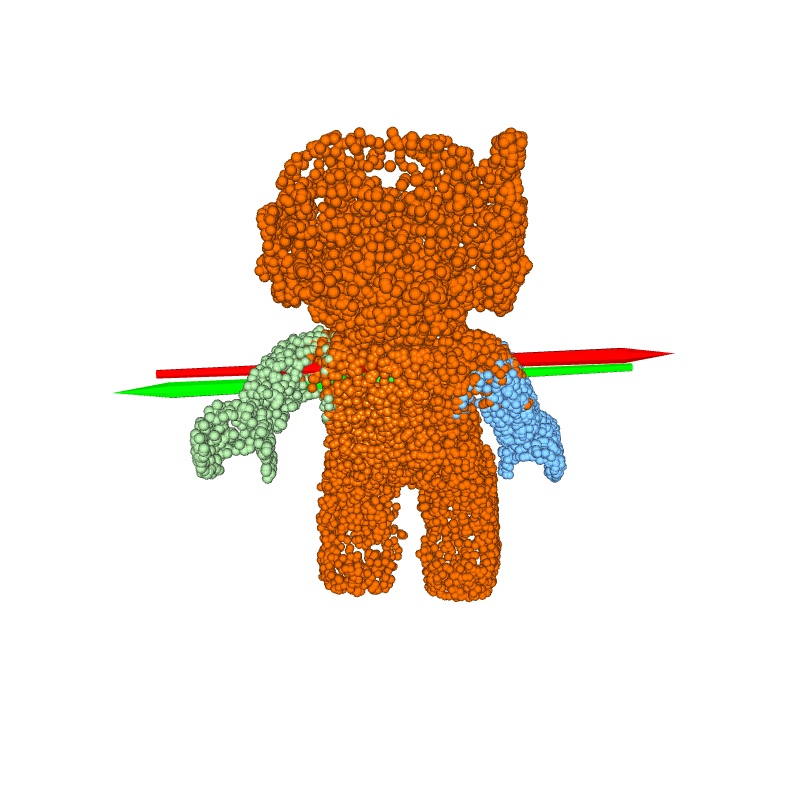}\\
    \end{tblr}
    }
    \vspace{1em}
    \caption{\textbf{Real-world objects.} We show the input images (left), novel articulation synthesis results (middle), and part-level segmentation with articulation axes (right). The first three objects are single-part articulated objects, while the last one is a multi-part articulated object. For the multi-part object, we collect three sets of views, one per articulation state. The segmentation results are shown in different colors for each part. The green lines denote the estimated articulation axes.}
    \label{fig:realworld}
\end{figure}

To demonstrate the generalization of our approach, we conduct experiments on four real-world objects (a bottle, projector, drawer, and LEGO toy), capturing four input images per articulation state. We employ the Segment Anything Model (SAM)~\cite{ravi2024sam2} to remove backgrounds prior to processing. For multi-part objects, like the LEGO toy, we capture three image sets (one per state) as detailed in \cref{sec:multi_part}. The results in \cref{fig:realworld} show that our method successfully generates seamless reconstructions for these real-world objects, validating its practical applicability. And we demonstrate our multpart extension on the LEGO toy, where we can successfully segment the toy into three parts (main body and two arms) and synthesize novel articulations by interpolating the joint parameters. 

%% file: tables/bigresults.tex
\begin{table*}[h]
    \centering
    \resizebox{0.95\textwidth}{!}{%
    \begin{tabular}{c c c| c |c c c c c c c c c c}
    \hline
     & &  Pipeline & Corr. & Foldchair & Fridge & Laptop & Oven & Scissor & Stapler & USB & Washer & Blade$^\dagger$ & Storage$^\dagger$ \\
    \hline

    \multirow{12}{*}{Motion} & \multirow{4}{*}{Ang Err $\downarrow$} & AGS\cite{guo2025articulatedgs} &  - & 62.083	&36.568	&86.058	&78.129	&62.083	&54.613	&88.677	&87.454	&59.562	&75.744 \\
    & & Ours &  GaussReg\cite{chang2024gaussreg} & failed	&failed	&failed	&failed	&failed	&failed	&failed &failed	&failed	&failed \\
    & & Ours &  FM-Dense\cite{cao2023self} & 0.974	& 13.267	& failed	&failed	&5.209	&failed	&12.603 &35.926	&3.716	&failed \\
    & & Ours & Ours & \cellcolor{orange!25}\textbf{0.095}	& \cellcolor{orange!25}\textbf{0.914}	& \cellcolor{orange!25}\textbf{1.020}	& \cellcolor{orange!25}\textbf{2.365}	& \cellcolor{orange!25}\textbf{0.839}	& \cellcolor{orange!25}\textbf{1.337}	& \cellcolor{orange!25}\textbf{1.872}	& \cellcolor{orange!25}\textbf{1.737}	& \cellcolor{orange!25}\textbf{0.878}	& \cellcolor{orange!25}\textbf{0.525} \\

    \cline{2-14}
    & \multirow{4}{*}{Pos Err $\downarrow$} & AGS\cite{guo2025articulatedgs} & - & 0.032	&0.124	&1.805	&0.252	&0.032	&0.149	&1.209	&0.231	&-	&-\\
    & & Ours & GaussReg\cite{chang2024gaussreg}& failed	&failed	&failed	&failed	&failed	&failed	&failed &failed	&-	&- \\
    & & Ours & FM-Dense\cite{cao2023self}& 0.729	&0.658	&failed	&failed	&0.394	&failed	&0.281	&0.417	&-	&-\\
    & & Ours & Ours & \cellcolor{orange!25}\textbf{0.180}	& \cellcolor{orange!25}\textbf{0.007}	& \cellcolor{orange!25}\textbf{0.106}	& \cellcolor{orange!25}\textbf{0.042}	& \cellcolor{orange!25}\textbf{0.058}	& \cellcolor{orange!25}\textbf{0.196}	& \cellcolor{orange!25}\textbf{0.099}	& \cellcolor{orange!25}\textbf{0.157}	& \cellcolor{orange!25}\textbf{-}	& \cellcolor{orange!25}\textbf{-}\\

    \cline{2-14}
    & \multirow{4}{*}{Motion Dist $\downarrow$} & AGS\cite{guo2025articulatedgs} & - & 70.900	&75.548	&46.925	&29.381	&70.900	&64.865	&61.955	&33.345	&1.275	&0.213\\
    & & Ours & GaussReg\cite{chang2024gaussreg} & failed	&failed	&failed	&failed	&failed	&failed	&failed &failed	&failed	&failed \\
    & & Ours & FM-Dense\cite{cao2023self} & 13.590	&25.261	&failed	&failed	&3.652	&failed	&11.285 &17.509	&0.189	&failed\\
    & & Ours &Ours & \cellcolor{orange!25}\textbf{0.861}	& \cellcolor{orange!25}\textbf{2.279}	& \cellcolor{orange!25}\textbf{0.788}	& \cellcolor{orange!25}\textbf{2.654}	& \cellcolor{orange!25}\textbf{1.516}	& \cellcolor{orange!25}\textbf{0.930}	& \cellcolor{orange!25}\textbf{1.877}	& \cellcolor{orange!25}\textbf{2.926}	& \cellcolor{orange!25}\textbf{0.017}	& \cellcolor{orange!25}\textbf{0.118} \\

    \hline
    \multirow{8}{*}{Rendering} & \multirow{3}{*}{PSNR $\uparrow$} & AGS\cite{guo2025articulatedgs} & -& 9.349	&11.294	&13.787	&13.871	&13.481	&12.898	&15.856	&17.644	&24.543	&14.333 \\
    & & Ours & GaussReg\cite{chang2024gaussreg} & failed	&failed	&failed	&failed	&failed	&failed	&failed &failed	&failed	&failed \\
    & & Ours & FM-Dense\cite{cao2023self} & 17.897	&19.256	&21.161	&19.356	&23.276	&19.002	&22.517	&20.632	&23.571	&failed\\

    & & Ours & Ours &\cellcolor{orange!25}\textbf{24.659}	&\cellcolor{orange!25}\textbf{26.920}	&\cellcolor{orange!25}\textbf{26.491}	&\cellcolor{orange!25}\textbf{23.434}	&\cellcolor{orange!25}\textbf{28.823}	&\cellcolor{orange!25}\textbf{26.532}	&\cellcolor{orange!25}\textbf{27.016}	&\cellcolor{orange!25}\textbf{27.416}	&\cellcolor{orange!25}\textbf{29.041}	&\cellcolor{orange!25}\textbf{25.493} \\
    \cline{2-14}
    & \multirow{4}{*}{SSIM $\uparrow$} & AGS\cite{guo2025articulatedgs} & - & 0.652	&0.722	&0.809	&0.723	&0.846	&0.831	&0.842	&0.877	&0.962	&0.765\\
    & & Ours & GaussReg\cite{chang2024gaussreg} & failed	&failed	&failed	&failed	&failed	&failed	&failed &failed	&failed	&failed \\
    & & Ours & FM-Dense\cite{cao2023self} & 0.889	&0.916	&0.923	&0.893	&0.953	&0.930	&0.821	&0.933	&0.942	&failed \\
    & & Ours &Ours & \cellcolor{orange!25}\textbf{0.940}	&\cellcolor{orange!25}\textbf{0.963}	&\cellcolor{orange!25}\textbf{0.955}	&\cellcolor{orange!25}\textbf{0.931}	&\cellcolor{orange!25}\textbf{0.967}	&\cellcolor{orange!25}\textbf{0.953}	&\cellcolor{orange!25}\textbf{0.956}	&\cellcolor{orange!25}\textbf{0.966}	&\cellcolor{orange!25}\textbf{0.974}	&\cellcolor{orange!25}\textbf{0.927} \\
    \hline
    \multirow{4}{*}{Geometry} & \multirow{4}{*}{CD $\downarrow$} & AGS\cite{guo2025articulatedgs} & - & 1188.824	&717.780	&893.149	&244.862	&1188.824	&1599.261	&467.633	&268.377	&90.288	&231.355\\
    & & Ours & GaussReg\cite{chang2024gaussreg} & failed	&failed	&failed	&failed	&failed	&failed	&failed &failed	&failed	&failed \\
    & & Ours & FM-Dense\cite{cao2023self} & 4.814	&6.399	&failed	&failed	&0.501	&failed	&8.935	&25.515	&2.789	&failed \\
    & & Ours & Ours & \cellcolor{orange!25}\textbf{0.856}	& \cellcolor{orange!25}\textbf{2.051}	& \cellcolor{orange!25}\textbf{0.610}	& \cellcolor{orange!25}\textbf{6.449}	& \cellcolor{orange!25}\textbf{0.264}	& \cellcolor{orange!25}\textbf{1.216}	& \cellcolor{orange!25}\textbf{1.430}	& \cellcolor{orange!25}\textbf{1.615}	& \cellcolor{orange!25}\textbf{1.788}	& \cellcolor{orange!25}\textbf{6.443} \\
    \hline
    \multirow{4}{*}{Segmentation} & \multirow{4}{*}{mIoU $\uparrow$} & AGS\cite{guo2025articulatedgs} & - & 0.356	&0.517	&0.540	&0.347	&0.691	&0.633	&0.638	&0.501	&0.386	&0.487\\
    & & Ours & GaussReg\cite{chang2024gaussreg} & failed	&failed	&failed	&failed	&failed	&failed	&failed &failed	&failed	&failed \\
    & & Ours & FM-Dense\cite{cao2023self} & 0.345	&0.399	&failed	&failed	&0.623	&failed	&0.412	&0.390	&0.372	&failed \\
    & & Ours & Ours & \cellcolor{orange!25}\textbf{0.965}	& \cellcolor{orange!25}\textbf{0.872}	& \cellcolor{orange!25}\textbf{0.903}	& \cellcolor{orange!25}\textbf{0.892}	& \cellcolor{orange!25}\textbf{0.913}	& \cellcolor{orange!25}\textbf{0.952}	& \cellcolor{orange!25}\textbf{0.881}	& \cellcolor{orange!25}\textbf{0.874}	& \cellcolor{orange!25}\textbf{0.842}	& \cellcolor{orange!25}\textbf{0.905} \\
    \hline
    \end{tabular}%
    }
    \vspace{1em}
    \caption{Comparisons to the baseline methods with 4-view images from each articulation states (source and target). Objects with $\dagger$ are with prismatic joints. Best results are highlighted in \colorbox{orange!25}{orange}.}
    \vspace{-1em}
    \label{tab:quantitative}
\end{table*}

%% file: sec/05conclusion.tex
\section{Conclusion}


We have introduced \nameshort, 
to our knowledge, the first framework for articulated object modeling from independently reconstructed sparse, unposed articulation states.
We identified that the central challenge in this setting is not just reconstruction, but the 3D-to-3D correspondence problem between unaligned 3D models.

Our solution is a test-time, optimization-based deformation field that finds dense and robust correspondences by leveraging both 3D geometry and 2D rendering losses. Our experiments show this approach is highly effective, whereas prominent baselines—including sparse keypoint methods and pre-trained dense networks—fail to solve the problem. Combined with robust part-motion initialization and staged joint refinement, our pipeline recovers geometry, appearance, part segmentation, and kinematic parameters under substantially weaker input assumptions than prior articulated-object modeling methods.

While effective, our method assumes rigid parts and requires at least two distinct articulation states with adequate view coverage. 
Our current multi-part extension assumes that each source-target pair isolates the motion of a single moving part. 
When multiple parts move simultaneously, the correspondence set can contain several non-dominant rigid motions, making sequential TEASER initialization ambiguous.
Future work could relax this assumption by jointly estimating multiple rigid motions from dense correspondences, rather than extracting one dominant motion at a time, and by enforcing part-consistency across more than two articulation states.
Furthermore, our alignment may fail if the initial, arbitrary coordinate systems of the 3D GS reconstructions are extremely misaligned. 
More visualization can be found in the supplementary.

%% file: main.bib
@String(CVPR  = {CVPR})

@String(ICCV  = {ICCV})

@String(ECCV  = {ECCV})

@String(NEURIPS= {NeurIPS})

@String(ICML  = {ICML})

@String(ICLR  = {ICLR})

@String(ICRA  = {ICRA})

@String(TOG   = {ACM TOG})

@String(ACCV  = {ACCV})

@String(CVPR   = {IEEE/CVF Conference on Computer Vision and Pattern Recognition})

@String(ICCV   = {IEEE/CVF International Conference on Computer Vision})

@String(ECCV   = {European Conference on Computer Vision})

@String(NEURIPS= {Advances in Neural Information Processing Systems})

@String(ICML   = {International Conference on Machine Learning})

@String(ICLR   = {International Conference on Learning Representations})

@String(ICRA   = {IEEE International Conference on Robotics and Automation})

@String(TOG    = {ACM Transactions on Graphics})

@String(ACCV   = {Asian Conference on Computer Vision})

@inproceedings{xu2025freesplatter,
  title={Freesplatter: Pose-free gaussian splatting for sparse-view 3d reconstruction},
  author={Xu, Jiale and Gao, Shenghua and Shan, Ying},
  booktitle={Proceedings of the IEEE/CVF International Conference on Computer Vision},
  pages={25442--25452},
  year={2025}
}

@article{yang2020teaser,
  title={Teaser: Fast and certifiable point cloud registration},
  author={Yang, Heng and Shi, Jingnan and Carlone, Luca},
  journal={IEEE Transactions on Robotics},
  volume={37},
  number={2},
  pages={314--333},
  year={2020},
  publisher={IEEE}
}

@InProceedings{Mo_2019_CVPR,
author = {Mo, Kaichun and Zhu, Shilin and Chang, Angel X. and Yi, Li and Tripathi, Subarna and Guibas, Leonidas J. and Su, Hao},
title = {{PartNet}: A Large-Scale Benchmark for Fine-Grained and Hierarchical Part-Level {3D} Object Understanding},
booktitle = CVPR,
month = {June},
year = {2019}
}

@inproceedings{guo2025articulatedgs,
  title={Articulatedgs: Self-supervised digital twin modeling of articulated objects using 3d gaussian splatting},
  author={Guo, Junfu and Xin, Yu and Liu, Gaoyi and Xu, Kai and Liu, Ligang and Hu, Ruizhen},
  booktitle=CVPR,
  pages={27144--27153},
  year={2025}
}

@inproceedings{liu2023paris,
  title={Paris: Part-level reconstruction and motion analysis for articulated objects},
  author={Liu, Jiayi and Mahdavi-Amiri, Ali and Savva, Manolis},
  booktitle=ICCV,
  pages={352--363},
  year={2023}
}

@inproceedings{deng2024articulate,
  title={Articulate your nerf: Unsupervised articulated object modeling via conditional view synthesis},
  author={Deng, Jianning and Subr, Kartic and Bilen, Hakan},
  booktitle=NEURIPS,
  volume={37},
  pages={119717--119741},
  year={2024}
}

@inproceedings{wang2025vggt,
  title={Vggt: Visual geometry grounded transformer},
  author={Wang, Jianyuan and Chen, Minghao and Karaev, Nikita and Vedaldi, Andrea and Rupprecht, Christian and Novotny, David},
  booktitle=CVPR,
  pages={5294--5306},
  year={2025}
}

@inproceedings{wang2024dust3r,
  title={Dust3r: Geometric 3d vision made easy},
  author={Wang, Shuzhe and Leroy, Vincent and Cabon, Yohann and Chidlovskii, Boris and Revaud, Jerome},
  booktitle=CVPR,
  pages={20697--20709},
  year={2024}
}

@inproceedings{leroy2024grounding,
  title={Grounding image matching in 3d with mast3r},
  author={Leroy, Vincent and Cabon, Yohann and Revaud, J{\'e}r{\^o}me},
  booktitle=ECCV,
  pages={71--91},
  year={2024},
  organization={Springer}
}

@inproceedings{schoenberger2016sfm,
    author={Sch\"{o}nberger, Johannes Lutz and Frahm, Jan-Michael},
    title={Structure-from-Motion Revisited},
    booktitle=CVPR,
    year={2016},
}

@inproceedings{schoenberger2016mvs,
    author={Sch\"{o}nberger, Johannes Lutz and Zheng, Enliang and Pollefeys, Marc and Frahm, Jan-Michael},
    title={Pixelwise View Selection for Unstructured Multi-View Stereo},
    booktitle=ECCV,
    year={2016},
}

@article{agarwal2011building,
  title={Building rome in a day},
  author={Agarwal, Sameer and Furukawa, Yasutaka and Snavely, Noah and Simon, Ian and Curless, Brian and Seitz, Steven M and Szeliski, Richard},
  journal={Communications of the ACM},
  volume={54},
  number={10},
  pages={105--112},
  year={2011},
  publisher={ACM New York, NY, USA}
}

@inproceedings{liu2024robust,
  title={Robust incremental structure-from-motion with hybrid features},
  author={Liu, Shaohui and Gao, Yidan and Zhang, Tianyi and Pautrat, R{\'e}mi and Sch{\"o}nberger, Johannes L and Larsson, Viktor and Pollefeys, Marc},
  booktitle=ECCV,
  pages={249--269},
  year={2024},
  organization={Springer}
}

@inproceedings{yang2025fast3r,
  title={Fast3r: Towards 3d reconstruction of 1000+ images in one forward pass},
  author={Yang, Jianing and Sax, Alexander and Liang, Kevin J and Henaff, Mikael and Tang, Hao and Cao, Ang and Chai, Joyce and Meier, Franziska and Feiszli, Matt},
  booktitle=CVPR,
  pages={21924--21935},
  year={2025}
}

@article{kerbl20233d,
  title={3D Gaussian splatting for real-time radiance field rendering.},
  author={Kerbl, Bernhard and Kopanas, Georgios and Leimk{\"u}hler, Thomas and Drettakis, George},
  journal={ACM Trans. Graph.},
  volume={42},
  number={4},
  pages={139--1},
  year={2023}
}

@inproceedings{huang20242d,
  title={2d gaussian splatting for geometrically accurate radiance fields},
  author={Huang, Binbin and Yu, Zehao and Chen, Anpei and Geiger, Andreas and Gao, Shenghua},
  booktitle={ACM SIGGRAPH 2024 conference papers},
  pages={1--11},
  year={2024}
}

@inproceedings{cheng2024gaussianpro,
  title={Gaussianpro: 3d gaussian splatting with progressive propagation},
  author={Cheng, Kai and Long, Xiaoxiao and Yang, Kaizhi and Yao, Yao and Yin, Wei and Ma, Yuexin and Wang, Wenping and Chen, Xuejin},
  booktitle=ICML,
  year={2024}
}

@inproceedings{fan2024lightgaussian,
  title={Lightgaussian: Unbounded 3d gaussian compression with 15x reduction and 200+ fps},
  author={Fan, Zhiwen and Wang, Kevin and Wen, Kairun and Zhu, Zehao and Xu, Dejia and Wang, Zhangyang and others},
  booktitle=NEURIPS,
  volume={37},
  pages={140138--140158},
  year={2024}
}

@inproceedings{lu2023scaffold,
  title={Scaffold-gs: Structured 3d gaussians for view-adaptive rendering. 2024 IEEE},
  author={Lu, Tao and Yu, Mulin and Xu, Linning and Xiangli, Yuanbo and Wang, Limin and Lin, Dahua and Dai, Bo},
  booktitle=CVPR,
  pages={20654--20664},
  year={2023}
}

@inproceedings{lee2024compact,
  title={Compact 3d gaussian representation for radiance field},
  author={Lee, Joo Chan and Rho, Daniel and Sun, Xiangyu and Ko, Jong Hwan and Park, Eunbyung},
  booktitle=CVPR,
  pages={21719--21728},
  year={2024}
}

@inproceedings{yan2024multi,
  title={Multi-scale 3d gaussian splatting for anti-aliased rendering},
  author={Yan, Zhiwen and Low, Weng Fei and Chen, Yu and Lee, Gim Hee},
  booktitle=CVPR,
  pages={20923--20931},
  year={2024}
}

@article{mildenhall2021nerf,
  title={Nerf: Representing scenes as neural radiance fields for view synthesis},
  author={Mildenhall, Ben and Srinivasan, Pratul P and Tancik, Matthew and Barron, Jonathan T and Ramamoorthi, Ravi and Ng, Ren},
  journal={Communications of the ACM},
  volume={65},
  number={1},
  pages={99--106},
  year={2021},
  publisher={ACM New York, NY, USA}
}

@article{wang2021neus,
  title={Neus: Learning neural implicit surfaces by volume rendering for multi-view reconstruction},
  author={Wang, Peng and Liu, Lingjie and Liu, Yuan and Theobalt, Christian and Komura, Taku and Wang, Wenping},
  journal={arXiv preprint arXiv:2106.10689},
  year={2021}
}

@article{muller2022instant,
  title={Instant neural graphics primitives with a multiresolution hash encoding},
  author={M{\"u}ller, Thomas and Evans, Alex and Schied, Christoph and Keller, Alexander},
  journal={ACM transactions on graphics (TOG)},
  volume={41},
  number={4},
  pages={1--15},
  year={2022},
  publisher={ACM New York, NY, USA}
}

@inproceedings{zhu2024fsgs,
  title={Fsgs: Real-time few-shot view synthesis using gaussian splatting},
  author={Zhu, Zehao and Fan, Zhiwen and Jiang, Yifan and Wang, Zhangyang},
  booktitle=ECCV,
  pages={145--163},
  year={2024},
  organization={Springer}
}

@inproceedings{li2024dngaussian,
  title={Dngaussian: Optimizing sparse-view 3d gaussian radiance fields with global-local depth normalization},
  author={Li, Jiahe and Zhang, Jiawei and Bai, Xiao and Zheng, Jin and Ning, Xin and Zhou, Jun and Gu, Lin},
  booktitle=CVPR,
  pages={20775--20785},
  year={2024}
}

@article{ye2024no,
  title={No pose, no problem: Surprisingly simple 3d gaussian splats from sparse unposed images},
  author={Ye, Botao and Liu, Sifei and Xu, Haofei and Li, Xueting and Pollefeys, Marc and Yang, Ming-Hsuan and Peng, Songyou},
  journal={arXiv preprint arXiv:2410.24207},
  year={2024}
}

@inproceedings{szymanowicz2024splatter,
  title={Splatter image: Ultra-fast single-view 3d reconstruction},
  author={Szymanowicz, Stanislaw and Rupprecht, Chrisitian and Vedaldi, Andrea},
  booktitle=CVPR,
  pages={10208--10217},
  year={2024}
}

@inproceedings{kong2025generative,
  title={Generative Sparse-View Gaussian Splatting},
  author={Kong, Hanyang and Yang, Xingyi and Wang, Xinchao},
  booktitle=CVPR,
  pages={26745--26755},
  year={2025}
}

@inproceedings{liu2023zero,
  title={Zero-1-to-3: Zero-shot one image to 3d object},
  author={Liu, Ruoshi and Wu, Rundi and Van Hoorick, Basile and Tokmakov, Pavel and Zakharov, Sergey and Vondrick, Carl},
  booktitle=ICCV,
  pages={9298--9309},
  year={2023}
}

@inproceedings{long2024wonder3d,
  title={Wonder3d: Single image to 3d using cross-domain diffusion},
  author={Long, Xiaoxiao and Guo, Yuan-Chen and Lin, Cheng and Liu, Yuan and Dou, Zhiyang and Liu, Lingjie and Ma, Yuexin and Zhang, Song-Hai and Habermann, Marc and Theobalt, Christian and others},
  booktitle=CVPR,
  pages={9970--9980},
  year={2024}
}

@inproceedings{tang2024lgm,
  title={Lgm: Large multi-view gaussian model for high-resolution 3d content creation},
  author={Tang, Jiaxiang and Chen, Zhaoxi and Chen, Xiaokang and Wang, Tengfei and Zeng, Gang and Liu, Ziwei},
  booktitle=ECCV,
  pages={1--18},
  year={2024},
  organization={Springer}
}

@inproceedings{charatan2024pixelsplat,
  title={pixelsplat: 3d gaussian splats from image pairs for scalable generalizable 3d reconstruction},
  author={Charatan, David and Li, Sizhe Lester and Tagliasacchi, Andrea and Sitzmann, Vincent},
  booktitle=CVPR,
  pages={19457--19467},
  year={2024}
}

@inproceedings{chen2024mvsplat,
  title={Mvsplat: Efficient 3d gaussian splatting from sparse multi-view images},
  author={Chen, Yuedong and Xu, Haofei and Zheng, Chuanxia and Zhuang, Bohan and Pollefeys, Marc and Geiger, Andreas and Cham, Tat-Jen and Cai, Jianfei},
  booktitle=ECCV,
  pages={370--386},
  year={2024},
  organization={Springer}
}

@inproceedings{chung2024depth,
  title={Depth-regularized optimization for 3d gaussian splatting in few-shot images},
  author={Chung, Jaeyoung and Oh, Jeongtaek and Lee, Kyoung Mu},
  booktitle=CVPR,
  pages={811--820},
  year={2024}
}

@inproceedings{mu2021sdf,
  title={A-sdf: Learning disentangled signed distance functions for articulated shape representation},
  author={Mu, Jiteng and Qiu, Weichao and Kortylewski, Adam and Yuille, Alan and Vasconcelos, Nuno and Wang, Xiaolong},
  booktitle=ICCV,
  pages={13001--13011},
  year={2021}
}

@inproceedings{jiang2022ditto,
  title={Ditto: Building digital twins of articulated objects from interaction},
  author={Jiang, Zhenyu and Hsu, Cheng-Chun and Zhu, Yuke},
  booktitle=CVPR,
  pages={5616--5626},
  year={2022}
}

@inproceedings{tseng2022cla,
  title={Cla-nerf: Category-level articulated neural radiance field},
  author={Tseng, Wei-Cheng and Liao, Hung-Ju and Yen-Chen, Lin and Sun, Min},
  booktitle=ICRA,
  pages={8454--8460},
  year={2022},
  organization={IEEE}
}

@inproceedings{kawana2023detection,
  title={Detection based part-level articulated object reconstruction from single RGBD image},
  author={Kawana, Yuki and Harada, Tatsuya},
  booktitle=NEURIPS,
  volume={36},
  pages={18444--18473},
  year={2023}
}

@inproceedings{weng2024neural,
  title={Neural implicit representation for building digital twins of unknown articulated objects},
  author={Weng, Yijia and Wen, Bowen and Tremblay, Jonathan and Blukis, Valts and Fox, Dieter and Guibas, Leonidas and Birchfield, Stan},
  booktitle=CVPR,
  pages={3141--3150},
  year={2024}
}

@inproceedings{liu2025building,
  title={Building interactable replicas of complex articulated objects via gaussian splatting},
  author={Liu, Yu and Jia, Baoxiong and Lu, Ruijie and Ni, Junfeng and Zhu, Song-Chun and Huang, Siyuan},
  booktitle=ICLR,
  year={2025}
}

@article{le2024articulate,
  title={Articulate-anything: Automatic modeling of articulated objects via a vision-language foundation model},
  author={Le, Long and Xie, Jason and Liang, William and Wang, Hung-Ju and Yang, Yue and Ma, Yecheng Jason and Vedder, Kyle and Krishna, Arjun and Jayaraman, Dinesh and Eaton, Eric},
  journal={arXiv preprint arXiv:2410.13882},
  year={2024}
}

@inproceedings{che2024op,
  title={Op-align: Object-level and part-level alignment for self-supervised category-level articulated object pose estimation},
  author={Che, Yuchen and Furukawa, Ryo and Kanezaki, Asako},
  booktitle=ECCV,
  pages={72--88},
  year={2024},
  organization={Springer}
}

@inproceedings{fu2024capt,
  title={Capt: Category-level articulation estimation from a single point cloud using transformer},
  author={Fu, Lian and Ishikawa, Ryoichi and Sato, Yoshihiro and Oishi, Takeshi},
  booktitle=ICRA,
  pages={751--757},
  year={2024},
  organization={IEEE}
}

@inproceedings{liu2023building,
  title={Building rearticulable models for arbitrary 3d objects from 4d point clouds},
  author={Liu, Shaowei and Gupta, Saurabh and Wang, Shenlong},
  booktitle=CVPR,
  pages={21138--21147},
  year={2023}
}

@inproceedings{wei2022self,
  title={Self-supervised neural articulated shape and appearance models},
  author={Wei, Fangyin and Chabra, Rohan and Ma, Lingni and Lassner, Christoph and Zollh{\"o}fer, Michael and Rusinkiewicz, Szymon and Sweeney, Chris and Newcombe, Richard and Slavcheva, Mira},
  booktitle=CVPR,
  pages={15816--15826},
  year={2022}
}

@inproceedings{liu2024cage,
  title={Cage: Controllable articulation generation},
  author={Liu, Jiayi and Tam, Hou In Ivan and Mahdavi-Amiri, Ali and Savva, Manolis},
  booktitle=CVPR,
  pages={17880--17889},
  year={2024}
}

@article{lei2023nap,
  title={Nap: Neural 3d articulation prior},
  author={Lei, Jiahui and Deng, Congyue and Shen, Bokui and Guibas, Leonidas and Daniilidis, Kostas},
  journal={arXiv preprint arXiv:2305.16315},
  year={2023}
}

@inproceedings{qian2022understanding,
  title={Understanding 3d object articulation in internet videos},
  author={Qian, Shengyi and Jin, Linyi and Rockwell, Chris and Chen, Siyi and Fouhey, David F},
  booktitle=CVPR,
  pages={1599--1609},
  year={2022}
}

@inproceedings{jiang2022opd,
  title={Opd: Single-view 3d openable part detection},
  author={Jiang, Hanxiao and Mao, Yongsen and Savva, Manolis and Chang, Angel X},
  booktitle=ECCV,
  pages={410--426},
  year={2022},
  organization={Springer}
}

@article{sun2023opdmulti,
  title={Opdmulti: Openable part detection for multiple objects},
  author={Sun, Xiaohao and Jiang, Hanxiao and Savva, Manolis and Chang, Angel Xuan},
  journal={arXiv preprint arXiv:2303.14087},
  year={2023}
}

@article{ravi2024sam2,
  title={SAM 2: Segment Anything in Images and Videos},
  author={Ravi, Nikhila and Gabeur, Valentin and Hu, Yuan-Ting and Hu, Ronghang and Ryali, Chaitanya and Ma, Tengyu and Khedr, Haitham and R{\"a}dle, Roman and Rolland, Chloe and Gustafson, Laura and Mintun, Eric and Pan, Junting and Alwala, Kalyan Vasudev and Carion, Nicolas and Wu, Chao-Yuan and Girshick, Ross and Doll{\'a}r, Piotr and Feichtenhofer, Christoph},
  journal={arXiv preprint arXiv:2408.00714},
  year={2024}
}

@article{fujimura2025ufv,
  title={UFV-Splatter: Pose-Free Feed-Forward 3D Gaussian Splatting Adapted to Unfavorable Views},
  author={Fujimura, Yuki and Kushida, Takahiro and Kitano, Kazuya and Funatomi, Takuya and Mukaigawa, Yasuhiro},
  journal={arXiv preprint arXiv:2507.22342},
  year={2025}
}

@INPROCEEDINGS{rusu2009fpfh,
  author={Rusu, Radu Bogdan and Blodow, Nico and Beetz, Michael},
  booktitle=ICRA, 
  title={Fast Point Feature Histograms (FPFH) for 3D registration}, 
  year={2009},
  volume={},
  number={},
  pages={3212-3217},
  doi={10.1109/ROBOT.2009.5152473}
}

@inproceedings{li2019usip,
  title={Usip: Unsupervised stable interest point detection from 3d point clouds},
  author={Li, Jiaxin and Lee, Gim Hee},
  booktitle=ICCV,
  pages={361--370},
  year={2019}
}

@inproceedings{zohaib2023sc3k,
  title={Sc3k: Self-supervised and coherent 3d keypoints estimation from rotated, noisy, and decimated point cloud data},
  author={Zohaib, Mohammad and Del Bue, Alessio},
  booktitle=ICCV,
  pages={22509--22519},
  year={2023}
}

@inproceedings{groueix20183d,
  title={3d-coded: 3d correspondences by deep deformation},
  author={Groueix, Thibault and Fisher, Matthew and Kim, Vladimir G and Russell, Bryan C and Aubry, Mathieu},
  booktitle=ECCV,
  pages={230--246},
  year={2018}
}

@inproceedings{deng2021deformed,
  title={Deformed implicit field: Modeling 3d shapes with learned dense correspondence},
  author={Deng, Yu and Yang, Jiaolong and Tong, Xin},
  booktitle=CVPR,
  pages={10286--10296},
  year={2021}
}

@inproceedings{cao2023self,
  title={Self-supervised learning for multimodal non-rigid 3d shape matching},
  author={Cao, Dongliang and Bernard, Florian},
  booktitle=CVPR,
  pages={17735--17744},
  year={2023}
}

@inproceedings{chang2024gaussreg,
  title={Gaussreg: Fast 3d registration with gaussian splatting},
  author={Chang, Jiahao and Xu, Yinglin and Li, Yihao and Chen, Yuantao and Feng, Wensen and Han, Xiaoguang},
  booktitle=ECCV,
  pages={407--423},
  year={2024},
  organization={Springer}
}

@inproceedings{le2024integrating,
  title={Integrating efficient optimal transport and functional maps for unsupervised shape correspondence learning},
  author={Le, Tung and Nguyen, Khai and Sun, Shanlin and Ho, Nhat and Xie, Xiaohui},
  booktitle=CVPR,
  pages={23188--23198},
  year={2024}
}

@inproceedings{chen2025dv,
  title={DV-Matcher: Deformation-based Non-Rigid Point Cloud Matching Guided by Pre-trained Visual Features},
  author={Chen, Zhangquan and Jiang, Puhua and Huang, Ruqi},
  booktitle=CVPR,
  pages={27264--27274},
  year={2025}
}

@inproceedings{schoenberger2016vote,
    author={Sch\"{o}nberger, Johannes Lutz and Price, True and Sattler, Torsten and Frahm, Jan-Michael and Pollefeys, Marc},
    title={A Vote-and-Verify Strategy for Fast Spatial Verification in Image Retrieval},
    booktitle=ACCV,
    year={2016},
}

@inproceedings{liu2024singapo,
  title={SINGAPO: Single Image Controlled Generation of Articulated Parts in Objects},
  author={Liu, Jiayi and Iliash, Denys and Chang, Angel X and Savva, Manolis and Amiri, Ali Mahdavi},
  booktitle=ICLR,
  year={2024}
}

@article{chen2024urdformer,
  title={Urdformer: A pipeline for constructing articulated simulation environments from real-world images},
  author={Chen, Zoey and Walsman, Aaron and Memmel, Marius and Mo, Kaichun and Fang, Alex and Vemuri, Karthikeya and Wu, Alan and Fox, Dieter and Gupta, Abhishek},
  journal={arXiv preprint arXiv:2405.11656},
  year={2024}
}

@article{wu2026urdf,
  title={Urdf-anything+: Autoregressive articulated 3d models generation for physical simulation},
  author={Wu, Zhuangzhe and Xin, Yue and Hou, Chengkai and Chen, Minghao and Lyu, Yaoxu and Zhang, Jieyu and Zhang, Shanghang},
  journal={arXiv preprint arXiv:2603.14010},
  year={2026}
}

@inproceedings{le2025articulate,
  title={Articulate-anything: Automatic modeling of articulated objects via a vision-language foundation model},
  author={Le, Long and Xie, Jason and Liang, William and Wang, Hung-Ju and Yang, Yue and Ma, Yecheng Jason and Vedder, Kyle and Krishna, Arjun and Jayaraman, Dinesh and Eaton, Eric},
  booktitle=ICLR,
  volume={2025},
  pages={17578--17602},
  year={2025}
}

@article{li2025art,
  title={ART: Articulated Reconstruction Transformer},
  author={Li, Zizhang and Zhang, Cheng and Li, Zhengqin and Howard-Jenkins, Henry and Lv, Zhaoyang and Geng, Chen and Wu, Jiajun and Newcombe, Richard and Engel, Jakob and Dong, Zhao},
  journal={arXiv preprint arXiv:2512.14671},
  year={2025}
}
